\RequirePackage{fix-cm}
\documentclass[twocolumn]{svjour3}
\smartqed  % flush right qed marks, e.g. at end of proof
\usepackage[ruled,linesnumbered]{algorithm2e}
\usepackage{algpseudocode}
\usepackage{listings}

\usepackage{graphicx}
\usepackage{booktabs}
\usepackage{colortbl}
\usepackage{makecell}
\usepackage{tikz,pgfplots,pgfplotstable}
\usepackage{color}
\usepackage{amsfonts}
\usepackage{amssymb}
\usepackage{multirow}
\usepackage{amsmath}
\usepgfplotslibrary[groupplots]
\definecolor{mediumelectricblue}{rgb}{0.12,0.314,0.588}
\definecolor{mossgreen2}{RGB}{138,154,91}
\definecolor{internationalorange}{RGB}{100,31,0}
\definecolor{lightsalmon}{RGB}{255,160,122}

\definecolor{selfblue}{RGB}{89,138,234}
\definecolor{selfgreen}{RGB}{133,235,133}
\definecolor{darkgreen}{rgb}{0.0, 0.5, 0.0}
\usepackage[pagebackref=false,breaklinks=true,colorlinks,bookmarks=false]{hyperref}

\begin{document}

% \title{UFAFormer: A Unified Transformer Framework for Detecting and Grounding Multi-Modal Manipulation%\thanks{Grants or other notes

\title{GraphBEV++: Multi-Modal Feature Alignment for Autonomous Driving}

\author{Ziying Song$^{1,2}$ \and Caiyan Jia$^{1\ast}$ \and Lin Liu$^{1}$     \and Shaoqing Xu$^3$ \and Lei Yang$^4$  \and Yadan Luo$^{5\ast}$}

% \authorrunning{Short form of author list} % if too long for running head
\institute{
$^\ast$Corresponding Author: Caiyan Jia, Yadan Luo\\
$^1$Beijing Key Laboratory of Traffic Data Mining and Embodied Intelligence, School of Computer Science and Technology, Beijing Jiaotong University Email: \{songziying, cyjia\}@bjtu.edu.cn\\
$^2$School of Artificial Intelligence (School of Software), Yanshan University.
$^3$University of Macau \\ $^4$Nanyang Technological University \\
$^5$The University of Queensland
}
\date{Received: date / Accepted: date}

\maketitle

\begin{abstract}
Feature misalignment in BEV perception is a critical yet often overlooked challenge in autonomous driving, especially under calibration uncertainties between LiDAR and camera sensors. To address this issue, we propose a robust multi-modal fusion framework, \textbf{GraphBEV++}, which systematically mitigates projection-induced misalignment. The framework consists of two key modules: LocalAlign-v2 and GlobalAlign-v2.
LocalAlign-v2 introduces neighborhood-aware depth features via graph matching to correct local misalignment. It supports both LSS-based and query-based BEV representations, making it compatible with BEVFusion and BEVFormer architectures for consistent cross-paradigm alignment. GlobalAlign-v2 encompasses two variants: Deformable and Diffusion. The Deformable variant addresses global misalignment in LSS-based multi-modal BEV by explicitly learning cross-modal feature offsets. In contrast, the Diffusion variant targets implicit misalignment in query-based BEV by injecting noise to simulate misalignment and employing a denoising process to recover aligned features.
Experimental results show that GraphBEV++ achieves state-of-the-art performance under misalignment noise on nuScenes and Waymo subset, improves long-range detection on Argoverse2, and generalizes effectively to the \textcolor{black}{3D occupancy prediction task}, consistently improving occupancy estimation accuracy and robustness under both clean and noisy settings. Furthermore, GraphBEV++ effectively alleviates misalignment issues in end-to-end autonomous driving. Compared with five baselines (UniAD, VAD, FusionAD, MomAD, and WoTE), it demonstrates superior performance in both open-loop (nuScenes) and closed-loop (Bench2Drive and NAVSIM) evaluations across perception, prediction, and planning tasks.  
% Code is publicly available at: \url{https://github.com/adept-thu/GraphBEVplus}.

\keywords{Autonomous Driving \and Multi-Modal Fusion \and Feature Alignment  \and Bird's-Eye View.}
\end{abstract}
\section{Introduction}\label{sec:introduction}

\begin{figure}[t]
\centering
 \includegraphics[width=1.0\linewidth]{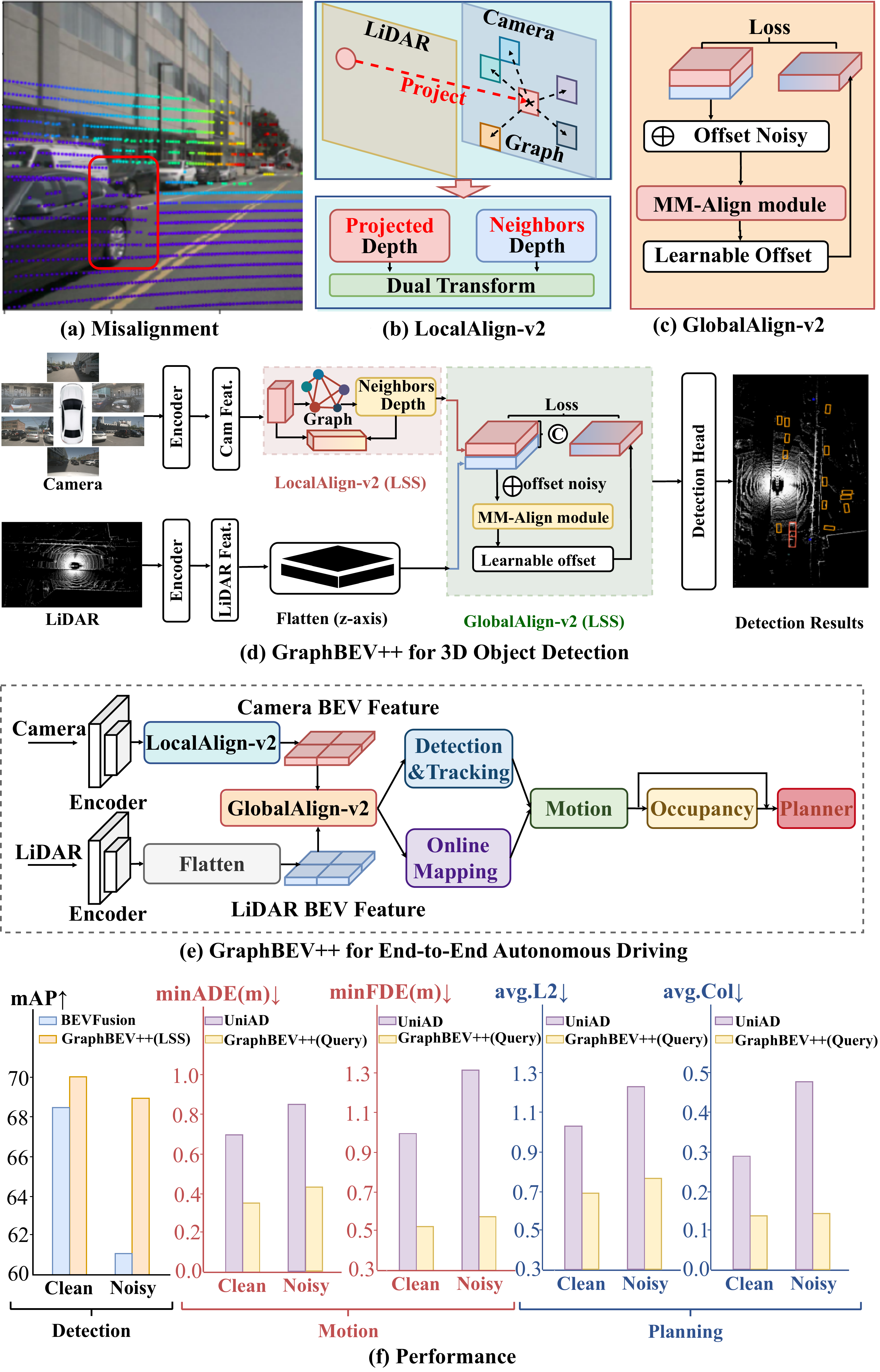}
\caption[]{\textbf{(a)} \textbf{Feature misalignment} often arises from projection matrix errors between LiDAR and camera, leading to inaccurate depth and distorted spatial relationships when projecting LiDAR points onto the image.  \textbf{(b)} \textbf{LocalAlign-v2} addresses local feature misalignment by encoding the projected neighborhood depth features through learnable representations. \textbf{(c)} GlobalAlign-v2 mitigates global BEV misalignment via learned offsets (Deformable) or diffusion-based feature alignment.\textbf{(d)}  The GraphBEV++ framework for 3D object detection effectively resolves BEV feature misalignment in both types of BEV representations: LSS-based (as in BEVFusion) and Query-based (as in BEVFormer). \textbf{(e)} GraphBEV++ systematically investigates the impact of feature misalignment on end-to-end autonomous driving and effectively mitigates it through principled BEV alignment mechanisms.
\textbf{(f)} Empirical results demonstrate that GraphBEV++ effectively mitigates the impact of feature misalignment in both 3D object detection and end-to-end autonomous driving tasks.
}
\label{fig:motivation}
\end{figure}

In autonomous driving systems, multi-modal  fusion plays a critical role. Different sensors provide complementary perceptual capabilities: cameras offer rich semantic and texture information that is valuable for recognizing fine-grained features such as traffic signs, lane markings, and pedestrians, while LiDAR provides accurate 3D geometric information that enables reliable estimation of object distances and spatial structures. Relying on a single modality often leads to perception blind spots or degraded performance—for example, cameras may fail under low-light or adverse weather conditions, whereas LiDAR suffers from sparsity and lacks semantic richness. Therefore, effectively fusing multi-modal data to leverage the strengths of each modality is essential for enhancing the robustness, accuracy, and safety of perception systems, and serves as a foundational technology for achieving high-level autonomous driving~\cite{wang2023multi}.

% 从多模态引出不对齐
Multi-modal fusion has evolved from early point-level~\cite{mvx-net,pointpainting,epnet,epnet++,wang2021pointaugmenting,mvp} and feature-level~\cite{VoxelNextFusion,autoalign,bi2024dyfusion,autoalignv2,graphalign,graphalign++}  methods to the currently prevalent methods of Bird’s-Eye View (BEV) fusion like BEVFusion~\cite{bevfusion-pku,bevfusion-mit}. Although current BEV methods are effective on clean datasets like nuScenes~\cite{nuscenes}, their performance deteriorate on misaligned data. 
This performance decline is primarily due to calibration errors between LiDAR and camera, exacerbated by factors like road vibrations~\cite{yukaicheng_benchmarking}. Feature misalignment presents a major challenge in practical multi-modal fusion tasks, especially when integrating LiDAR and camera data~\cite{yukaicheng_benchmarking,zhujun_benchmarking}, as illustrated in Figure~\ref{fig:motivation}. Such misalignment~\cite{yukaicheng_benchmarking,zhujun_benchmarking,song2024robustness,feng2020deep,feng2020leveraging,shin2019roarnet,bijelic2020seeing,drews2022deepfusion,pfeuffer2018optimal,frossard2021strobe} arises from a range of factors, including: (1) inaccuracies in the extrinsic calibration matrix between LiDAR and camera sensors; (2) temporal desynchronization between asynchronous sensing modalities, such as spinning LiDAR and event-based cameras; (3) calibration noise induced by mechanical vibrations or sensor mounting instability; (4) discrepancies in the fields of view across different sensors; and (5) modality-specific sensitivity to external factors such as lighting and adverse weather conditions. These challenges collectively hinder reliable feature fusion and substantially degrade downstream perception performance.
For instance, adverse weather such as rain or snow can lead to signal scattering and reflection in LiDAR, resulting in sparse or noisy point clouds. This, in turn, increases the depth estimation error—especially for distant or boundary objects—which severely affects the LiDAR-to-Camera projection quality.

\begin{table}[t]
\centering
\caption{Correspondence between GraphBEV++ variants, downstream tasks, and alignment modules.}
\renewcommand\arraystretch{1.1}
\tabcolsep=0.3mm
\resizebox{\linewidth}{!}{
\begin{tabular}{l|l|l}
\toprule
Version & Tasks & Alignment Modules \\
\midrule

\multirow{2}{*}{GraphBEV++ (LSS)}
& 3D Object Detection
& \multirow{2}{*}{LocalAlign-v2 (LSS) + GlobalAlign-v2 (Deformable)} \\
& BEV Segmentation
& \\
\midrule

\multirow{3}{*}{GraphBEV++ (Query)}
& 3D Object Detection
& \multirow{3}{*}{LocalAlign-v2 (Query) + GlobalAlign-v2 (Diffusion)} \\
& 3D Occupancy Prediction
& \\
& End-to-End Autonomous Driving
& \\
\bottomrule

\end{tabular}}
\label{tab_Correspondence}
\end{table}

% 特征不对齐在多模态解决方法
Most feature-level multi-modal methods~\cite{DeepFusion,autoalign,Transfusion,DeepInteraction,CAT-Det} employ the Cross Attention operation to query features of a specific modality, circumventing the need for projection matrices. A few feature-level multi-modal methods~\cite{3dcvf,graphalign,graphalign++,autoalignv2,HMFI,Robust-FusionNet,Logonet,3DDualFusion} have sought to mitigate these errors through the use of projection offsets or neighboring projections. A few BEV-based methods, such as ObjectFusion~\cite{ObjectFusion}, eliminate the camera-to-BEV transformation during fusion to align object-centric features across different modalities. MetaBEV~\cite{ge2023metabev} utilizes the Cross Deformable Attention for feature misalignment, but overlooks depth estimation errors in view transformation and aligns features only during LiDAR and camera BEV fusion.

% BEV存在的问题

Currently, mainstream BEV representations can be broadly categorized into two types: Lift-Splat-Shoot (LSS)-based methods and Query-based methods~\cite{song2024robustness}. BEVFusion~\cite{bevfusion-mit,bevfusion-pku} and BEVFormer~\cite{bevformer} respectively exemplify these two prominent paradigms. Both approaches face distinct forms of \textbf{feature misalignment} issues when fusing camera and LiDAR data into the BEV space to enhance detection performance.
BEVFusion~\cite{bevfusion-mit,bevfusion-pku} unites camera and LiDAR data in BEV space to enhance detection, but overlooks \textbf{feature misalignment} in real-world applications. It is primarily evident in two aspects. 1) BEVFusion~\cite{bevfusion-mit} transforms multi-image features into a unified BEV representation using BEVDepth's~\cite{bevdepth} explicit depth supervision from LiDAR-to-camera. %While
Although this LiDAR-to-camera strategy offers more reliable depth than LSS~\cite{lss}, it overlooks the misalignment between LiDAR and camera in real-world scenarios, leading to \textbf{local misalignment}. 2) In the LiDAR-camera BEV fusion, the misalignment of BEV features due to depth inaccuracies is overlooked by directly concatenating representations and applying basic convolution, as described in BEVFusion~\cite{bevfusion-mit}, %leading to 
resulting in \textbf{global misalignment}. Moreover, query-based BEV representation methods such as BEVFormer~\cite{bevformer} also suffer from feature misalignment issues. When projecting 3D query points onto the image plane, these methods are highly sensitive to extrinsic calibration errors and depth estimation inaccuracies. Such projection deviations result in discrepancies between the sampled image features and the actual object locations, leading to local misalignment and degraded BEV feature representations.

\textcolor{black}{
It is important to note that the local and global misalignment discussed in this paper correspond to two different but complementary error granularities in BEV perception. Specifically, \textbf{local misalignment} refers to feature-level correspondence errors caused by inaccurate geometric projection, including depth estimation errors, LiDAR-camera calibration noise, and query projection deviations. These errors mainly affect local neighborhood consistency during image-to-BEV transformation and feature sampling. In contrast, \textbf{global misalignment} refers to representation-level inconsistencies after BEV construction, where camera and LiDAR BEV features exhibit spatial shifts, structural distortions, or semantic discrepancies due to the accumulation of local projection errors. In practice, both types of misalignment often coexist and form a hierarchical error propagation process: local projection errors introduced during BEV construction may gradually accumulate and eventually lead to global inconsistencies in fused BEV representations. Therefore, the two alignment modules in GraphBEV++ are designed to operate at different stages and address different error granularities. LocalAlign-v2 focuses on improving local geometric correspondence during BEV feature generation, while GlobalAlign-v2 further aligns heterogeneous BEV representations at the fusion stage. As a result, the two modules are complementary rather than conflicting, jointly providing robust feature alignment under real-world misalignment scenarios.
}

To address the aforementioned feature misalignment challenges, we propose a robust fusion framework, GraphBEV++, which enables reliable 3D object detection and end-to-end autonomous driving, especially under misalignment scenarios. Specifically, to handle local misalignment in the camera-to-BEV transformation, we introduce LocalAlign-v2 module with two variants: LocalAlign-v2 (LSS) and LocalAlign-v2 (Query). LocalAlign-v2 (LSS) operates within the view transformation step of the BEVFusion camera branch, where it leverages LiDAR-to-camera explicit depth supervision and utilizes a graph neural network to extract neighborhood-aware depth features. In contrast, LocalAlign-v2 (Query) is built upon BEVFormer~\cite{bevformer} and addresses projection errors of reference points onto the image plane by aligning 2D-3D correspondences during feature sampling.
To further address global misalignment during BEV-level fusion of LiDAR and camera features, we introduce GlobalAlign-v2 module. This module encodes both the projected LiDAR-to-camera depth and the neighborhood depth through dual-stream encoding to generate a reliable depth representation. It then dynamically estimates spatial offsets to align heterogeneous BEV features from the two modalities. Overall, the corresponding baselines, tasks, and datasets for LocalAlign-v2 and GlobalAlign-v2 are summarized in Table \ref{tab_Correspondence}.

We evaluate GraphBEV++ on the nuScenes~\cite{nuscenes}, Waymo subset~\cite{waymo}, and Argoverse 2~\cite{Argoverse2} benchmark for 3D object detection. Experimental results show that GraphBEV++ achieves state-of-the-art performance in clean settings and improves mAP by 8.3\% over BEVFusion under the misaligned noise setting proposed by Dong et al.~\cite{zhujun_benchmarking}.
In addition to perception tasks, we also investigate feature misalignment in end-to-end autonomous driving, where projection errors not only compromise perception quality but also jeopardize the stability and safety of the entire driving system. We conduct comprehensive evaluations on the nuScenes dataset under extrinsic perturbation settings, using UniAD~\cite{uniad}, FusionAD~\cite{fusionad}, VAD~\cite{jiang2023vad}, MomAD~\cite{momad}, and WoTE~\cite{wote} as baselines. Furthermore, we perform closed-loop evaluations on Bench2Drive~\cite{jia2024bench2drive} and NAVSIM~\cite{dauner2024navsim}, demonstrating the effectiveness and generalizability of our multi-modal fusion strategy in both perception and planning pipelines. We believe that addressing misalignment issues is crucial for mitigating the detrimental impact on multi-modal end-to-end  autonomous driving systems.

In summary, the main contributions of this study are as follows.

\begin{enumerate}
\item 
We propose a robust multi-modal fusion framework, named \textbf{GraphBEV++}, to address feature misalignment arising from projection errors between different sensors. Our framework extends beyond 3D object detection—its original focus—towards more comprehensive autonomous driving tasks, including end-to-end prediction and planning.

\item 
By thoroughly analyzing the root causes of feature misalignment, we design \textbf{LocalAlign-v2} to mitigate local misalignment caused by imprecise depth estimation, and \textbf{GlobalAlign-v2} to correct global misalignment between different BEV features.

\item 
Extensive experiments validate the effectiveness of \textbf{GraphBEV++}, demonstrating competitive performance on nuScenes, Waymo subset and Argoverse2 dataset, and at the close-loop settings for the datasets Bench2Drive and NAVSIM. Notably, GraphBEV++ effectively mitigates feature misalignment on the nuScenes dataset for both 3D object detection and end-to-end autonomous driving tasks, %while also
improving long-range detection performance on Argoverse2. %What's more, the validation studies in closed-loop end-to-end simulation scenarios further demonstrate the effectiveness of our framework.

\end{enumerate}

This work's preliminary version of GraphBEV~\cite{song2024graphbev} was presented at ECCV 2024. Compared to our previous conference version~\cite{song2024graphbev}, this paper introduces the following substantial improvements in both methodology and task scope.

\begin{enumerate}
\item 
\textbf{Module-wise Upgrades.} We extend the original GraphBEV by proposing four alignment variants: LocalAlign-v2 (LSS), LocalAlign-v2 (Query), GlobalAlign-v2 (Deformable), and GlobalAlign-v2 (Diffusion). Specifically, LocalAlign-v2 generalizes the neighbor-based alignment strategy to support both LSS- and query-based BEV paradigms, thereby addressing query-feature misalignment and enabling flexible neighbor retrieval across architectures like BEVDepth and BEVFormer. GlobalAlign-v2 incorporates deformable attention and a diffusion-based denoising mechanism, which simulates BEV noise and refines feature alignment through reverse diffusion, demonstrating improved robustness.
\item 
\textbf{Task-wise Extension.} Beyond 3D object detection and segmentation, GraphBEV++ is extended to support a full-stack end-to-end autonomous driving pipeline, covering perception, mapping, prediction, and planning. We introduce modality-specific alignment strategies for LiDAR-camera and radar-camera pairs, accounting for their inherent disparities in resolution, semantics, and spatial coverage. This design emphasizes the critical role of alignment in downstream driving decisions.
\item 
\textbf{Comprehensive Evaluation.} Compared with the six tasks reported in the original GraphBEV, GraphBEV++ includes 19 benchmarks across diverse tasks: 3D object detection, multi-object tracking, online mapping, occupancy prediction, motion forecasting, and planning trajectory prediction. We introduce radar-camera fusion, expand evaluations to Waymo and Argoverse datasets, and investigate misalignment effects in BEVFormer. For end-to-end driving, both open-loop (nuScenes) and closed-loop scenarios (NAVSIM, Bench2Drive) are covered. These extensive evaluations validate the robustness and generalizability of GraphBEV++, while offering valuable resources to the research community for benchmarking multi-modal alignment and planning systems.
\end{enumerate}

\section{Related Work}\label{sec:related_work}
\subsection{LiDAR-based 3D Object Detection}
LiDAR-based 3D object detection methods can be categorized into three primary types based on point cloud representation: Point-based, Voxel-based, and PV-based (Point-Voxel). 
Point-based methods~\cite{lidarrcnn,Frustumpointnets,Pointnet,Pointnet++,Pointrcnn} extend PointNet's~\cite{Pointnet,Pointnet++} principle, directly processing raw point clouds with stacked Multi-Layer Perceptrons (MLPs) to extract point features.
VoxelNet~\cite{Voxelnet} is innovated by partitioning raw point clouds into uniform voxel grids. 
Voxel-based methods~\cite{Voxelnet,Voxelrcnn,Second,LargeKernel3D,SATGCN} typically convert point clouds into voxels and apply 3D sparse convolutions for voxel feature extraction. In addition, PointPillars~\cite{Pointpillars} converts irregular raw point clouds into pillars and encodes them on a 2D backbone, achieving a very high FPS. Some Voxel-based methods~\cite{vsettransformer,voxeltransformer,Sparsetransformer} further exploit Transformers~\cite{transformer} post-voxelization to capture long-range voxel relationships. PV-based methods~\cite{vpnet,pvcnn,PVGNet,pvrcnn++,Std} combine voxel and point-based strategies and extract features from point clouds' diverse representations using both approaches, achieving higher accuracy albeit with increased computational demand.

\subsection{Camera-based 3D Object Detection}
Camera-based 3D object detection methods have gained increasing attention in academia and industry, mainly due to the significantly lower cost of camera sensors compared to LiDAR~\cite{song2024robustness}. Early methods~\cite{brazil2019m3d,xu2018multi,simonelli2019disentangling} have focused on augmenting 2D object detectors with additional 3D bounding box regression heads. Current camera-based methods have rapidly evolved since LSS~\cite{lss} introduced the concept of unifying multi-view information onto a BEV through ``Lift and splat''. LSS-based methods~\cite{lss,bevdepth,BEVStereo,park2022time,yang2023bevheight,yang2023bevheight++} like BEVDepth~\cite{bevdepth} extract 2D features from multi-view images and provide effective depth supervision via LiDAR-to-camera projections before unifying multi-view features onto the BEV. Subsequent works~\cite{BEVStereo,park2022time} have introduced multi-view stereo techniques to improve depth estimation accuracy and achieve SOTA (state-of-the-art) performance. Additionally, inspired by the success of transformer-based architectures such as DETR~\cite{detr} and Deformable DETR~\cite{deformabledetr} in 2D detection, transformer-based detectors have emerged for 3D object detection. Following DETR3D~\cite{wang2022detr3d}, some methods design a set of object queries~\cite{jiang2023polarformer,liu2022petr,liu2023petrv2} or BEV grid queries~\cite{bevformer,bevformerv2}, then perform view transformation through cross-attention between queries and image features.

\subsection{Multi-modal 3D Object Detection}
Multi-modal 3D object detection refers to using data features from different sensors and integrating these features to achieve complementarity, thus enabling the detection of 3D objects. Previous multi-modal methods can be coarsely classified into three types, i.e., point-level, feature-level, and BEV-based methods. Point-level methods~\cite{mvx-net,pointpainting,epnet,epnet++,wang2021pointaugmenting,mvp} and feature-level methods~\cite{VoxelNextFusion,autoalign,bi2024dyfusion,autoalignv2,graphalign,graphalign++,zhang2023urformer} typically leverage image features to augment LiDAR points or 3D object proposals. BEV-based methods~\cite{bevfusion-mit,bevfusion-pku,ge2023metabev,ObjectFusion,song2024contrastalign} efficiently unify the representations of LiDAR and camera into BEV space. Although BEVFusion~\cite{bevfusion-mit,bevfusion-pku} achieve high performance, they are typically tested on clean datasets like nuScenes~\cite{nuscenes}, overlooking real-world complexities, especially \textbf{feature misalignment}, which hampers their applications.

\subsection{3D Occupancy Prediction}
\textcolor{black}{3D occupancy prediction has emerged as a promising paradigm for holistic scene understanding in autonomous driving, as it provides a unified representation of geometry and semantics in 3D space~\cite{cao2022monoscene,zheng2024occworld,huang2023tri,zhang2023occformer}. Existing methods have explored diverse scene representations for occupancy estimation, ranging from dense voxel grids~\cite{cao2022monoscene,li2023voxformer,zhang2023occformer} to more efficient structures such as Tri-Perspective View (TPV)~\cite{huang2023tri}, 3D Gaussians~\cite{huang2024gaussianformer}, point-based representations~\cite{zuo2023pointocc}, and fully sparse occupancy frameworks~\cite{liu2024fully}. In parallel, large-scale benchmarks and comprehensive occupancy frameworks have been developed to advance occupancy perception in real-world autonomous driving scenarios~\cite{tian2023occ3d,wang2023openoccupancy,gan2024comprehensive}. 
\textcolor{black}{\textit{Gan et al.\cite{gan2024comprehensive} present
a comprehensive framework for 3D occupancy estimation, which
reveals several key components for 3D occupancy estimation, such
as network design, optimization, and evaluation.
SparseOcc \cite{liu2024fully} introduces a fully sparse occupancy framework for efficient 3D scene understanding.
These advances have significantly improved the accuracy and efficiency of 3D scene representation and understanding.}}
Despite these successes, feature misalignment remains an underexplored challenge in occupancy prediction. Since occupancy estimation relies on multi-view feature projection and aggregation, inaccurate geometric alignment may directly affect occupancy quality. Although GraphBEV++ is designed for multi-modal 3D object detection, the proposed LocalAlign-v2 and GlobalAlign-v2 modules provide a general alignment mechanism that could potentially benefit occupancy prediction tasks. Exploring robust feature alignment for occupancy estimation is an interesting direction for future research.}

\subsection{End-to-end Autonomous Driving}
Recent works in autonomous driving have shifted towards exploring end-to-end tasks~\cite{e2etpamisurvey}. These works~\cite{uniad,fusionad,sun2024sparsedrive,jiang2023vad,chen2024vadv2,jia2023driveadapter,yang2024deepinteraction++} are now designed to execute integrated tasks encompassing perception, prediction, and planning, spanning the entire process from scene perception to ego-planning. The advantage of end-to-end autonomous driving lies in its ability to provide interpretable intermediate results and avoids local optima by considering the system holistically, leading to significant breakthroughs in planning tasks~\cite{e2etpamisurvey}. However, most studies rely on a single modality (especially cameras) and ignore the impact of extrinsic calibration changes causing feature misalignment, which can disrupt end-to-end tasks. To address this, we propose the multi-modal end-to-end framework GraphBEV++ that improves feature alignment through graph matching, enhancing its robustness in real-world scenarios and offering significant potential for the future deployment of end-to-end autonomous driving systems.

\section{Method}\label{sec:method}

\begin{figure*}[t]
\centering
\includegraphics[width=1\textwidth]{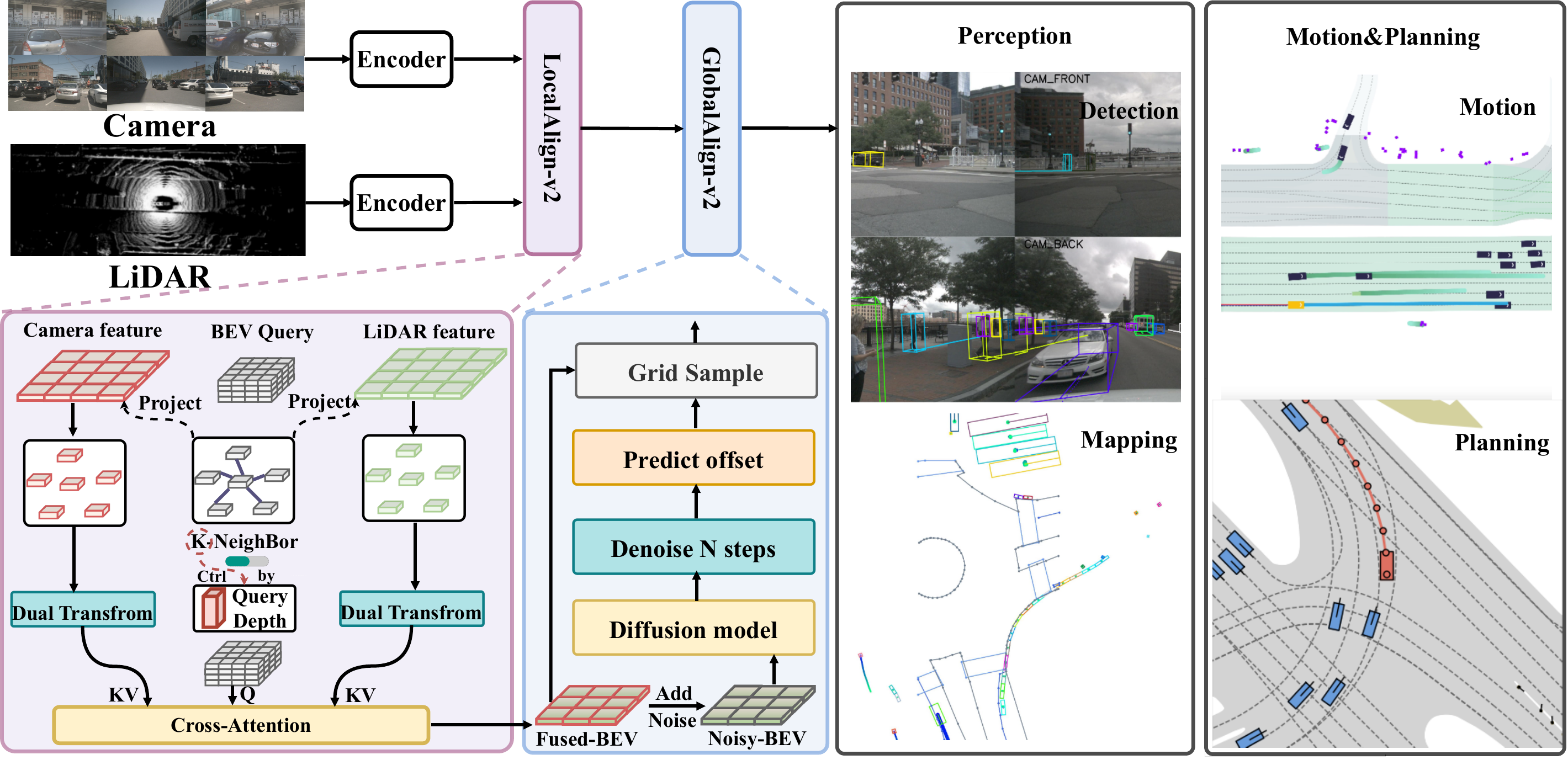 }
\caption[]{The overview of \textbf{GraphBEV++} within an end-to-end autonomous driving framework. We primarily demonstrate our method based on the multi-modal end-to-end baseline FusionAD, aiming to mitigate the impact of BEV feature misalignment on downstream autonomous driving tasks.}
\label{fig:frameworkv2}
\end{figure*}

To address the challenge of feature misalignment in autonomous driving tasks~\cite{bevfusion-mit,bevfusion-pku,uniad}, we propose a robust multi-modal fusion framework, GraphBEV++, designed for both 3D object detection and end-to-end autonomous driving. GraphBEV++ is highly extensible and supports both Lift-Splat-Shoot (LSS)-based and query-based BEV representations, effectively mitigating feature misalignment across modalities. The overall architecture for end-to-end autonomous driving is illustrated in Figure~\ref{fig:frameworkv2}.
GraphBEV++ takes inputs from multiple sensors, including LiDAR and cameras, and extracts modality-specific features using dedicated encoders. We design the LocalAlign-v2 module to transform image features into the BEV space, alleviating local misalignment caused by projection errors between LiDAR and camera in conventional BEV fusion frameworks~\cite{bevfusion-mit,bevfusion-pku}. Additionally, we introduce the GlobalAlign-v2 module to further correct global misalignment between LiDAR and camera BEV features during the fusion process.
% \begin{figure*}[t]
% \centering
% \includegraphics[width=1\textwidth]{mainv3.pdf }
% \caption[]{The overview of \textbf{GraphBEV++} framework. The LiDAR branch largely follows the baselines~\cite{bevfusion-mit,Transfusion} to generate LiDAR BEV features. In the camera branch, we first extract camera BEV features using the proposed LocalAlign-v2 module, which aims to address local misalignment due to sensor calibration errors. Subsequently, we simulate the offset noise of LiDAR and camera BEV features, followed by aligning global multi-modal features through learnable offsets. Notably, we only add offset noise to the GlobalAlign-v2 module during \textbf{training} to simulate global misalignment issues. Finally, we employ the dense detection head in~\cite{Transfusion} to accomplish 3D detection tasks.
% }
% \label{fig:framework}
% \end{figure*}

\begin{figure}[t]
\centering
\includegraphics[width=1.0\linewidth]{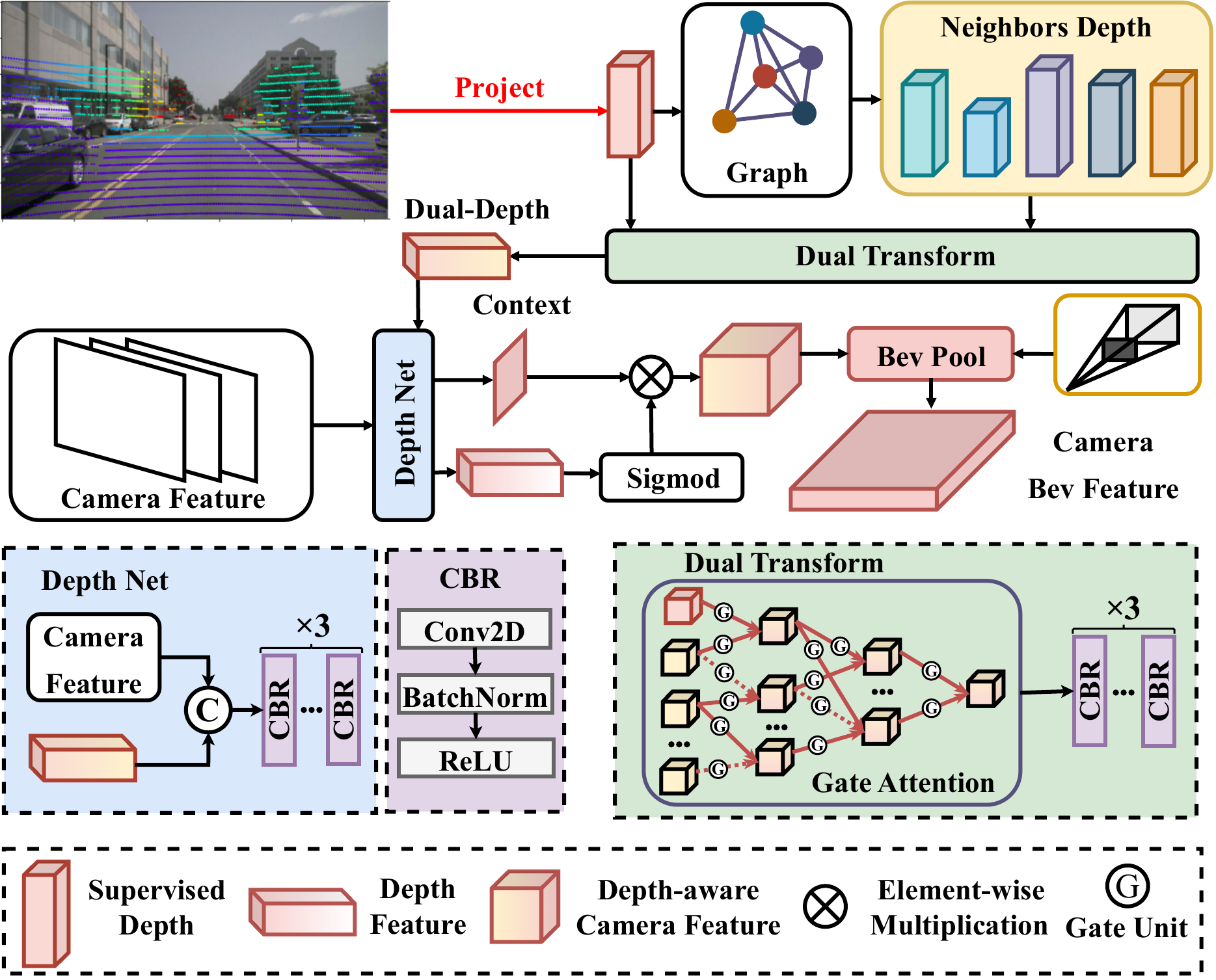}
\caption[]{Overview of the \textbf{LocalAlign-v2 (LSS)} pipeline. The LocalAlign-v2 (LSS) module mitigates local misalignment in LSS-based BEV representations by enhancing the camera-to-BEV transformation. It incorporates neighboring depth features obtained via KD-Tree–based nearest-neighbor search to refine LiDAR-to-camera projections.} 
\label{fig:LocalAlign_lss}
\end{figure}
\begin{figure}
\centering
\includegraphics[width=1\linewidth]{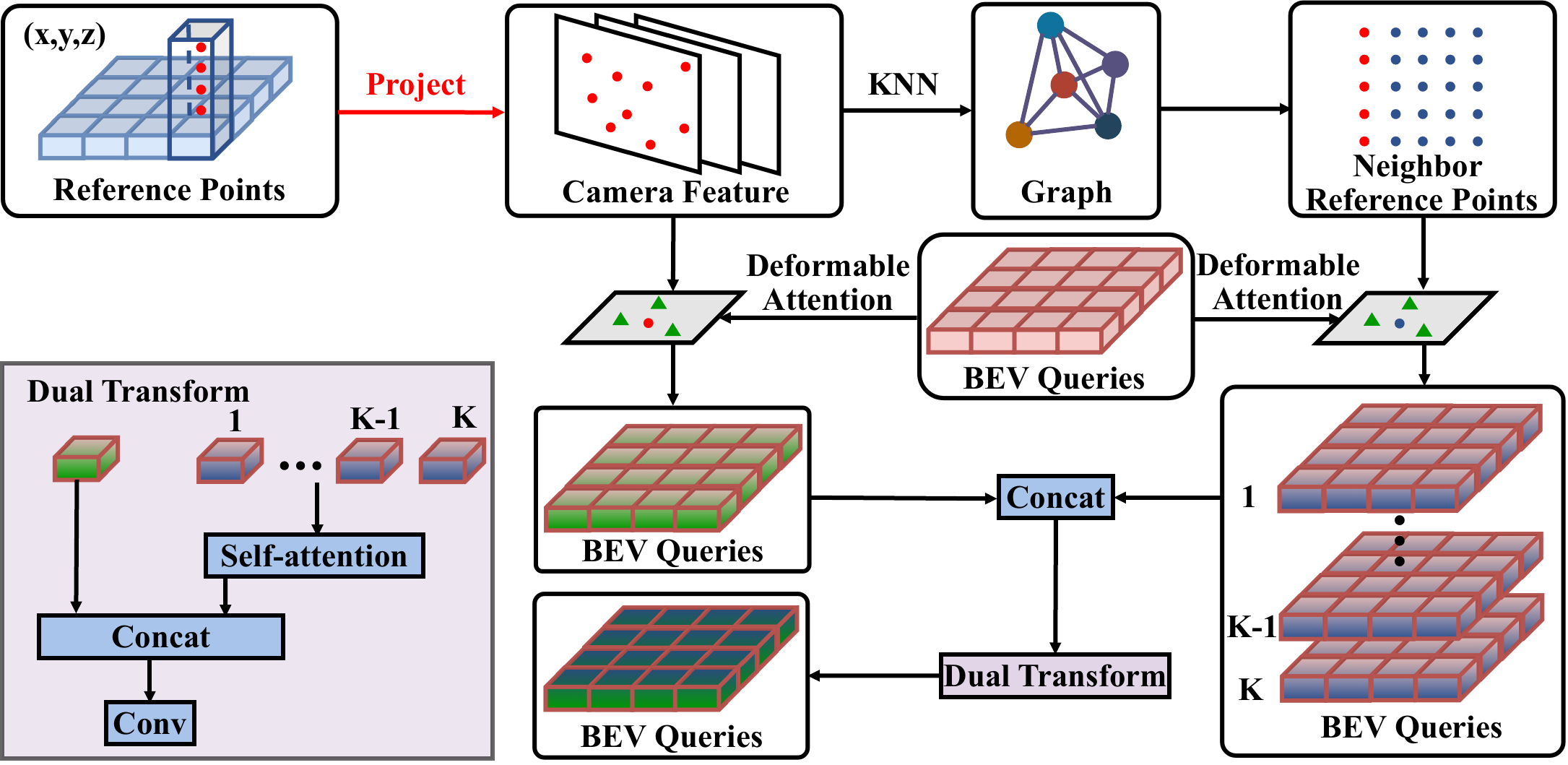}
\caption[center]{Overview of the \textbf{LocalAlign-v2 (Query)} pipeline. The LocalAlign-v2 (Query) module addresses projection-induced local misalignment in query-based BEV representations. It refines the image-to-BEV transformation by leveraging adjacent BEV query features, with neighborhood relations established via KD-Tree.   } 
\label{fig:LocalAlign_query}
\end{figure}

\subsection{LocalAlign-v2}
LocalAlign-v2 is designed to address local misalignment caused by projection discrepancies between heterogeneous sensors. It is compatible with various BEV representations, including both LSS-based and query-based paradigms. In the following, we present the details of \textbf{LocalAlign-v2 (LSS)} and \textbf{LocalAlign-v2 (Query)}, as shown in Figure~\ref{fig:LocalAlign_lss} and Figure~\ref{fig:LocalAlign_query}, respectively.

\subsubsection{LocalAlign-v2 (LSS)}
To facilitate the transformation of camera features into BEV features, 
the LSS-based methods like  BEVFusion~\cite{bevfusion-mit,bevfusion-pku}  leverage LiDAR-to-camera to provide projected depth, thereby enabling the fusion of depth and image features. In the process of camera-to-BEV, the methods like BEVFusion~\cite{bevfusion-mit,bevfusion-pku} 
operate under the assumption that the depth information provided by LiDAR-to-camera projection is accurate and reliable. However, they overlook the complexities inherent in real-world scenarios, where most of projection matrices between LiDAR and cameras are calibrated manually. Such calibration inevitably introduces projection errors, leading to \textbf{depth misalignment}—where the depths of surrounding neighbors are projected as the pixel's depth. This depth misalignment results in inaccuracies within the depth features, causing \textbf{local misalignment}  during multi-view transformation into BEV representations. %Given that LSS-based methods~\cite{lss,bevdepth} rely on depth estimation from pixel-level features with inaccuracies in detailing, this leads to \textbf{local misalignment} within camera BEV features. 
It underscores the challenges of ensuring precise depth estimation within BEVFusion~\cite{bevfusion-mit} and highlights the importance of robust methods to address projection errors.
%我们提出了LocalAlign-v2 来解决

Therefore, we propose a LocalAlign-v2 (LSS) module to address local misalignment, with its pipeline depicted in Figure~\ref{fig:LocalAlign_lss}. Specifically, LiDAR-to-camera provides projected depth, defined as \(D_{S} \in \mathbb{R}^{B_S \times N_{C} \times 1 \times H \times W}\), where \(B_S\) represents the batch size, \(N_{C}\) denotes the number of multi-views (six in the case of nuScenes), and \(H\) and \(W\) are the height and the width of images, respectively. The projection from LiDAR-to-camera maps 3D point clouds onto an image plane, from which we can obtain the indices of the projected pixels, defined as \(M_{\text{Coords}} \in \mathbb{R}^{N_{P} \times 2}\), where \(N_{P} \) refers to the number of points projected onto pixels, and 2 represents the pixel coordinates (\(u, v\)) as illustrated below.
\begin{align}\label{equ3d2d}
z_{c}\left[\begin{array}{c}
u \\
v \\
1
\end{array}\right]=h \mathcal{K}\left[\begin{array}{ll}
R & T
\end{array}\right]\left[\begin{array}{c}
P_{x} \\
P_{y} \\
P_{z} \\
1
\end{array}\right],
\end{align}
where $P_{x}$, $P_{y}$, $P_{z}$ denote the LiDAR point's 3D location, $(u, v)$ denotes the corresponding 2D location, and $z_{c}$ represents the depth of its projection on the image plane, $\mathcal{K}$ denotes the camera intrinsic parameter, $R$ and $T$ denote the rotation and the translation of the LiDAR with respect to the camera reference system, and $h$ denotes the scale factor due to down-sampling.

\begin{algorithm}[t]
\SetAlgoLined
\caption[]{Graph for Finding Neighbors} \label{algorithm:KD-Tree}
\KwIn{

The indices of the projected pixels \(M_{\text{Coords}} \in \mathbb{R}^{N_{P} \times 2}\).

Hyper-parameter: Number of neighbors $K_{\text{graph}} = 8$.
}
\KwOut{Neighbors $M_{K_{\text{Coords}}} \in \mathbb{R}^{N_{P} \times K_{\text{graph}} \times 2}$.
}

\While{LocalAlign-v2}{
\SetKwProg{Fn}{Function}{}{}
\Fn{KD-Tree ($M_{\text{Coords}}$, ${M_{\text{Coords}}}_{i}$, $K_{\text{graph}}$)}{

        Compute the Euclidean distance between ${M_{\text{Coords}}}_{i}$ and $M_{\text{Coords}}$
        
        Indices = argsort(distances)
        
    \Return  $M_{\text{Coords}}[1:K_{\text{graph}}]$
}

\For{$i=1 \dots   N_{P} $}{
    Neighbors = KD-Tree($M_{\text{Coords}}$, ${M_{\text{Coords}}}_{i}$, $K_{\text{graph}}$)
    
    ${M_{\text{Coords}}}_{i}$ = Neighbors
}

}

\end{algorithm}

We employ the KD-Tree algorithm to obtain the indices of the projected pixels' neighbors, defined as $M_{K_{\text{Coords}}} \in \mathbb{R}^{N_{P} \times K_{\text{graph}} \times 2}$, where $K_{\text{graph}}$ denotes the number of neighbors for each projected pixel. The process is outlined in Algorithm (\ref{algorithm:KD-Tree}). It is worth noting that we simplify the process of the KD-Tree algorithm, and the code can be referred to %in
scipy\footnote{\url{https://github.com/minrk/scipy-1/blob/master/scipy/spatial/ckdtree.c}}. Then, we obtain the surrounding neighbor depth \(D_{K} \in \mathbb{R}^{B_S \times N_{C} \times K_{\text{graph}} \times H \times W} \) by indexing \(D_{S}\) with \(M_{\text{Coords}}\). Then, \(D_{S}\) and \(D_{K}\) simultaneously enter the Dual Transform module for deep feature encoding. The shapes of $D_{S}$ and $D_{K}$ are respectively modified to $[B_S \times N_{C} , 1, H, W]$ and $[B_S \times N_{C} , K_{\text{graph}} , H, W]$ before being fed into the Dual Transform module. This module comprises straightforward components, including convolutional layers, Batch Normalization, and ReLU activations, as illustrated in Figure~\ref{fig:LocalAlign_lss}. The outcome of this process is the Dual Depth feature, denoted as $D_{\text{SK}}$, and its shape is $[B_S \times N_{C}, C_{\text{SK}}, \frac{H}{8}, \frac{W}{8}]$. 
The camera Encoder outputs multi-scale image features from the FPN, including $F_{\text{Cam.}} \in \mathbb{R}^{(B_S \times N_{C}) \times C_{\text{Cam}}\times \frac{H}{8} \times \frac{W}{8}}$ for richer semantic information and another at a reduced resolution of $\frac{H}{16}, \frac{W}{16}$. We opt to utilize the feature with the resolution $\frac{H}{8}, \frac{W}{8}$ due to its more comprehensive semantic content. 

We then design DepthNet model 
%The design of DepthNet is straightforward, 
as illustrated in the right upper corner of Figure~\ref{fig:LocalAlign_lss}, 
%We input both 
where $F_{\text{Cam.}}$ and $D_{\text{SK}}$ are fed into DepthNet to fuse depth features with multi-view camera features. 
Initially, $F_{\text{Cam.}}$ and $D_{\text{SK}}$ are concatenated, followed by processing through three sets of CBR-module (see Figure~\ref{fig:LocalAlign_lss}) %
compromising a 2D convolution layer with Batch Normalization and ReLU activations. %
This results in the generation of a new depth-aware camera feature, denoted as $F_{\text{DC}} \in \mathbb{R}^{(B_S \times N_{C}) \times C_{\text{DC}}\times \frac{H}{8} \times \frac{W}{8}}$. 
Subsequently, $F_{\text{DC}}$ is split along the $C_{\text{DC}}$ dimension into two new features: a novel depth feature, define as $\hat{F_{D}}  \in \mathbb{R}^{(B_S \times N_{C}) \times \hat{C_{D}}\times \frac{H}{8} \times \frac{W}{8}}$ and a novel image context feature, define as $\hat{F_{C}}  \in \mathbb{R}^{(B_S \times N_{C}) \times \hat{C_{C}} \times \frac{H}{8} \times \frac{W}{8}}$. 
It's important to note that $C_{\text{DC}}$ = $\hat{C_{C}} $ + $\hat{C_{D}}$, indicating the division of the combined feature space into distinct depth and image feature components. Subsequently, $\hat{F_{D}} $ is subjected to a softmax operation and then multiplied with $\hat{F_{C}} $, resulting in a novel image feature with depth information, represented as $\hat{F}_{\text{DC}} \in \mathbb{R}^{(B_S \times N_{C}) \times \hat{C_{C}}  \times \hat{C_{D}} \times \frac{H}{8} \times \frac{W}{8}}$. Finally, adopting operations consistent with LSS~\cite{lss} and BEVDepth~\cite{bevdepth}, we utilize pre-generated 3D spatial coordinates and \(\hat{F_{DC}}\) with BEV Pooling to output the camera BEV feature, thereby completing the camera-to-BEV transformation, and finally outputs the camera BEV feature, define as $ F^{C}_{B} \in \mathbb{R}^{B_S \times \hat{C_{C}}  \times H_{B} \times W_{B}}$.

\subsubsection{LocalAlign-v2 (Query)}

Following our discussion of how LocalAlign-v2 (LSS) addresses feature misalignment in LSS-based BEV representations such as BEVDepth and BEVFusion, we now turn to LocalAlign-v2 (Query), which focuses on aligning BEV features in query-based architectures like BEVFormer. Although the underlying mechanisms for BEV generation differ between these two paradigms, the core idea of mitigating feature misalignment through neighborhood-aware alignment remains consistent.

As illustrated in Figure~\ref{fig:LocalAlign_query}, BEVFormer first samples 3D query points $P_{3D}^{loc} \in \mathbb{R}^{B_{S}\times N_{P}\times 3}$ and initializes them as BEV queries $Q_{BEV} \in \mathbb{R}^{B_{S}\times N_{P}\times C}$. These query points are projected onto the corresponding images to obtain their 2D coordinates $P_{2D}^{loc} \in \mathbb{R}^{B_{S}\times N_{P}\times 2}$, which are then used in deformable attention to extract image features for BEV construction. However, due to inevitable projection matrix errors in real-world settings, these 2D coordinates are often inaccurate, leading to misalignment between the sampled image features and their true semantic counterparts.

To address this issue, LocalAlign-v2 (Query) introduces a neighborhood-based correction mechanism. For each BEV query, it identifies $K$ neighboring BEV queries $Q_{BEV}^{neighbor} \in \mathbb{R}^{B_{S} \times K \times N_{P} \times C}$ based on proximity in the image plane. These neighbors, along with the original BEV query, are fed into a dual-transform module that leverages a self-attention mechanism to adaptively fuse information from the local neighborhood. This process enhances the robustness and alignment of BEV features by compensating for projection-induced errors.

The significance of LocalAlign-v2 (Query) lies in its ability to generalize feature alignment to the query-based BEV paradigm, which is inherently more sensitive to calibration errors due to its reliance on reference point projections. By introducing a principled and learnable neighborhood-based correction mechanism, GraphBEV++ (Query) effectively extends our framework’s applicability from dense BEV fusion methods to sparse, query-driven architectures. This ensures that GraphBEV++ not only maintains compatibility with a broader class of BEV-based perception systems but also enhances their resilience to real-world misalignment, thereby improving downstream tasks such as detection, prediction, and planning in end-to-end autonomous driving.

\subsubsection{Adaptive KNN  for Efficient  LocalAlign-v2}

Feature misalignment across modalities varies with the spatial scale and location of objects. We observe that large, nearby objects are generally more robust to local misalignment, whereas small or distant objects are more sensitive to projection-induced noise. The original GraphBEV adopts a fixed number of neighbors $K$ in the LocalAlign module, which fails to account for such discrepancies. Applying a uniform $K$ across all points leads to redundant computation for well-aligned regions and may result in overfitting uninformative areas.

To address this limitation, we introduce an \textbf{Adaptive KNN} mechanism that dynamically determines the number of neighbors $K_i$ for each projected point. This allows LocalAlign-v2 to focus more effectively on regions prone to misalignment while improving computational efficiency.

During training, where ground-truth 3D bounding boxes are available, we use object size as a proxy for alignment difficulty. For each projected point $i$, we compute its associated object’s scale $\text{size}_i$ as the average of the bounding box’s length, width, and height. The number of neighbors $K_i$ is then determined by:
\begin{align}
K_i = \text{clip}\left( \frac{\gamma}{\text{size}_i + \epsilon},\; K_{\min},\; K_{\max} \right)
\label{eq:adaptive_knn_train}
\end{align}
where $\gamma$ is a scaling factor, $\epsilon$ prevents division by zero, and $\text{clip}(\cdot)$ constrains $K_i$ to a reasonable range (e.g., $[4, 16]$). This rule ensures smaller or thinner objects receive more neighbors for better alignment, while large, well-perceived objects require fewer.

During inference, ground-truth labels are unavailable. Instead, we use the depth value $z_{c, i}$ of each projected point, obtained via LiDAR-to-camera projection, as an indirect indicator of alignment difficulty. The neighbor count is then computed as:
\begin{align}
K_i = \text{clip}\left( K_{\min} + \alpha \cdot \log(1 + z_{c, i}),\; K_{\min},\; K_{\max} \right)
\label{eq:adaptive_knn_infer}
\end{align}
Here, $\alpha$ controls the sensitivity of $K_i$ to depth. This \textbf{depth-aware adaptation} increases the receptive field for distant or uncertain points, improving robustness without incurring significant overhead.

Overall, the Adaptive KNN strategy offers three key benefits. First, it improves efficiency by avoiding unnecessary computation for well-aligned regions. Second, it enhances robustness by allocating more context to points prone to misalignment. Third, it introduces no additional network parameters and can be seamlessly integrated into standard KNN-based neighbor search pipelines. This approach is also \textbf{backbone-agnostic}, making it compatible with both LSS-based and query-based BEV representations within the LocalAlign-v2 framework.

\subsection{GlobalAlign-v2}
\begin{figure}[t]
\centering
\includegraphics[width=\linewidth]{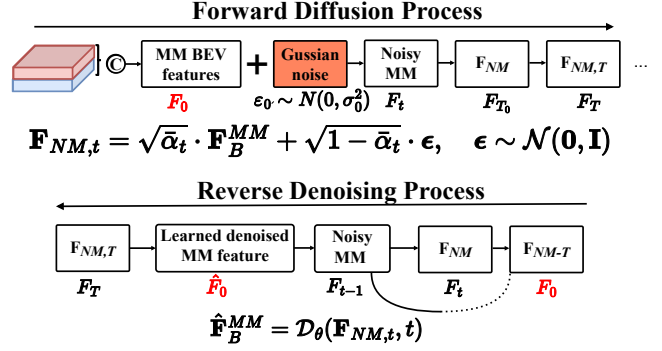}
\caption[]{\textbf{The overview of \textbf{GlobalAlign-v2 (\textcolor{black}{Diffusion})} pipeline.} The GlobalAlign-v2 (Diffusion) module tackles the challenge of global misalignment in query-based multi-modal BEV representations. By treating global misalignment as a form of diffusion noise, the model progressively denoises noisy BEV features to recover robust and well-aligned representations.} 
\label{fig:GlobalAlign_diffusion}
\end{figure}

GlobalAlign-v2 addresses the problem of global feature misalignment in multi-modal BEV fusion. It comprises two main variants: \textbf{GlobalAlign-v2 (Deformable)}, as illustrated in Figure~\ref{fig:GlobalAlign_deformable}, and \textbf{GlobalAlign-v2 (Diffusion)}, as shown in Figure~\ref{fig:GlobalAlign_diffusion}. The Deformable variant is designed for explicit multi-modal BEV representations (e.g., BEVFusion), utilizing deformable convolutions for feature alignment. In contrast, the Diffusion variant targets implicit BEV representations (e.g., BEVFormer), and employs a multi-step diffusion-based denoising process to achieve more robust global alignment.

\subsubsection{GlobalAlign-v2 (Deformable)}

\begin{figure}[t]
\centering
\includegraphics[width=\linewidth]{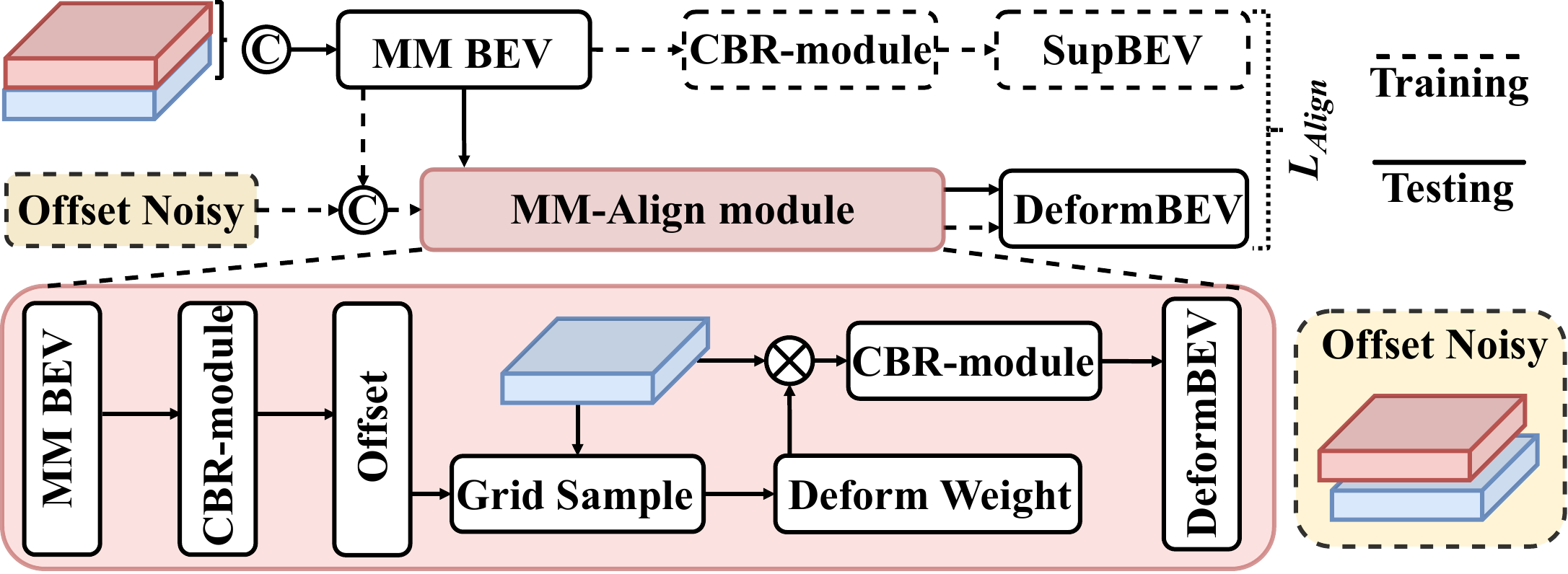}
\caption[]{The overview of \textbf{GlobalAlign-v2 (Deformable)} pipeline. The GlobalAlign-v2 (Deformable) module addresses the issue of LSS-based multi-modal BEV feature misalignment. During training, we add offset noise to simulate the global misalignment problem in %the 
camera and LiDAR BEV features. It is supervised through a simple CBR-module to learn the offsets of camera BEV features. We do not introduce noise during testing and employ learnable offsets for forward inference.} 
\label{fig:GlobalAlign_deformable}
\end{figure}

In the real world, feature misalignment is inevitable due to calibration matrix discrepancies between LiDAR and camera sensors. %While
Although LocalAlign-v2 module mitigates the local misalignment issue in the camera-to-BEV process, deviations may still exist in camera-BEV features. During LiDAR-camera BEV fusion, despite being in the same spatial domain, inaccuracies in depth lead to global misalignment from the view transformer and overlooking of global offsets between LiDAR and camera BEV features.
To tackle the global misalignment issue described above, we introduce the GlobalAlign-v2 (Deformable) module, employing learnable offsets to achieve the alignment of global multi-modal BEV features. As shown in Figure~\ref{fig:GlobalAlign_deformable}, we use clean datasets such as nuScenes~\cite{nuscenes} for training, which exhibit minimal deviations that can be considered negligible. The supervised information is derived from the features obtained after the fusion and convolution of LiDAR and camera BEV features. During training, we introduce global offset noise and employ learnable offsets. In the LiDAR branch, the LiDAR feature is flattened along the Z-axis to form the LiDAR BEV feature, defined as \( F^{L}_{B} \in \mathbb{R}^{B_S \times \hat{C_{L}}  \times H_{B} \times W_{B}} \). Initially, we concatenate \( F^{L}_{B} \) and \( F^{C}_{B} \) to obtain a fused BEV feature, denoted as \( F^{MM}_{B} \in \mathbb{R}^{B \times (\hat{C_{C}}  + \hat{C_{L}} ) \times H_{B} \times W_{B}} \). Subsequently, \( F^{MM}_{B} \) undergoes a convolution operation, resulting in a new fused feature, denoted as \( \hat{F_B} \in \mathbb{R}^{B_S \times \hat{C_{L}}  \times H_{B} \times W_{B}} \). Notably, \( \hat{F_B}\) will be utilized as a supervision signal during the training process.

As shown in Figure~\ref{fig:GlobalAlign_deformable}, we introduce random offset noise to the camera dimension of $ F^{MM}_{B}$ to obtain a new noisy feature $F^{MM}_{N} \in \mathbb{R}^{B_S \times (\hat{C_{C}}  + \hat{C_{L}} ) \times H_{B} \times W_{B}}$, simulating the global misalignment issue originating from camera BEV features. Notably, the LiDAR BEV feature is directly flattened, thus more accurate. Then, $F^{MM}_{N}$ is input into the MM-Align module for global offset learning. $F^{MM}_{N}$ is processed through the CBR-module with basic convolution operations to learn offsets, defined as $ F^{O} \in \mathbb{R}^{B_S \times 2 \times H_{B} \times W_{B}}$, where %two refers 
2 corresponds to the offset coordinates $(u,v)$. Subsequently, LiDAR BEV features $ F^{L}_{B} $ and $ F^{O}$ undergo grid sampling to generate new deform weights, defined as $ F^{D}_{W} \in \mathbb{R}^{B_S \times \hat{C_{L}}  \times H_{B} \times W_{B}} $. The purpose of grid sampling is to utilize offsets for spatial transformation of LiDAR BEV feature $ F^{L}_{B}$, with learnable shifts dynamically adjusting to capture spatial dependencies more flexibly than standard convolution operations. Afterward, $ F^{D}_{W} $ is multiplied by LiDAR BEV features $ F^{L}_{B} $ to dynamically adjust features, followed by standard convolution operations through the CBR-module, culminating in the output Deform BEV defined as $ F^D_B \in \mathbb{R}^{B_S \times \hat{C_{L}} \times H_{B} \times W_{B}} $. Finally, during training, we supervise $F^D_B$ using $\hat{F_B}$ previously mentioned, and employ the $L_{\text{Align}}$ for supervision as follows.

\begin{align}
\mathcal{L}_{\text{Align}} = \frac{1}{N_{B}} \sum_{i=1}^{N_{B}} ({\hat{F_B}}_i - {F^D_B}_i)^2,
\end{align}
where $N_{B} = B_S \times H_{B} \times W_{B}$ represents the total number of elements, and ${\hat{F_B}}_i$ and ${F^D_B}_i$ denote the value of the $i$th element in $\hat{F_B}$ and $F^D_B$, respectively. This formula calculates the mean of the squared differences between corresponding positions in the two feature maps, serving as the loss.

\subsubsection{GlobalAlign-v2 (Diffusion)}
Query-based multi-modal BEV representations are inherently implicit, rendering it infeasible to apply GlobalAlign-v2 (Deformable), as depicted in Figure~\ref{fig:GlobalAlign_diffusion}, for explicit global misalignment correction. To overcome this limitation, we introduce GlobalAlign-v2 (Diffusion), a diffusion-based framework specifically designed to address global misalignment in query-based BEV representations. Unlike traditional offset learning approaches with fixed noise injection, the proposed method leverages a progressive and learnable denoising process, enabling more effective modeling of the complex spatial misalignment patterns inherent to multi-modal fusion.
\textcolor{black}{Unlike conventional image-generation diffusion models, GlobalAlign-v2 (Diffusion) performs lightweight feature-level alignment on compact BEV representations. We adopt $T=4$ throughout all experiments to achieve a favorable trade-off between robustness and efficiency.}

\textcolor{black}{
It is worth noting that GlobalAlign-v2 (Diffusion) shares the same alignment objective as GlobalAlign-v2 (Deformable): both aim to estimate and compensate for global feature misalignment through spatial correction. The key difference lies in how the correction is performed. GlobalAlign-v2 (Deformable) adopts a one-shot alignment strategy that directly predicts spatial offsets from the current BEV representation, which is effective when misalignment can be explicitly observed in dense BEV features. However, query-based BEV representations encode spatial information implicitly within latent query embeddings, making global misalignment difficult to directly model using explicit offset prediction. Simply stacking multiple deformable alignment layers only increases network depth while still relying on deterministic offset estimation. In contrast, GlobalAlign-v2 (Diffusion) reformulates global alignment as a progressive denoising process. By gradually injecting and removing misalignment noise, the model performs multi-step offset refinement rather than a single offset prediction. Therefore, the proposed diffusion framework can be viewed as a generalized iterative alignment mechanism that extends deformable alignment to implicit query-based BEV representations.
}

As shown in Figure \ref{fig:GlobalAlign_diffusion}, we inject random offset noise into the camera BEV features to simulate global misalignment, denoted as:
\begin{align}
F^{MM}_{N,0} = F^{MM}_{B} + \epsilon_0, \quad \epsilon_0 \sim \mathcal{N}(0, \sigma_0^2),
\end{align}
where $F^{MM}_{B}$ is the clean fused BEV feature, and $\epsilon_0$ represents initial Gaussian noise with variance $\sigma_0^2$.

We then apply a forward diffusion process over $T$ discrete time steps, gradually adding noise according to a predefined schedule $\{\beta_t\}_{t=1}^T$, producing a series of noisy features $\{F^{MM}_{N,t}\}_{t=1}^T$. This process models the progressive corruption of BEV features by global misalignment noise.

The core of our strategy lies in training a neural network $\mathcal{D}_\theta$ to learn the reverse denoising process:
\begin{align}
\hat{F}^{MM}_{B} = \mathcal{D}_\theta(F^{MM}_{N,t}, t),
\end{align}
which estimates the clean fused BEV feature $\hat{F}^{MM}_{B}$ from a noisy input $F^{MM}_{N,t}$ at any diffusion step $t$. Here, $\mathcal{D}_\theta$ adopts a similar architecture to the original MM-Align module of GlobalAlign-v2 (Deformable) but is enhanced to condition on the diffusion timestep and progressively remove noise.

At each reverse step, the denoised feature $\hat{F}^{MM}_{B}$ is used to predict spatial offsets $F^{O}_t$, which in turn guide the grid sampling operation applied on LiDAR BEV features $F^{L}_{B}$ to obtain deformable aligned features:
\begin{align}
F^{D}_{W,t} = \text{GridSample}(F^{L}_{B}, F^{O}_t), \quad F^{D}_{B,t} = F^{D}_{W,t} \odot F^{L}_{B},
\end{align}
where $\odot$ denotes element-wise multiplication. This iterative correction allows the model to refine global alignment in a noise-adaptive manner.

We optimize the diffusion model parameters $\theta$ by minimizing a combination of reconstruction loss and denoising score matching loss across all diffusion steps:
\begin{align}
\mathcal{L}_{\text{diff}} = \mathbb{E}_{t, F^{MM}_{B}, \epsilon_t} \left\| F^{MM}_{B} - \mathcal{D}_\theta(F^{MM}_{N,t}, t) \right\|^2.
\end{align}

By integrating diffusion-based denoising into the global alignment module, our approach models complex and varying global misalignments in a progressive manner, leading to more accurate multi-modal fusion.

\textcolor{black}{
Conceptually, the proposed diffusion framework differs from a deeper or recurrent alignment network in that the denoising trajectory explicitly models the distribution of global misalignment noise. Rather than repeatedly applying the same alignment operator, the model learns to recover clean BEV representations from progressively corrupted states, enabling adaptive correction under different levels of global misalignment. Consequently, diffusion-based alignment provides a more robust solution for query-based BEV representations, where misalignment is implicitly encoded and difficult to correct using conventional deformable alignment alone.
}

\subsection{3D Object Detection (3DOD)}
To comprehensively validate the effectiveness of our method in addressing feature misalignment in 3D object detection, we design and evaluate our approach across multiple multi-modal fusion scenarios, including both \textbf{LiDAR-Camera} and \textbf{Radar-Camera} combinations.

\noindent \textbf{LiDAR-Camera Fusion.}
We consider two mainstream BEV construction paradigms:
\begin{itemize}
    \item \textbf{LSS-based}: We integrate our alignment modules into two representative LSS-based frameworks: \textit{UniTR}~\cite{wang2023unitr} and \textit{BEVFusion}~\cite{bevfusion-mit}.
    \item \textbf{Query-based}: We apply our method to \textit{BEVFormer-M}~\cite{BEVFormerpami}, which adopts implicit BEV queries for BEV feature construction.
\end{itemize}

\noindent \textbf{Radar-Camera Fusion.}
We further validate the generality of our approach by incorporating it into radar-camera fusion. Specifically, we adopt the \textbf{LSS-based} framework \textit{HVDetFusion}~\cite{lei2023hvdetfusion} as the baseline for evaluation.\textcolor{black}{
Unlike LiDAR, Radar measurements are sparse and contain limited elevation information. Nevertheless, LocalAlign-v2 relies on local geometric neighborhood consistency rather than dense depth estimation. Therefore, the same KD-Tree-based neighborhood construction can be directly applied to projected Radar observations, enabling robust alignment under Radar-Camera fusion.
}

\subsection{3D Occupancy Prediction (3DOP)}
\textcolor{black}{To further investigate the applicability of feature alignment beyond 3D object detection, we extend GraphBEV++ to the \textbf{3D occupancy estimation} task, where feature misalignment remains a critical challenge due to the reliance on multi-view geometric projection and feature aggregation.}

\textcolor{black}{We conduct experiments based on \textit{GaussianFormer-T}~\cite{huang2024gaussianformer}, a representative query-based occupancy framework. Since GaussianFormer adopts a BEVFormer-style query paradigm for occupancy prediction, we integrate our \textbf{GraphBEV++ (Query)} variant into the framework. This setting enables us to evaluate whether the proposed LocalAlign-v2 and GlobalAlign-v2 modules can effectively improve feature alignment quality and enhance occupancy perception under complex driving scenarios.}

\subsection{End-to-End Autonomous Driving (E2E-AD)}
Feature misalignment permeates the entirety of autonomous driving tasks, adversely affecting not only perception performance but also influencing downstream components such as motion prediction and planning. We are the first article dedicated to resolving misalignment issues within the end-to-end framework. To further demonstrate the scalability and generalizability of our approach, we extend GraphBEV++---originally designed for 3D object detection---into a unified end-to-end autonomous driving framework. This extension simultaneously enhances the performance of scene perception, motion forecasting, and ego-planning modules.

Our method is applicable to a wide range of BEV-based end-to-end driving systems, including \textit{UniAD}~\cite{uniad}, \textit{FusionAD}~\cite{fusionad}, \textit{VAD}~\cite{jiang2023vad}, \textit{MomAD}~\cite{momad}, and \textit{WoTE}~\cite{wote}. Specifically, for vision-only BEV frameworks such as \textit{UniAD}~\cite{uniad}, \textit{VAD}~\cite{jiang2023vad}, and \textit{MomAD}~\cite{momad}, we adopt the LocalAlign-v2 (Query) module to mitigate local misalignment. For multi-modal BEV systems like \textit{FusionAD}~\cite{fusionad} and \textit{WoTE}~\cite{wote}, we employ both LocalAlign-v2 (Query) and the proposed GlobalAlign-v2 (Diffusion) to address both local and global misalignments across modalities.

In summary, the alignment strategy introduced in GraphBEV++ exhibits strong extensibility and robustness, making it well-suited for a diverse set of perception and planning tasks within modern end-to-end autonomous driving pipelines.

% We extend the core concept of GraphBEV++ to multi-modal end-to-end autonomous driving, committing to its the advancement. %of multi-modal end-to-end autonomous driving. 

% 由于二者获取bev 特征方式的不同导致了LocalAlign-v2模块实现的差异，但是对齐的思路保持是一致的。前者的对齐模块如图2所示，方法具体细节和章节二保持一致

\section{Experiments}\label{sec:exps}

\subsection{Datasets}

\noindent \textbf{Bench2Drive  (E2E-AD).}
We conduct training and evaluation of GraphBEV++ on \textbf{Bench2Drive}~\cite{jia2024bench2drive}, a closed-loop evaluation protocol based on the CARLA Leaderboard 2.0~\cite{CARLA} for E2E-AD. It provides a base training set of 1000 clips, with 950 used for training and 50 for open-loop validation. Each clip captures approximately 150 meters of continuous driving in a specific traffic scenario. For closed-loop evaluation, we use the official 220 routes, covering 44 interactive scenarios with 5 routes each.

\noindent \textbf{NAVSIM  (E2E-AD).}
We conduct training and evaluation of GraphBEV++ on \textbf{NAVSIM}~\cite{dauner2024navsim} dataset. NAVSIM is a real-world, planning-oriented dataset that builds upon OpenScene~\cite{contributors2023openscene}, a compact redistribution of nuPlan~\cite{caesar2021nuplan}, the largest publicly available annotated driving dataset. It leverages eight cameras to achieve a full 360° field of view, along with a merged LiDAR point cloud derived from five sensors. Annotations are provided at 2 Hz and include both HD maps and object bounding boxes. The dataset is specifically designed to emphasize challenging driving scenarios involving dynamic changes in driving intentions, while deliberately excluding trivial cases such as stationary scenes or constant-speed cruising.

\noindent \textbf{NuScenes (E2E-AD\&3DOD\&3DOP).} 
We conduct extensive open-loop experiments on the \textbf{nuScenes} dataset~\cite{nuscenes}, which consists of 1000 driving scenes (700 for training, 150 for validation and 150 for test). Each scene lasts 20 seconds and includes around 40 key-frames annotated at 2 Hz. Each sample contains six images from surround-view cameras (covering 360° FOV), and point clouds from both LiDAR and radar sensors.

\noindent \textbf{NuScenes-C  (E2E-AD\&3DOD).} 
\textbf{NuScenes-C}~\cite{zhujun_benchmarking} is a corrupted benchmark derived from the nuScenes validation set, introducing various types of noise to assess the robustness of planning models. It includes 27 corruption types applied at 5 severity levels. To evaluate robustness under adverse weather conditions, we select three representative weather corruptions — Rain, Snow, and Fog — as our test scenarios.  In addition, to validate the robustness of \textbf{feature alignment}, we have followed Ref.~\cite{zhujun_benchmarking} to simulate misalignment caused by LiDAR and camera projection errors. It is worth noting that Ref.~\cite{zhujun_benchmarking} only adds noise to the validation dataset, rather than the train and test datasets. For the 3D detection task, we use noise severity levels from 1 to 5 and report the mean values. In the end-to-end autonomous driving tasks, we use noise severity levels from 1 to 10, with mean values reported.

\noindent \textbf{Argoverse 2 (3DOD).} We further conduct long-range experiments on the recently released Argoverse 2 (AV2) dataset~\cite{Argoverse2} to demonstrate the superiority of our GraphBEV++ in long-range detection. AV2 has a large-scale data, and it contains 1000 sequences in total, 700 for training, 150 for validation, and 150 for testing. In addition to average precision (AP), AV2 adopts a composite score as an evaluation metric, which takes both AP and localization errors into account. The perception range in AV2 is 200 meters (cover area of 400m × 400m). We test on the AV2 dataset due to its extensive range of long-distance scenarios, where the issue of feature misalignment intensifies with increasing distance.

\noindent \textbf{Waymo and Waymo-C (3DOD).} Waymo Open dataset (WOD)~\cite{waymo} is a well-known benchmark for large-scale outdoor 3D perception, comprising 1150 scenes which are divided into 798 scenes for training, 202 scenes for validation, and 150 scenes for testing. Each scene includes about 200 frames, covering a perception range of 150m $\times$ 150m. Similar to nuScenes-C, Waymo-C~\cite{zhujun_benchmarking} is constructed by applying all 27 corruptions to the Waymo validation set with 5 severities.

\subsection{Evaluation Metrics}
\noindent \textbf{Bench2Drive  (E2E-AD).} The \textbf{Bench2Drive}~\cite{jia2024bench2drive} includes five metrics for closed-loop evaluation: Driving Score (DS), Success Rate (SR), Efficiency, Comfortness, and Multi-Ability. The Success Rate quantifies the proportion of routes successfully completed within the allotted time. The Driving Score follows CARLA [11], incorporating both route completion status and violation penalties, where infractions reduce the score via discount factors. Efficiency and Comfortness are used to measure the speed performance and comfort of the autonomous driving system during the driving process, respectively. Multi-Ability measures 5 advanced skills, including `Merging, Overtaking, Emergency Brake, Give Way, and Traffic Sign', independently for urban driving. 

\noindent \textbf{NAVSIM  (E2E-AD).}
NAVSIM~\cite{dauner2024navsim} benchmarks planning performance using nonreactive simulations and closed-loop metrics for comprehensive evaluation. In this study, we employ the proposed PDM score (PDMS)~\cite{dauner2024navsim}, which is a weighted combination of several sub-scores: no at-fault collisions (NC), drivable area compliance (DAC), time-to-collision (TTC), comfort (Comf.), and ego progress (EP).

\noindent \textbf{NuScenes and NuScenes-C (3DOD).} 
For object detection on the nuScenes~\cite{nuscenes} and NuScenes-C~\cite{zhujun_benchmarking} dataset, evaluation metrics include mAP and the nuScenes detection score (NDS). mAP is calculated by averaging over the distance thresholds of 0.5m, 1m, 2m, and 4m across all categories. NDS is a weighted average of mAP and five other true positive metrics that measure translation, scaling, orientation, velocity, and attribute errors.

\noindent \textbf{NuScenes and NuScenes-C (E2E-AD).} 
For the metrics of perception tasks, we use mAP and NDS to evaluate the detection tasks, adopt Average Multi-object Tracking Accuracy (AMOTA) and Average Multi-object Tracking Precision (AMOTP) to evaluate the tracking tasks, use intersection-over-union (IoU) to evaluate the mapping tasks. To evaluate the prediction and planning tasks, we adopt conventional metrics, including End-to-end Prediction Accuracy (EPA), Average Displacement Error (ADE), Final Displacement Error (FDE), and Miss Rate (MR) to evaluate the performance of motion prediction. For future occupancy prediction, we use the metrics Future Video Panoptic Quality (VPQ) and IoU for near (30 × 30m) and far (100 × 100m) range, following FIERY~\cite{hu2021fiery}.
For planning evaluation, we adopt the commonly used L2 displacement error (L2) and collision rate as the primary metrics. 
% However, the evaluation protocols differ between UniAD~\cite{uniad} and VAD~\cite{jiang2023vad}. We clarify the specific metric settings in the corresponding experimental tables.

\noindent \textbf{Argoverse 2 (3DOD).}  For object detection on the Argoverse2~\cite{Argoverse2} dataset, mAP is adopted as the evaluation metric. We test on the AV2 dataset due to its extensive range of long-distance scenarios, where the issue of feature misalignment intensifies with increasing distance.

\noindent \textbf{Waymo and Waymo-C (3DOD).}  For evaluation metrics, WOD~\cite{waymo} employs 3D mean Average Precision (mAP) and mAP weighted by heading accuracy (mAPH). Each object is divided into two difficulty levels: L1 is for objects detected with more than five points and L2 is for those at least one point. For Waymo-C, the official evaluation metrics are mAP and mAPH by taking the heading accuracy into consideration. We similarly calculate the corruption robustness and relative corruption error on Waymo-C.

\noindent \textbf{NuScenes (3D  Occupancy Prediction).}  \textcolor{black}{For 3D semantic occupancy prediction, we use the intersection-over-union (IoU) of occupied voxels, regardless of their semantic categories, as the evaluation metric for the scene completion (SC) task. For the semantic scene completion (SSC) task, we adopt the mean IoU (mIoU) over all semantic classes:}

\begin{equation}
\mathrm{IoU} =
\frac{TP}
{TP + FP + FN},
\end{equation}

\begin{equation}
\mathrm{mIoU} =
\frac{1}{C}
\sum_{i=1}^{C}
\frac{TP_i}
{TP_i + FP_i + FN_i},
\end{equation}

where $TP$, $FP$, and $FN$ denote the numbers of true positives, false positives, and false negatives, respectively, and $C$ is the total number of semantic classes.

\subsection{Implementation Details}

We implement GraphBEV++ within the PyTorch~\cite{paszke2019pytorch}, built upon the open-source BEVFusion~\cite{bevfusion-mit} and OpenPCDet~\cite{openpcdet}. For the LiDAR branch, feature encoding is performed using SECOND~\cite{Second} to obtain LiDAR BEV features, with voxel dimensions set to [0.075m, 0.075m, 0.2m] and point cloud ranges specified as [-54m, -54m, -5m, 54m, 54m, 3m] across the X, Y, and Z axes, respectively. The camera branch employs a Swin Transformer~\cite{swimtransformer} as the backbone, integrating Heads of numbers 3, 6, 12, 24, and utilizing FPN~\cite{maskrcnn} to fuse multi-scale feature maps. The resolution of input images is adjusted and cropped to 256 $\times$ 704. In the LSS~\cite{lss} configuration, frustum ranges are set with X coordinates [-54m, 54m, 0.3m], Y coordinates [-54m, 54m, 0.3m], Z coordinates [-10m, 10m, 20m], and depth range [1m, 60m, 0.5m].
% jfy building upon the open-source BEVFusion -> built

\textcolor{black}{
For occupancy prediction, the spatial range is set to
$[-50\,\mathrm{m}, 50\,\mathrm{m}]$ along the X- and Y-axes and
$[-5\,\mathrm{m}, 3\,\mathrm{m}]$ along the Z-axis. The final occupancy volume has a resolution of $200 \times 200 \times 16$ with a voxel size of $0.5\,\mathrm{m}$. The input image resolution is $1600 \times 900$.
The network adopts a hierarchical architecture with $M=4$ levels, where skip connections are not applied at Level~0. For the nuScenes dataset, we employ ResNet101-DCN~\cite{resnet,dcn} initialized with FCOS3D~\cite{wang2021fcos3d} pretrained weights as the image backbone. The features from Stages 1--3 are processed by an FPN~\cite{fpn} to generate multi-scale image features. The numbers of 2D--3D spatial attention layers are set to 1, 3, and 6 for the three levels, respectively.}

During training, we employ data augmentation for ten epochs, including random flips, rotations (within the range [$-\frac{\pi}{4}$, $\frac{\pi}{4}$]), translations (with std=0.5), and scaling in [0.9, 1.1] for LiDAR data enhancement. We use CBGS~\cite{CBGS} to resample the training data. Additionally, we use random rotation in [$-5.4^\circ$, $5.4^\circ$] and random resizing in [0.38, 0.55] to augment the images. The Adam optimizer~\cite{adam} is used with a one-cycle learning rate policy, setting the maximum learning rate to 0.001 and weight decay to 0.01. The batch size is 24, and our training is conducted on 8 NVIDIA GeForce RTX 3090 24G GPUs. During inference, we remove Test Time Augmentation (TTA) data augmentation, and the batch size is set to 1 on an A100 GPU. All latency measurements are taken on the same workstation with an A100 GPU.

\subsection{Comparisons with State-of-the-Art Methods}

\begin{table*}[t]
\centering
  \caption[]{Comparison with the SOTA methods on the nuScenes  \textcolor{blue}{validation} and \textcolor{red}{test}  set. `C.V.', `Motor.', `Ped.' and `T.C.' are short for construction vehicles, motorcycles, pedestrians, and traffic cones. The Modality column: ‘L’ denotes LiDAR-only input, while ‘LC’ indicates the use of both LiDAR and camera data. $^{\dagger}$ means using TTA (test-time augmentation). * denotes the re-implementation. The best results are highlighted in \textbf{bold}, and the second-best results are indicated with \underline{underlining}.
 }
  % }
  \renewcommand\arraystretch{1}
  \tabcolsep=2.4mm %%%%%%%%%
  \resizebox{\linewidth}{!}{
  %\begin{tabular*}{\linewidth} {@{}@{\extracolsep{\fill}}!{\color{white}\vline}l|c|c|c|c|c|c|c|c|c|c|c|c @{}}
\begin{tabular}{l|c|c|cc|ccccc ccccc}
\toprule
Method & Venue & Modality & mAP & NDS & Car & Truck & C.V. & Bus & Trailer & Barrier & Motor. & Bike & Ped. & T.C. \\
\midrule
\multicolumn{14}{c}{Performances on \textcolor{blue}{validation} set} \\
\midrule
SparseBEV\cite{chen2026sparsebev} & \textcolor{black}{TPAMI'26} & C & 43.2 & 54.5&-&-&-&-&-&-&-&-&-&-  \\
Refine3D\cite{li2026refine3d} & \textcolor{black}{AAAI'26} & C & 53.9 & 62.4&-&-&-&-&-&-&-&-&-&-   \\
DGFSD\cite{zhang2025dgfsd} & \textcolor{black}{MM'25} & L & 66.8 & 71.3 & 87.9 & 63.4 & 26.4 & \underline{77.9} & \underline{47.2} & 69.8 & 75.8 & 56.1 & 88.1 & 75.7 \\
TransFusion-L~\cite{Transfusion} & CVPR'22 & L & 65.1 & 70.1 & 86.5 & 59.6 & 25.4 & 74.4 & 42.2 & 74.1 & 72.1 & 56.0 & 86.6 & 74.1 \\
FSDv2\cite{fan2024fsd} & \textcolor{black}{TPAMI'24} & L & 64.7 & 70.4&-&-&-&-&-&-&-&-&-&- \\
DepthFusion\cite{ji2026depthfusion} & \textcolor{black}{TMM'26} & LC & \underline{71.2} & \textbf{74.0}&-&-&-&-&-&-&-&-&-&- \\
HD-Fusion\cite{jing2026hd} & \textcolor{black}{TII'26} & LC & 70.5 & 72.9 & 88.8 & 60.5 & \textbf{35.4} & \textbf{80.7} & \textbf{48.1} & 68.1 & 79.8 & \textbf{69.7} & \underline{90.3} & \underline{83.3} \\
ObjectFusion~\cite{ObjectFusion} & ICCV'23 & LC & 69.8 & 72.3 & 89.7 & 65.6 & 31.8 & 77.7 & 42.8 & 75.2 & 79.4 & 65.0 & 89.3 & 81.1 \\
GraphBEV~\cite{song2024graphbev} & \textcolor{black}{ECCV'24} & LC & 70.1 & 72.9 & 89.9 & 64.7 & 31.1 & 76.0 & 43.8 & 76.0 & 80.1 & 67.5 & 89.2 & 82.2 \\
BEVFormer-M~\cite{BEVFormerpami} & \textcolor{black}{TPAMI'25} & LC & \underline{71.2} & 73.2&-&-&-&-&-&-&-&-&-&- \\
BEVFormer-M*~\cite{BEVFormerpami} & \textcolor{black}{TPAMI'25} & LC & 70.9 & 73.0 & 90.4 & \underline{65.7} & 32.1 & 77.1 & 45.0 & \underline{77.5} & \underline{81.2} & 68.0 & 89.7 & 82.2 \\
\midrule
\textbf{GraphBEV++ (LSS)} & - & LC & 70.7 & 73.2 & \underline{90.5} & 65.3 & 31.8 & 76.4 & 44.4 & 76.6 & 80.8 & 68.1 & 89.7 & \textbf{83.4} \\
\textbf{GraphBEV++ (Query)} & - & LC & \textbf{71.4} & \underline{73.4} & \textbf{91.6} & \textbf{66.2} & \underline{32.3} & 77.4 & 45.5 & \textbf{77.8} & \textbf{81.9} & \underline{68.4} & \textbf{90.4} & 82.5 \\
\midrule
\midrule
\multicolumn{14}{c}{Performances on \textcolor{red}{test} set} \\
\midrule
OcRFDet\cite{ji2025ocrfdet} & \textcolor{black}{ICCV'25} & C & 57.2 & 64.8 & - & - & -& -& -& -& -& -& -& -\\
PointPillars~\cite{Pointpillars} & CVPR'19 & L & 40.1 & 55.0 & 76.0 & 31.0 & 11.3 & 32.1 & 36.6 & 56.4 & 34.2 & 14.0 & 64.0 & 45.6 \\
FSDv2\cite{fan2024fsd} & \textcolor{black}{TPAMI'24} & L &
66.2 & 71.7 &
83.7 & 51.6 &
\textbf{66.4} &
59.1 & 32.5 &
87.1 & 71.4 & 51.7 & 80.3 & 78.7 \\

CenterPoint~\cite{Centerpoint}$^{\dagger}$ & CVPR'21 & L &
60.3 & 67.3 &
85.2 & 53.5 & 20.0 & 63.6 & 56.0 &
71.1 & 59.5 & 30.7 & 84.6 & 78.4 \\

PointPainting~\cite{pointpainting} & CVPR'20 & LC &
46.4 & 58.1 &
77.9 & 35.8 & 15.8 & 36.2 & 37.3 &
60.2 & 41.5 & 24.1 & 73.3 & 62.4 \\

PointAugmenting~\cite{wang2021pointaugmenting}$^{\dagger}$ & CVPR'21 & LC &
66.8 & 71.0 &
87.5 & 57.3 & 28.0 & 65.2 & 60.7 &
72.6 & 74.3 & 50.9 & 87.9 & 83.6 \\

MVP~\cite{mvp} & NeurIPS'21 & LC &
66.4 & 70.5 &
86.8 & 58.5 & 26.1 & 67.4 & 57.3 &
74.8 & 70.0 & 49.3 & 89.1 & 85.0 \\

GraphAlign~\cite{graphalign} & ICCV'23 & LC &
66.5 & 70.6 &
87.6 & 57.7 & 26.1 & 66.2 & 57.8 &
74.1 & 72.5 & 49.0 & 87.2 & 86.3 \\

AutoAlignV2~\cite{autoalignv2} & ECCV'22 & LC &
68.4 & 72.4 &
87.0 & 59.0 & 33.1 & 69.3 & 59.3 &
- & 72.9 & 52.1 & 87.6 & - \\

TransFusion~\cite{Transfusion} & CVPR'22 & LC &
68.9 & 71.7 &
87.1 & 60.0 & 33.1 & 68.3 & 60.8 &
78.1 & 73.6 & 52.9 & 88.4 & 86.7 \\

DeepInteraction~\cite{DeepInteraction} & NeurIPS'22 & LC &
70.8 & 73.4 &
87.9 & 60.2 & 37.5 & 70.8 & 63.8 &
\underline{80.4} &
75.4 & 54.5 & 90.3 & 87.0 \\

BEVFusion-PKU~\cite{bevfusion-pku} & NeurIPS'22 & LC &
69.2 & 71.8 &
88.1 & 60.9 & 34.4 & 69.3 & 62.1 &
78.2 & 72.2 & 52.2 & 89.2 & 85.2 \\

ObjectFusion~\cite{ObjectFusion} & ICCV'23 & LC &
71.0 & 73.3 &
\underline{89.4} &
59.0 &
\underline{40.5} &
71.8 & 63.1 &
76.6 & 78.1 & 53.2 & 90.7 & 87.7 \\

MV2DFusion\cite{wang2025mv2dfusion} & \textcolor{black}{TPAMI'25} & LC &
\textbf{74.5} &
\textbf{76.7}&-&-&-&-&-&-&-&-&-&- \\

MSMDFusion~\cite{MSMDFusion} & CVPR'23 & LC &
71.5 &
\underline{74.0} &
88.4 &
\underline{61.0} &
35.2 &
71.4 &
64.2 &
\textbf{80.7} &
76.9 &
58.3 &
90.6 &
88.1 \\

SparseFusion~\cite{SparseFusion} & ICCV'23 & LC &
\underline{72.0} &
73.8 &
88.0 &
60.2 &
38.7 &
72.0 &
\underline{64.9} &
79.2 &
\underline{78.5} &
\underline{59.8} &
\underline{90.9} &
87.9 \\

CMT~\cite{cmt} & ICCV'23 & LC &
\underline{72.0} &
73.8 &
88.0 &
\textbf{63.3} &
37.3 &
\textbf{75.4} &
\textbf{65.4} &
78.2 &
\textbf{79.1} &
\textbf{60.6} &
87.9 &
84.7 \\

BEVFusion-MIT~\cite{bevfusion-mit} & ICRA'23 & LC &
70.2 & 72.9 &
88.6 & 60.1 & 39.3 & 69.8 & 63.8 &
80.0 & 74.1 & 51.0 & 89.2 & 86.5 \\

DepthFusion\cite{ji2026depthfusion} & \textcolor{black}{TMM'26} & LC &
70.9 &
\underline{74.2} &-&-&-&-&-&-&-&-&-&-\\

GraphBEV~\cite{song2024graphbev} & \textcolor{black}{ECCV'24} & LC &
71.7 &
73.6 &
89.2 &
60.0 &
\underline{40.8} &
72.1 &
64.5 &
80.1 &
76.8 &
53.3 &
\underline{90.9} &
\underline{88.9} \\

\midrule
\rowcolor{pink!10}
\textbf{GraphBEV++ (LSS)} & - & LC &
\underline{72.0} &
74.0 &
\textbf{89.5} &
60.5 &
\underline{41.3} &
\underline{72.4} &
\underline{64.9} &
80.3 &
77.1 &
53.6 &
\textbf{91.1} &
\textbf{89.3} \\
\bottomrule
\end{tabular} }
\label{tab_nuscenes_val_test}
\end{table*}

\begin{table}[t]
% \scriptsize
\centering
  \caption[]{Comparison with SOTAs on the \textbf{nuScenes (Clean)} and \textbf{nuScenes-C (Noisy)} dataset. \textcolor{darkgreen}{Green} denotes the relative performance drop from the clean to the noisy settings.   All latency measurements are conducted on the same workstation with an A100 GPU.
  }
  \renewcommand\arraystretch{1}
  \tabcolsep=2.8mm %%%%%%%%%
  \resizebox{\linewidth}{!}{
  \begin{tabular}{l|c|cc }
    \toprule
 Method  & Setting          & mAP  & NDS   \\ 
\midrule
\multirow{2}{*}{SparseFusion~\cite{SparseFusion}}& Clean&70.4& 72.8    \\
&\cellcolor{darkgreen!5}Noisy&  \cellcolor{darkgreen!5}64.7\textit{\fontsize{6}{0}\selectfont\textcolor{darkgreen}{-8.1\%}} & \cellcolor{darkgreen!5}67.1\textit{\fontsize{6}{0}\selectfont\textcolor{darkgreen}{-7.8\%}}\\
\midrule

\multirow{2}{*}{BEVFusion-MIT~\cite{bevfusion-mit}}& Clean& 68.5 & 71.4   \\
&\cellcolor{darkgreen!5}Noisy&  \cellcolor{darkgreen!5}60.8\textit{\fontsize{6}{0}\selectfont\textcolor{darkgreen}{-11.2\%}} & \cellcolor{darkgreen!5}65.7\textit{\fontsize{6}{0}\selectfont\textcolor{darkgreen}{-8.2\%}}\\

\midrule
\multirow{2}{*}{BEVFormer-M~\cite{BEVFormerpami}}& Clean&70.9& 73.0   \\
&\cellcolor{darkgreen!5}Noisy&  \cellcolor{darkgreen!5}63.2\textit{\fontsize{6}{0}\selectfont\textcolor{darkgreen}{-10.8\%}} & \cellcolor{darkgreen!5}66.3\textit{\fontsize{6}{0}\selectfont\textcolor{darkgreen}{-9.2\%}}\\
\midrule

\multirow{2}{*}{GraphBEV~\cite{song2024graphbev}}& Clean& 70.1& 72.9   \\
&\cellcolor{darkgreen!5}Noisy&  \cellcolor{darkgreen!5}69.1\textit{\fontsize{6}{0}\selectfont\textcolor{darkgreen}{-1.4\%}} & \cellcolor{darkgreen!5}72.0\textit{\fontsize{6}{0}\selectfont\textcolor{darkgreen}{-1.2\%}}\\
\midrule
\rowcolor{pink!10}\multirow{2}{*}{GraphBEV++ (LSS)~\cite{bevfusion-mit}}& Clean& 70.7& 73.2   \\
&\cellcolor{darkgreen!5}Noisy&  \cellcolor{darkgreen!5}69.3\textit{\fontsize{6}{0}\selectfont\textcolor{darkgreen}{-2.0\%}} &  \cellcolor{darkgreen!5}72.3\textit{\fontsize{6}{0}\selectfont\textcolor{darkgreen}{-1.2\%}}\\
\midrule
\rowcolor{pink!10}\multirow{2}{*}{GraphBEV++ (Query)~\cite{bevfusion-mit}}& Clean& 71.4 & 73.4   \\
&\cellcolor{darkgreen!5}Noisy&  \cellcolor{darkgreen!5}69.1\textit{\fontsize{6}{0}\selectfont\textcolor{darkgreen}{-3.2\%}} & \cellcolor{darkgreen!5}71.2\textit{\fontsize{6}{0}\selectfont\textcolor{darkgreen}{-3.0\%}}\\
\bottomrule
\end{tabular} }
\label{tab_nuscenes_C_misalignment}
\end{table}

\begin{table}[htp]
\caption[]{Comparison with SOTA methods on the nuScenes~\cite{nuscenes} validation set under \textcolor{blue}{clean} and \textcolor{red}{noise} misalignment settings. ‘RC’ indicates the use of both Radar and camera data. }
  \renewcommand\arraystretch{1}
  \tabcolsep=2.5mm %%%%%%%%%
  \resizebox{\linewidth}{!}{

  \begin{tabular}{l|c|c|cc}
    \toprule
   Methods &Venue& Modality  &  mAP (\textcolor{blue}{clean})& mAP (\textcolor{red}{noise}) \\ 
   \midrule
StreamPETR\cite{wang2023exploring}&ICCV'23&C&45.0& 40.8\\
RayFormer\cite{chu2024rayformer}&\textcolor{black}{MM'24}&C&45.9& 41.4\\
RCBEVDet\cite{lin2024rcbevdet}&\textcolor{black}{CVPR'24}&RC& 45.3& 41.2\\
CRN\cite{kim2023crn}&ICCV'23&RC& 49.0& 43.8\\
HyDRa\cite{wolters2025unleashing}&\textcolor{black}{ICRA'25}&RC& 49.4& 44.1\\
SparseInteraction\cite{xu2024sparseinteraction}&\textcolor{black}{MM’24}&RC&45.8&-\\
\midrule
HVDetFusion~\cite{lei2023hvdetfusion}&ArXiv'23&RC& 45.1& 40.1\\
\rowcolor{pink!10}\textbf{GraphBEV++ (LSS)}&-&RC& 45.9\textit{\fontsize{6}{0}\selectfont\textcolor{black}{+0.8}}& 45.0\textit{\fontsize{6}{0}\selectfont\textcolor{red}{+4.9}}\\
\midrule
RaCFormer \cite{chu2025racformer}&\textcolor{black}{CVPR'25}&RC& 54.1& 50.6\\
\rowcolor{pink!10}\textbf{GraphBEV++ (LSS)}&-&RC& 54.8\textit{\fontsize{6}{0}\selectfont\textcolor{black}{+0.7}}& 54.2\textit{\fontsize{6}{0}\selectfont\textcolor{red}{+3.6}}\\
\bottomrule
\end{tabular} }
\label{tab_nus_C_noise_radar}
\end{table}

\begin{table*}[htp]

\caption[]{Comparison with UniTR (LSS)~\cite{wang2023unitr} on vehicle results under \textbf{Waymo} (clean) and \textbf{Waymo-C} (noise misalignment) settings (10\% Waymo Training Data).}

  \renewcommand\arraystretch{1}
  \tabcolsep=7.5mm %%%%%%%%%
  \resizebox{\linewidth}{!}{
  %\begin{tabular*}{\linewidth} {@{}@{\extracolsep{\fill}}!{\color{white}\vline}l|c|c|c|c|c|c|c|c|c|c|c|c @{}}
  \begin{tabular}{l|c|cc|cc}
    \toprule
   \multirow{2}{*}{Methods}& \multirow{2}{*}{Modality}     & \multicolumn{2}{c|}{mAP/mAPH \textcolor{black}{(clean)}}   & \multicolumn{2}{c}{mAP/mAPH \textcolor{red}{(noise)}} \\ 
   && L1& L2&L1&L2
   \\
\midrule

UniTR (LSS)~\cite{wang2023unitr}& LC   &44.86/40.41 & 37.58/33.69& 38.88/34.24 & 32.44/29.95  \\ 
\rowcolor{pink!10}\textbf{GraphBEV++ (LSS)}& LC  &45.73/41.17 & 38.51/35.14& 43.50/38.97 & 36.48/33.08  \\
&
&\textit{\fontsize{6}{0}\selectfont\textcolor{black}{+0.87\textcolor{black}{/}+0.76}}
&\textit{\fontsize{6}{0}\selectfont\textcolor{black}{+0.93\textcolor{black}{/}+1.45}}
&\textit{\fontsize{6}{0}\selectfont\textcolor{red}{+4.62\textcolor{black}{/}+4.73}}
&\textit{\fontsize{6}{0}\selectfont\textcolor{red}{+4.04\textcolor{black}{/}+3.13}}
\\
\bottomrule
\end{tabular} }
\label{tab_waymo_C_noise}
\end{table*}

\begin{table*}[t]
\scriptsize
\centering
  \caption[]{Comparison with prior methods on \textbf{Argoverse2} validation set. Metrics: mAP (\%)↑ for the overall results, AP (\%)↑ for each category. * denotes result from paper~\cite{chen2023voxelnext}. $^{\dagger}$ denotes result re-implentment.}
  \renewcommand\arraystretch{1}
  \tabcolsep=0.1mm %%%%%%%%%
  \resizebox{\linewidth}{!}{
  %\begin{tabular*}{\linewidth} {@{}@{\extracolsep{\fill}}!{\color{white}\vline}l|c|c|c|c|c|c|c|c|c|c|c|c @{}}
  \begin{tabular}{l|c|c|ccccc ccccc ccccc ccccc }
    \toprule
Method  &Venue    &mAP& Veh.& Bus& Ped.& Stop.& Box.& Boll.& C-B.& M.-list& MPC.& M.-cycle& Bicycle& A-B.& School.& Truck.& C-C.& V-T.& Sign& Large.& Str.& Bic.-list \\ 
\midrule
PETR\cite{liu2022petr}&ECCV'22&17.6&-&-&-&-&-&-&-&-&-&-&-&-&-&-&-&-&-&-&-&-\\
Sparse4Dv2\cite{lin2023sparse4d}&Aexiv'23&18.9&-&-&-&-&-&-&-&-&-&-&-&-&-&-&-&-&-&-&-&-\\
StreamPETR\cite{wang2023exploring}&ICCV'23&20.3&-&-&-&-&-&-&-&-&-&-&-&-&-&-&-&-&-&-&-&-\\
Far3D\cite{jiang2024far3d}&\textcolor{black}{AAAI'24}&24.4&-&-&-&-&-&-&-&-&-&-&-&-&-&-&-&-&-&-&-&-\\
CenterPoint*~\cite{Centerpoint}&CVPR'21&22.0& 67.6& 38.9& 46.5& 16.9& 37.4& 40.1& 32.2& 28.6& 27.4& 33.4& 24.5& 8.7& 25.8& 22.6& 29.5& 22.4& 6.3& 3.9& 0.5& 20.1\\
FSD*~\cite{fsd}&NeurPS'22& 28.2& 68.1& 40.9 &59.0& 29.0& 38.5& 41.8& 42.6& 39.7& 26.2& 49.0& 38.6& 20.4& 30.5& 14.8& 41.2& 26.9& 11.9& 5.9 &13.8 &33.4\\
FSDv2~\cite{fsd}&\textcolor{black}{TPAMI'24}& 37.6&  77.0 & 47.6&  70.5&  43.6 & 41.5&  53.9 & 58.5 & 56.8&  39.0 & 60.7 & 49.4 & 28.4 & 41.9 & 30.2&  44.9&  33.4&  16.6&  7.3&  32.5&  45.9\\
VoxelNeXt*~\cite{chen2023voxelnext}&CVPR'23& 30.0& 71.7& 39.2& 63.1& 39.2& 40.0& 52.5& 63.7& 42.2& 34.9& 42.7& 40.1& 20.1& 25.2& 16.9& 45.7& 22.3& 15.8& 5.9& 9.8& 33.5\\
HEDNet\cite{zhang2023hednet} &NeurPS'23& 37.1 &78.2 &47.7 &67.6 &46.4 &45.9 &56.9 &67.0 &48.7 &46.5 &58.2 &47.5& 23.3 &40.9 &21.6& 46.8& 27.9& 20.6& 6.9& 27.2& 38.7\\
SAFDNet\cite{zhang2024safdnet}&\textcolor{black}{CVPR'24}& 39.7 &78.5& 49.4 &70.7 &51.5 &44.7 &65.7& 72.3 &54.3& 49.7 &60.8 &50.0& 31.3 &44.9& 24.7& 55.4& 31.4 &22.1& 7.1 &31.1 &42.7\\
LION-Mamba\cite{liu2024lion}&\textcolor{black}{NeurPS'24}&  41.5& 75.1& 43.6& 73.9& 53.9& 45.1 &66.4 &74.7 &61.3 &48.7 &65.1 &56.2 &21.7 &42.7& 25.3 &58.4 &28.9& 23.6 &8.3 &49.5& 47.3\\
UniMamba\cite{jin2025unimamba} &\textcolor{black}{CVPR'25}& 42.0& 78.9& 47.9& 74.3 &51.8& 46.8& 67.8 &76.9& 55.8 &51.7& 62.8 &52.4& 30.2 &44.6 &24.6 & 59.4 &32.2 &23.2& 6.7 &41.5& 48.5\\
GeoFormer\cite{jin2025geoformer}&\textcolor{black}{ICCV'25}&41.7&77.4 &50.7 &73.7&-&-&-&-&-&-&-&-&-&-&-&-&-&-&-&-&-\\
FSHNet\cite{liu2025fshnet}&\textcolor{black}{CVPR'25}&40.2& 75.1& 47.1 &50.3 &76.2 &54.9 &46.0 &62.1 &28.5& 45.8& 29.1& 25.4 &64.8 &64.1& 48.9 &44.5& 61.3 &26.0 &44.1 &23.6 &32.5\\
M3Net\cite{yuan2023m}&Arxiv'23&40.9 &74.9& 47.8& 57.4& 77.1& 55.3 &48.5& 63.9& 29.9& 47.5& 28.2& 25.4& 67.0& 64.2 &46.8& 43.8& 59.5& 24.0& 41.5 &21.8& 30.5\\
PGDC\cite{li2026look}&\textcolor{black}{CVPR'26}&41.7& 76.2& 49.0& 58.5& 78.1& 56.8 &49.8 &65.1& 31.2& 48.5 &31.0& 26.9 &68.1 &65.8 &50.3& 46.0& 62.5& 27.5& 45.8& 25.0 &33.8\\
BEVFusion$^{\dagger}$~\cite{bevfusion-mit}&ICRA'23& 43.1&83.1& 48.1& 64.2& 48.2& 54.5& 58.1& 65.3& 45.1& 39.8& 49.3& 48.1& 33.1& 38.1& 28.5& 54.1& 37.4& 34.6& 31.2& 34.4& 38.1\\

GraphBEV~\cite{song2024graphbev}&\textcolor{black}{ECCV'24}& 46.1& 85.2& 51.3& 67.3& 52.5& 58.7& 60.9& 67.4& 49.9& 40.1& 52.9& 52.4& 35.8& 42.9& 31.5& 57.8 &41.5 &36.7& 35.9& 37.6& 42.6\\
\rowcolor{pink!10}\textbf{GraphBEV++ (LSS)}&-& \textbf{46.7}&  \textbf{85.9}& \textbf{51.7}&\textbf{67.8}&\textbf{53.0}&\textbf{59.1}&\textbf{61.3}&\textbf{67.8}&\textbf{50.3}&\textbf{40.6}&\textbf{53.3}&\textbf{52.8}&\textbf{36.3}&\textbf{43.4}&\textbf{32.0}&\textbf{58.2}&\textbf{41.9}&\textbf{37.2}&\textbf{36.4}&\textbf{38.1}&\textbf{43.0}\\

\bottomrule
\end{tabular} }
\label{tab_Argoverse2_val}
\end{table*}

\begin{table*}[htp]
\scriptsize
\centering
  \caption[]{Comparisons with the SOTA methods on BEV map segmentation on \textbf{nuScenes} \textcolor{black}{validation} set. The Modality column: ‘L’ denotes LiDAR-only input, while ‘LC’ indicates the use of both LiDAR and camera data.}
  \renewcommand\arraystretch{1}
  \tabcolsep=3.5mm %%%%%%%%%
  \resizebox{\linewidth}{!}{
  %\begin{tabular*}{\linewidth} {@{}@{\extracolsep{\fill}}!{\color{white}\vline}l|c|c|c|c|c|c|c|c|c|c|c|c @{}}
  \begin{tabular}{l|c|ccc cccc }
    \toprule
Method      &Venue & Drivable & Ped. Cross. & Walkway & Stop Line & Carpark & Divider & Mean \\ 
\midrule
LSS\cite{lss} &ECCV'20& 	75.4	& 38.8& 	46.3& 	30.3& 	39.1	& 36.5& 	44.4\\
CVT\cite{cvt} & CVPR'22& 	74.3	&36.8	&39.9&	25.8&	35.0	&29.4&	40.2\\
M2BEV\cite{m2bev} & ArXiv'22& 	77.2&	-&	-&	-	&-&	40.5&	-\\
MapPrior\cite{zhu2023mapprior}&ICCV'23& 81.7	&54.6&	58.3&	46.7&	53.3	&45.1&	56.7\\
X-Align\cite{borse2023x} & 	WACV'23&	82.4&	55.6&	59.3&	49.6	&53.8&	47.4&	58.0\\
MetaBEV\cite{ge2023metabev} & 	ICCV'23& 	83.3&	56.7	&61.4&	50.8&	55.5&	48.0&	59.3\\
DDP\cite{ji2023ddp} & 	ICCV'23&	83.6&	58.3	&61.6&	52.4&	51.4&	49.2&	59.4\\
RGC\cite{chen2024residual} & 	\textcolor{black}{WACV'24}&	81.7&	57.1	&60.5	&51.7&	53.8&	53.5	&59.7\\
BridgeTA\cite{kim2025bridgeta}& 	\textcolor{black}{ArXiv'25}&	83.3&	58.6&	62.9&	53.6&	56.6&	50.1&	60.8\\
PointPillars~\cite{Pointpillars} & CVPR'19 & 72.0 & 43.1 & 53.1 & 29.7 & 27.7 & 37.5 & 43.8\\
CenterPoint~\cite{Centerpoint} & CVPR'21 & 75.6 &  48.4 & 57.5 & 36.5 & 31.7 & 41.9 & 48.6\\
PointPainting~\cite{pointpainting} & CVPR'20 & 75.9 & 48.5 & 57.1 & 36.9 & 34.5 & 41.9 & 49.1\\
MVP~\cite{mvp} &  NeurIPS'21 & 76.1 & 48.7 & 57.0 & 36.9 & 33.0 & 42.2 & 49.0 \\
MapFusion\cite{hao2025mapfusion}&\textcolor{black}{IF'25}&88.9& 69.6& 74.0& 63.0& 56.5& 61.5 &68.9\\
NRSeg\cite{li2026nrseg}&\textcolor{black}{TIP'26}&59.1& 16.9& 21.9& 12.1& 16.8& 21.0& 24.6\\
\midrule
BEVFusion~\cite{bevfusion-mit}& ICRA'23 & 85.5 & 60.5 & 67.6 & 52.0 & 57.0 & 53.7 & 62.7\\ 
GraphBEV~\cite{song2024graphbev}& \textcolor{black}{ECCV'24} & 86.3& 60.9& 69.1& 53.1& 57.5& 53.1& 63.3\\ 
\rowcolor{pink!10}\textbf{GraphBEV++ (LSS)}& - & 86.8& 61.5& 69.7& 53.6& 58.1& 53.1& 63.8\\ 
\bottomrule
\end{tabular} }

\label{tab_nuscenes_bevmap}
\end{table*}
\definecolor{nbarrier}{RGB}{255, 120, 50}
\definecolor{nbicycle}{RGB}{255, 192, 203}
\definecolor{nbus}{RGB}{255, 255, 0}
\definecolor{ncar}{RGB}{0, 150, 245}
\definecolor{nconstruct}{RGB}{0, 255, 255}
\definecolor{nmotor}{RGB}{200, 180, 0}
\definecolor{npedestrian}{RGB}{255, 0, 0}
\definecolor{ntraffic}{RGB}{255, 240, 150}
\definecolor{ntrailer}{RGB}{135, 60, 0}
\definecolor{ntruck}{RGB}{160, 32, 240}
\definecolor{ndriveable}{RGB}{255, 0, 255}
\definecolor{nother}{RGB}{139, 137, 137}
\definecolor{nsidewalk}{RGB}{75, 0, 75}
\definecolor{nterrain}{RGB}{150, 240, 80}
\definecolor{nmanmade}{RGB}{213, 213, 213}
\definecolor{nvegetation}{RGB}{0, 175, 0}

\newcommand\crule[3][black]{\textcolor{#1}{\rule{#2}{#3}}}
\definecolor{nvcolor}{RGB}{119,185,0}
\definecolor{roadcolor}{RGB}{234,51,246}
\definecolor{sidewalkcolor}{RGB}{68,8,72}
\definecolor{parkingcolor}{RGB}{241,156,249}
\definecolor{othergroundcolor}{RGB}{160,32,76}
\definecolor{buildingcolor}{RGB}{246,202,69}
\definecolor{carcolor}{RGB}{111,149,238}
\definecolor{truckcolor}{RGB}{74,32,172}
\definecolor{bicyclecolor}{RGB}{136,227,242}
\definecolor{motorcyclecolor}{RGB}{37,59,146}
\definecolor{othervehiclecolor}{RGB}{96,81,242}
\definecolor{vegetationcolor}{RGB}{79, 173, 50}
\definecolor{trunkcolor}{RGB}{126, 65, 22}
\definecolor{terraincolor}{RGB}{171, 238, 105}
\definecolor{personcolor}{RGB}{234, 60, 49}
\definecolor{bicyclistcolor}{RGB}{234, 66, 195}
\definecolor{motorcyclistcolor}{RGB}{138, 42, 90}
\definecolor{fencecolor}{RGB}{238, 128, 69}
\definecolor{polecolor}{RGB}{252, 241, 161}
\definecolor{trafficsigncolor}{RGB}{233, 51, 35}
\definecolor{other-struct.color}{RGB}{255, 150, 0}
\definecolor{other-objectcolor}{RGB}{50, 255, 255}
\definecolor{lane-markingcolor}{RGB}{150, 255, 170}
\definecolor{color1}{RGB}{176, 36, 24}
\definecolor{color2}{RGB}{0, 176, 80}
\definecolor{color3}{RGB}{0, 0, 200}

\begin{table*}[t] %
    \caption[]{\textcolor{black}{3D semantic occupancy prediction results on nuScenes validation set. Since GaussianFormer follows a BEVFormer-style query-based architecture, we adopt GraphBEV++ (Query) in the comparison.}}
    \setlength{\tabcolsep}{0.003\linewidth}  

    \renewcommand\arraystretch{1.2}
    \centering
    \resizebox{\textwidth}{!}{
    \begin{tabular}{l|c|c c | c c c c c c c c c c c c c c c c}
        \toprule
        Method
        &Venue
        &  \makecell{IoU} & \makecell{mIoU}
        & \rotatebox{90}{\textcolor{nbarrier}{$\blacksquare$} barrier}
        & \rotatebox{90}{\textcolor{nbicycle}{$\blacksquare$} bicycle}
        & \rotatebox{90}{\textcolor{nbus}{$\blacksquare$} bus}
        & \rotatebox{90}{\textcolor{ncar}{$\blacksquare$} car}
        & \rotatebox{90}{\textcolor{nconstruct}{$\blacksquare$} const. veh.}
        & \rotatebox{90}{\textcolor{nmotor}{$\blacksquare$} motorcycle}
        & \rotatebox{90}{\textcolor{npedestrian}{$\blacksquare$} pedestrian}
        & \rotatebox{90}{\textcolor{ntraffic}{$\blacksquare$} traffic cone}
        & \rotatebox{90}{\textcolor{ntrailer}{$\blacksquare$} trailer}
        & \rotatebox{90}{\textcolor{ntruck}{$\blacksquare$} truck}
        & \rotatebox{90}{\textcolor{ndriveable}{$\blacksquare$} drive. suf.}
        & \rotatebox{90}{\textcolor{nother}{$\blacksquare$} other flat}
        & \rotatebox{90}{\textcolor{nsidewalk}{$\blacksquare$} sidewalk}
        & \rotatebox{90}{\textcolor{nterrain}{$\blacksquare$} terrain}
        & \rotatebox{90}{\textcolor{nmanmade}{$\blacksquare$} manmade}
        & \rotatebox{90}{\textcolor{nvegetation}{$\blacksquare$} vegetation}
        \\
        \midrule
        MonoScene~\cite{cao2022monoscene} &CVPR'22 & 23.96 &  7.31 &  4.03 &  0.35 &  8.00 &  8.04 &  2.90 &  0.28 &  1.16 &  0.67 &  4.01 &  4.35 & 27.72 &  5.20 & 15.13 & 11.29 &  9.03 & 14.86 \\
        
        Atlas~\cite{murez2020atlas}&  ECCV'20        & 28.66 & 15.00 & 10.64 &  5.68 & 19.66 & 24.94 &  8.90 &  8.84 &  6.47 &  3.28 & 10.42 & 16.21 & 34.86 & 15.46 & 21.89 & 20.95 & 11.21 & 20.54 \\
        
        BEVFormer~\cite{bevformer} &  ECCV'22 & 30.50 & 16.75 & 14.22 &  6.58 & 23.46 & 28.28 &  8.66 & 10.77 &  6.64 &  4.05 & 11.20 & 17.78 & 37.28 & 18.00 & 22.88 & 22.17 & 13.80 & 22.21\\
        
        TPVFormer~\cite{huang2023tri} &CVPR'23 & 11.51 & 11.66 & 16.14 &  7.17 & 22.63	& 17.13 &  8.83 & 11.39 & 10.46 &  8.23 &  9.43 & 17.02 &  8.07 & 13.64 & 13.85 & 10.34 &  4.90 &  7.37\\

        OccFormer~\cite{zhang2023occformer} &ICCV'23 & 31.39 & 19.03 & 18.65 & 10.41 & 23.92 & 30.29 & 10.31 & 14.19 & 13.59 & 10.13 & 12.49 & 20.77 & 38.78 & 19.79 & 24.19 & 22.21 & 13.48 & 21.35\\

        GaussianFormer~\cite{huang2024gaussianformer} &\textcolor{black}{ECCV'24}
                                    & 29.83 & 19.10 & 19.52 & 11.26 & 26.11 & 29.78 & 10.47 & 13.83 & 12.58 & 8.67 & 12.74 & 21.57 & 39.63 & 23.28 & 24.46 & 22.99 & 9.59 & 19.12 \\
        
        SurroundOcc~\cite{wei2023surroundocc} &CVPR'23
                                      & 31.49 & 20.30 & 20.59 & 11.68      & 28.06 & 30.86 & 10.70  & 15.14      & 14.09 & 12.06  & 14.38 & 22.26 & 37.29 &         
                                        23.70 &   24.49     &     22.77   & 14.89 & 21.86  \\

QuadricFormer\cite{zuo2026quadricformer}&\textcolor{black}{NeurIPS'26}&31.22 &20.12 &19.58 &13.11 &27.27 &29.64 &11.25& 16.26 &12.65& 9.15& 12.51 &21.24 &40.20 &24.34 &25.69& 24.24 &12.95 &21.86\\
GaussianFormer-2\cite{huang2025gaussianformer}&\textcolor{black}{CVPR'25}&31.74 &20.82 &21.39 &13.44 &28.49 &30.82& 10.92 &15.84& 13.55& 10.53 &14.04 &22.92& 40.61& 24.36 &26.08& 24.27& 13.83& 21.98\\
Gau-Occ\cite{lv2026gau} &\textcolor{black}{CVPR'26}&44.30& 32.70& 33.10& 16.50& 41.10 &43.90& 21.90 &26.60 &31.10 &23.10 &24.50& 34.00& 49.00& 26.30 &32.90& 33.40&39.30 &45.70\\
        GaussianWorld\cite{zuo2025gaussianworld}&\textcolor{black}{CVPR'25} & 33.40 & 22.13 & 21.38 & 14.12 & 27.71 & 31.84 & 13.66 & 17.43 & 13.66 & 11.46 & 15.09 & 23.94 & 42.98 &                                     24.86 & 28.84 & 26.74 & 15.69 & 24.74 \\

        \midrule
        GaussianFormer-T\cite{huang2024gaussianformer}&\textcolor{black}{ECCV'24}& 31.34& 20.42& 20.82& 12.07& 26.89& 30.94 &10.52 &16.48 &13.15& 10.46& 12.90& 21.79& 41.13& 24.22& 26.29& 24.89& 12.80& 21.45\\
        \rowcolor{pink!10}GraphBEV++  &-   & 32.47       & 21.90       & 21.66& 12.83 &27.70& 31.75&
11.39& 17.15& 13.89& 11.16&
13.80& 22.74& 41.90 &24.98&
27.12& 25.84 &13.48& 22.23 \\
        \midrule
\multicolumn{19}{c}{Performances on \textcolor{red}{Noisy} setting} \\
       \midrule
GaussianFormer-T\cite{huang2024gaussianformer} &\textcolor{black}{ECCV'24}    & 27.86 & 17.63 & 17.42 & 8.38 & 22.91 & 27.48 & 7.56 & 12.03 & 9.81 & 6.87 & 8.72 & 18.02 & 37.18 & 20.44 & 22.67 & 21.18 & 9.22 & 17.74\\        
 \rowcolor{pink!10}GraphBEV++&- & 29.41 & 19.37 & 18.74 & 10.43 & 24.62 & 28.39 & 8.71 & 13.92 & 11.28 & 8.74 & 10.61 & 19.58 & 38.71 & 21.56 & 24.03 & 22.77 & 10.92 & 19.31 \\

        \bottomrule
    \end{tabular}}
    \label{tab_nusc_occ}

\end{table*}

\begin{table}[htp]
\scriptsize
\centering
  \caption[]{The results of \textbf{multi-object tracking} in end-to-end autonomous driving  (E2E-AD) tasks on nuScenes\cite{nuscenes} dataset. 
  % Our GraphBEV++ remarkably outperforms previous methods.
  }
  \renewcommand\arraystretch{1}
  \tabcolsep=2.5mm %%%%%%%%%
  \resizebox{\linewidth}{!}{
  %\begin{tabular*}{\linewidth} {@{}@{\extracolsep{\fill}}!{\color{white}\vline}l|c|c|c|c|c|c|c|c|c|c|c|c @{}}
  \begin{tabular}{l|c|cc}
    \toprule
Method          &    Venue                                &     AMOTA (\%) ↑           & AMOTP (m) ↓          \\ 

\midrule
ViP3D~\cite{gu2023vip3d}      &CVPR'23   &     21.7                 &    1.625            \\
QD3DT~\cite{QD3DT}             &TPAMI'22  &     24.2                 &    1.518            \\ 
MUTR3D~\cite{zhang2022mutr3d}        &CVPR'22               &     29.4                 &    1.498            \\
DEFT~\cite{chaabane2021deft} &Arxiv'21      & 20.1 & - \\
DQTrack~\cite{DQTrack}     &ICCV'23          & 36.7 & 1.351 \\
STAR-Track~\cite{doll2023star} &RAl'23            & 37.9 & 1.358 \\
CC-3DT~\cite{fischer2022cc}     &Arxiv'22        & 42.9 & 1.257 \\
PF-Track~\cite{pang2023standing}   &CVPR'23        & 40.8 & 1.343 \\
SparseDrive~\cite{sun2025sparsedrive} & \textcolor{black}{ICRA'25}       & 38.6 & 1.254 \\
BridgeAD~\cite{zhang2025bridging}  &\textcolor{black}{CVPR'25 }           & 39.8 & 1.232 \\
MomAD\cite{momad} &\textcolor{black}{CVPR'25 } &39.1 &1.243\\
UC-Track \cite{wu2026uc}  &\textcolor{black}{TITS'26  }              & 43.8 & 1.290 \\
UniAD~\cite{uniad}     &CVPR'23          &     35.9                 &    1.320            \\
FusionAD~\cite{fusionad}       &Arxiv'23  &     50.1                 &    1.065            \\
\rowcolor{pink!10}\textbf{GraphBEV++ (LSS)}   &-     &   49.8& 1.082               \\
\rowcolor{pink!10}\textbf{GraphBEV++ (Query)}           &-              &   51.1          & 1.022     \\
\bottomrule
\end{tabular} }

\label{tab_nuscenes_multi_object_tracking}
\end{table}

\begin{table}[htp]
\scriptsize
\centering
  \caption[]{The results of \textbf{online mapping} in end-to-end autonomous driving tasks  on nuScenes\cite{nuscenes} dataset. 
  % Our GraphBEV++ remarkably outperforms previous methods.
  }
  \renewcommand\arraystretch{1}
  \tabcolsep=2.5mm %%%%%%%%%
  \resizebox{\linewidth}{!}{
  %\begin{tabular*}{\linewidth} {@{}@{\extracolsep{\fill}}!{\color{white}\vline}l|c|c|c|c|c|c|c|c|c|c|c|c @{}}
  \begin{tabular}{l|c|cc}
    \toprule
Method        &  Venue                                       &    IoU-Lanes (\%) ↑  &IoU-Drivable (\%) ↑          \\ 

\midrule
VPN~\cite{VPN}&RAL'20 & 18.0 & 76.0\\
LSS~\cite{lss}&ECCV'20 & 18.3 & 73.9\\
BEVFormer~\cite{bevformer}&ECCV'22 & 23.9 & 77.5\\

FusionAD~\cite{fusionad}&Arxiv'23 & 36.7& 73.1\\
UniAD~\cite{uniad}&CVPR'23 & 31.3 & 69.1\\
ParaDrive\cite{paradrive}&\textcolor{black}{CVPR'24}&71.0&33.0\\
DriveTransformer\cite{jia2025drivetransformer}&\textcolor{black}{ICLR'25}&77.0&39.0\\
\rowcolor{pink!10}\textbf{GraphBEV++ (LSS)}&- & 36.4 & 73.9\\
\rowcolor{pink!10}\textbf{GraphBEV++ (Query)}&- & 37.1 & 73.4\\
\bottomrule
\end{tabular} }

\label{tab_nuscenes_online_mapping}
\end{table}

\begin{table}[htp]
\scriptsize
\centering
  \caption[]{The results of \textbf{motion forecasting} in end-to-end autonomous driving tasks.}
  \renewcommand\arraystretch{1}
  \tabcolsep=0.4mm %%%%%%%%%
  \resizebox{\linewidth}{!}{
  %\begin{tabular*}{\linewidth} {@{}@{\extracolsep{\fill}}!{\color{white}\vline}l|c|c|c|c|c|c|c|c|c|c|c|c @{}}
  \begin{tabular}{l|c|cccc}
    \toprule
Method      &    Venue                                    &    minADE (m) ↓       & minFDE (m) ↓    &  MR (\%) ↓         &  EPA (\%) ↑       \\ 

\midrule
UniAD~\cite{uniad}      &    CVPR‘23                        &      0.71            & 1.02           & 15.1         &45.6         \\
VAD~\cite{jiang2023vad}    &ICCV'23                         &      0.68            & 0.88           & 8.3         &-             \\
FusionAD~\cite{fusionad}     &Arxiv’23                        &      0.39            & 0.62           & 8.6         &62.6         \\
SparseDrive\cite{sun2024sparsedrive}&\textcolor{black}{ICRA'25}&0.62& 0.99& 13.6& 48.2\\

SparseWorld&\textcolor{black}{Arxiv'26}&0.78& 1.03 &11.3& 61.9\\
MomAD\cite{momad}&\textcolor{black}{CVPR'25}&0.61& 0.98&13.7& 49.9\\
CoT-Drive\cite{liao2025cot}&\textcolor{black}{TAI'25}&1.56 &3.49& 52.0\\
\rowcolor{pink!10}\textbf{GraphBEV++ (LSS)}    &-                             &   0.40      & 0.59  &8.5 &\textbf{64.7}\\
\rowcolor{pink!10}\textbf{GraphBEV++ (Query)}        &-                         &   \textbf{0.38}      & \textbf{0.52}  &\textbf{7.7} &64.5\\

\bottomrule
\end{tabular} }

\label{tab_nuscenes_motion_forecasting}
\end{table}

% \begin{table}[htp]
% \scriptsize
% \centering
%   \caption[]{The results of \textbf{occupancy prediction} in end-to-end autonomous driving tasks  on nuScenes\cite{nuscenes} dataset. `n.' and `f.' indicate near (30 $\times$ 30m) and far (100 $\times$ 100m) evaluation ranges, respectively.}
%   \renewcommand\arraystretch{1}
%   \tabcolsep=0.3mm %%%%%%%%%
%   \resizebox{\linewidth}{!}{
%   %\begin{tabular*}{\linewidth} {@{}@{\extracolsep{\fill}}!{\color{white}\vline}l|c|c|c|c|c|c|c|c|c|c|c|c @{}}
%   \begin{tabular}{l|cccc}
%     \toprule
% Method           & IoU-n. (\%) ↑ & IoU-f. (\%) ↑ & VPQ-n. (\%) ↑ & VPQ-f. ↑ (\%)\\
% \midrule
% % FIERY~\cite{hu2021fiery}       & 59.4    &  36.7   &  50.2   & 29.9   \\ 
% % StretchBEV~\cite{akan2022stretchbev}   & 55.5    &  37.1   &  46.0   & 29.0   \\ 
% ST-P3~\cite{ST_P3}       &  -      &  38.9   &   -     & 32.1   \\ 
% FusionAD~\cite{fusionad}            & 71.2& 51.5& 65.5& 51.1  \\
% UniAD~\cite{uniad}            & 63.4    &  40.2   &  54.7   & 33.5   \\
% \rowcolor{pink!10}\textbf{GraphBEV++ (LSS)}       & \textbf{72.6}    &  \textbf{52.4}   &  \textbf{66.5}   & 51.4   \\
% \rowcolor{pink!10}\textbf{GraphBEV++ (Query)}       & 72.3    &  52.2   & 66.1   & \textbf{51.5}   \\
% \bottomrule
% \end{tabular} }

% \label{tab_nuscenes_occupancy_prediction}
% \end{table}

\begin{table}[htp]
\centering
  \caption[]{The results of \textbf{planning} in end-to-end autonomous driving tasks on nuScenes~\cite{nuscenes} dataset. As Ref.~\cite{li2024ego} states, we \textbf{deactivate} the \textbf{ego status} information for a fair comparison.}
  \renewcommand\arraystretch{1}
  \tabcolsep=0.8mm %%%%%%%%%
  \resizebox{\linewidth}{!}{
  %\begin{tabular*}{\linewidth} {@{}@{\extracolsep{\fill}}!{\color{white}\vline}l|c|c|c|c|c|c|c|c|c|c|c|c @{}}
  \begin{tabular}{l|c|cccc | cccc}
    \toprule
\multirow{2}{*}{Method}   &\multirow{2}{*}{Venue}          & \multicolumn{4}{c|}{L2 (m)↓} & \multicolumn{4}{c}{Col. Rate (\%)↓}\\
& &1s      & 2s      &    3s    &   Avg.  &  1s   &   2s   &  3s   & Avg.\\
\midrule
NMP~\cite{NMP} & CVPR'19    & -     & -    &2.31 & -   & -  & -  &1.92& -   \\
SA-NMP~\cite{NMP}  & CVPR'19    & -     & -    &2.05 & -   & -  & -  &1.59& -   \\ 
FF~\cite{FF} & CVPR'21       &0.55   &1.20  &2.54 & 1.43&0.66&0.17&1.07&0.43 \\
EO~\cite{EO}    & ECCV'22   &0.67   &1.36  &2.78 &1.60 &0.04&0.09&0.88&0.33 \\
ST-P3~\cite{ST_P3} & ECCV'22    &1.33   &2.11  &2.90 &2.11 &0.23&0.62&1.27&0.71 \\ 
GPT-Driver~\cite{gpt_driver}&ArXiv'23&0.27   &0.74  &1.52 &0.84 &0.07&0.15&1.10&0.44 \\ 
UniAD~\cite{uniad}  & CVPR'23     &0.48   &0.96  &1.65 &1.03 &0.05&0.17&0.71&0.31 \\
FusionAD~\cite{fusionad} & ArXiv'23     &0.02& 0.08& 0.27 &0.81 &0.05&0.17&0.71&0.12 \\
$\operatorname{VAD}$~\cite{jiang2023vad}&ICCV'23 &0.41& 0.70& 1.05& 0.72 &0.11& 0.24& 0.42 &0.26 \\
$\operatorname{DiffusionDrive}$~\cite{liao2024diffusiondrive}&\textcolor{black}{CVPR'25} &0.29& 0.58& 0.96& 0.61& 0.02& 0.05& 0.22& 0.09 \\
$\operatorname{DIVER}$~\cite{diver} &\textcolor{black}{Arxiv'26}&-& -& -& -& 0.01& 0.05&0.15& 0.07 \\
$\operatorname{FocalAD}$~\cite{sun2026focalad}&\textcolor{black}{Aut. Inno.'26} &0.27& 0.57& 0.96& 0.60 &0.00& 0.04 &0.24& 0.09 \\
$\operatorname{GuideFlow}$~\cite{liu2026guideflow}&\textcolor{black}{CVPR'26} &-& -& -& -&0.00& 0.02& 0.18& 0.07 \\
$\operatorname{SparseDrive}$~\cite{sun2024sparsedrive}&\textcolor{black}{ICRA'25} &0.30& 0.58&0.96& 0.61& 0.01 &0.05& 0.23& 0.10 \\
$\operatorname{LAW}$~\cite{li2024enhancing_law}&\textcolor{black}{ICLR'25}&0.26 &0.57 &1.01 &0.61& 0.14 &0.21& 0.54 &0.30\\
$\operatorname{GenAD}$~\cite{zheng2024genad}&\textcolor{black}{ECCV'24}&0.28 &0.49 &0.78 &0.52 &0.08 &0.14& 0.34& 0.19\\
$\operatorname{Drive-OccWorld}$~\cite{zheng2024occworld}&\textcolor{black}{ECCV'24}&0.25& 0.44 &0.72& 0.47 &0.03 &0.08 &0.22 &0.11\\
$\operatorname{SSR}$~\cite{ssr}&\textcolor{black}{ICLR'25}&0.19 &0.36& 0.62& 0.39 &0.10 &0.10& 0.24& 0.15\\
$\operatorname{MomAD}$~\cite{momad}&\textcolor{black}{CVPR'25}&0.31& 0.57 &0.91 &0.60& 0.01& 0.05& 0.22& 0.09\\
$\operatorname{DriveWorld-VLA}$~\cite{liu2026driveworld}&\textcolor{black}{ICML'26}&0.28& 0.58 &0.99 &0.61& 0.00& 0.10& 0.38& 0.16 \\
\rowcolor{pink!10}\textbf{GraphBEV++ (LSS)} &-   &0.33& 0.66& 1.02& 0.67& 0.06& 0.21& 0.36& 0.21 \\
\rowcolor{pink!10}\textbf{GraphBEV++ (Query)} &-   &0.40& 0.67& 1.04& 0.70& 0.03& 0.07& 0.29& 0.13 \\
\bottomrule
\end{tabular} }
\label{tab_nuscenes_planning}
\end{table}

\begin{table}[htp]
\centering
\caption[]{Open-loop and Closed-loop results on \textbf{Bench2Drive}~\cite{jia2024bench2drive} (V0.0.3) under base training set.
$^{*}$ denotes expert feature distillation. $\dagger$ denotes the re-implementation. 
`mmt' refers to the multi-mode trajectory variant.
}

  \tabcolsep=0.3mm %%%%%%%%%
  \resizebox{\linewidth}{!}{
  \begin{tabular}{l|c|c|cccc}
    \toprule
   \multirow{2}{*}{\textbf{$\operatorname{Method}$}} 
   & \multirow{2}{*}{\textbf{$\operatorname{Venue}$}} 
   & \textbf{$\operatorname{Open-loop\ Metric}$}     
   & \multicolumn{4}{c}{\textbf{$\operatorname{Closed-loop\ Metric}$}}  \\
   \cmidrule(lr){3-3} \cmidrule(lr){4-7} 
   & & $\operatorname{Avg.\ L2}\downarrow$ 
   & $\operatorname{DS}\uparrow$ 
   & $\operatorname{SR\ (\%)}\uparrow$ 
   & $\operatorname{Effi}\uparrow$  
   & $\operatorname{Comf}\uparrow$ \\
\midrule
ThinkTwice$^{*}$~\cite{thiktwice} & CVPR'23 & 0.95 & 62.44 & 31.23 & 69.33 & 16.22 \\
DriveAdapter$^{*}$~\cite{jia2023driveadapter} & ICCV'23 & 1.01 & 64.22 & 33.08 & 70.22 & 16.01 \\
VAD~\cite{jiang2023vad} & ICCV'23 & 0.91 & 42.35 & 15.00 & 157.94 & 46.01 \\
DriveDPO~\cite{shang2025drivedpo} & \textcolor{black}{NeurIPS'25} & - & 62.02 & 30.62 & - & - \\
Raw2Drive~\cite{Raw2Drive} & \textcolor{black}{ NeurIPS'25 }& - & 71.36 & 50.24 & - & - \\
DriveTrans$^{*}$~\cite{jia2025drivetransformer} &  \textcolor{black}{ICLR'25} & 0.62 & 63.46 & 35.01 & 100.64 & 20.78 \\
MomAD~\cite{momad} &  \textcolor{black}{CVPR'25} & 0.85 & 45.35 & 17.44 & 162.09 & 49.34 \\
WoTE$^{*}$~\cite{wote} &  \textcolor{black}{ICCV'25} & - & 61.71 & 31.36 & - & - \\
${\operatorname{ThinkTwice}_{\operatorname{mmt}}}^{*\dagger}$~\cite{thiktwice} & CVPR'23 & 0.93 & 63.34 & 33.23 & 71.56 & 18.32 \\
\midrule
\rowcolor{pink!10}\textbf{GraphBEV++ (Query)} & - & 0.82 & 45.71 & 17.92 & 162.82 & 50.14 \\
\bottomrule
\end{tabular} }
\label{tab_b2d}
\end{table}

\begin{table}[t]
\centering
  \caption[]{Comparison on planning-oriented NAVSIM~\cite{dauner2024navsim} navtest split with $\operatorname{\textbf{Closed-Loop}}$ metrics.}
\renewcommand\arraystretch{1}
  \tabcolsep=0.8mm %%%%%%%%%
  \resizebox{\linewidth}{!}{
\begin{tabular}{l|c|c|cccccc}
\toprule
\multicolumn{1}{l|}{Method} & \multicolumn{1}{c|}{Venue} & \multicolumn{1}{c|}{Input} & NC$\uparrow$ & \multicolumn{1}{c}{DAC}$\uparrow$ & TTC$\uparrow$ & Comf.$\uparrow$ & EP$\uparrow$ & PDMS$\uparrow$ \\
\midrule
UniAD~\cite{uniad} & CVPR'23 & C & 97.8 & 91.9 & 92.9 & 100 & 78.8 & 83.4 \\
LTF~\cite{TransFuser} & TPAMI'22 & C & 97.4 & 92.8 & 92.4 & 100 & 79.0 & 83.8 \\
PARA-Drive~\cite{paradrive} & \textcolor{black}{CVPR'24} & C & 97.9 & 92.4 & 93.0 & 99.8 & 79.3 & 84.0 \\
VADv2~\cite{chen2024vadv2} & \textcolor{black}{ICLR'26} & LC & 97.2 & 89.1 & 91.6 & 100 & 76.0 & 80.9 \\
Hydra-MDP~\cite{li2024hydra} & \textcolor{black}{Arxiv'24} & LC & 98.3 & 96.0 & 94.6 & 100 & 78.7 & 86.5 \\
DiffusionDrive~\cite{liao2024diffusiondrive} & \textcolor{black}{CVPR'25} & LC & 98.2 & 96.2 & 94.7 & 100 & 82.2 & 88.1 \\
\midrule
WoTE~\cite{wote} & \textcolor{black}{ICCV'25} & LC & 98.5 & 96.8 & 81.9 & 94.9 & 99.9 & 88.3 \\ 
\rowcolor{pink!10}GraphBEV++ (Query) & - & LC & 98.7 & 97.1 & 82.0 & 95.1 & 99.9 & 88.7 \\ 
\bottomrule
\end{tabular}}
\label{tab_navsim}
\end{table}

\begin{table}[t]
% \scriptsize
\centering
  \caption[]{\textcolor{black}{Efficiency comparison with SOTA multi-modal 3D detection methods. FPS is reported or estimated on an A100 GPU.
  }}
  \renewcommand\arraystretch{1}
  \tabcolsep=1.8mm %%%%%%%%%
  \resizebox{\linewidth}{!}{
  \begin{tabular}{l|c|ccc}
    \toprule
 Method  & Modality & Model Size & FLOPs & FPS \\ 
\midrule
RoboFusion~\cite{song2024robofusion} & LC & 97.54M& 1.23T& 3.1 \\
DeepInteraction~\cite{DeepInteraction} & LC & 57.82M& 775.2G& 4.9 \\
TransFusion~\cite{Transfusion} & LC & 36.96M& 612.6G& 6.0 \\
BEVFusion~\cite{bevfusion-mit} & LC & 40.81M& 506.4G & 7.5 \\
GraphBEV~\cite{song2024graphbev} & LC & 47.80M& 535.7G& 7.2 \\
BEVFormer-S-C~\cite{BEVFormerpami} & LC & 68.70M& 1303.5.7G& 5.3 \\
\rowcolor{pink!10}\textbf{GraphBEV++ (LSS)} & LC & 48.12M& 457.6G& 7.1 \\
\rowcolor{pink!10}\textbf{GraphBEV++ (Query)} & LC & 69.76M& 1357.6G& 4.9 \\
\bottomrule
\end{tabular} }
\label{tab_efficiency_comparison}
\end{table}

\begin{table}[t]
% \scriptsize
\centering
  \caption[]{Roles of different modules in GraphBEV++ (LSS) for feature alignment on nuScenes (\textcolor{blue}{clean}) and nuScenes-C (\textcolor{red}{noisy}) dataset. \textcolor{darkgreen}{Green} denotes the relative performance drop from the clean to the noisy settings.   `+L (LSS)' indicates the addition of only the LocalAlign-v2 (LSS) module, and `+G (Deformable)' indicates only the GlobalAlign-v2 (Deformable) module. GraphBEV++ (LSS) denotes the addition of both LocalAlign-v2 (LSS) and GlobalAlign-v2 (Deformable) modules.  All latency measurements are conducted on the same workstation with an A100 GPU.
  }
  \renewcommand\arraystretch{1}
  \tabcolsep=1.8mm %%%%%%%%%
  \resizebox{\linewidth}{!}{
  %\begin{tabular*}{\linewidth} {@{}@{\extracolsep{\fill}}!{\color{white}\vline}l|c|c|c|c|c|c|c|c|c|c|c|c @{}}
  \begin{tabular}{l|c|ccc }
    \toprule
 &Method            & mAP  & NDS & Latency(ms)   \\ 
\cmidrule(r){2-5}
&TransFusion~\cite{Transfusion}& 67.3& 71.2&164.6 \\
\cmidrule(r){2-5}
\multirow{7}{*}{\textcolor{blue}{\rotatebox{90}{\textbf{Clean}}}} &Baseline~\cite{bevfusion-mit}&  68.5 & 71.4 &133.2  \\
\cmidrule(r){2-5}
&\multirow{2}{*}{+L (LSS)}& 69.8&72.6&135.1\\
 &
 &\textit{\fontsize{6}{0}\selectfont\textcolor{black}{+1.3}}
 &\textit{\fontsize{6}{0}\selectfont\textcolor{black}{+1.2}}
 &\textit{\fontsize{6}{0}\selectfont\textcolor{black}{+2.9}}
 \\
\cmidrule(r){2-5}
&\multirow{2}{*}{+G (Deformable)}&  68.9 & 71.7 &138.1\\
 &
 &\textit{\fontsize{6}{0}\selectfont\textcolor{black}{+0.4}}
 &\textit{\fontsize{6}{0}\selectfont\textcolor{black}{+0.3}}
 &\textit{\fontsize{6}{0}\selectfont\textcolor{black}{+4.9}}
 \\
\cmidrule(r){2-5}
&\textbf{GraphBEV++ (LSS)}&   70.7&73.2&140.9\\ 
 &
 &\textit{\fontsize{6}{0}\selectfont\textcolor{black}{+1.6}}
 &\textit{\fontsize{6}{0}\selectfont\textcolor{black}{+1.5}}
 &\textit{\fontsize{6}{0}\selectfont\textcolor{black}{+7.7}}
 \\
\midrule
\midrule
&TransFusion~\cite{Transfusion}& 66.4\textit{\fontsize{6}{0}\selectfont\textcolor{darkgreen}{-1.3\%}}& 70.6\textit{\fontsize{6}{0}\selectfont\textcolor{darkgreen}{-0.8\%}}&164.6 \\
\cmidrule(r){2-5}
\multirow{8}{*}{\textcolor{red}{\rotatebox{90}{\textbf{Noisy}}}}&Baseline~\cite{bevfusion-mit}&  60.8\textit{\fontsize{6}{0}\selectfont\textcolor{darkgreen}{-11.2\%}} & 65.7\textit{\fontsize{6}{0}\selectfont\textcolor{darkgreen}{-8.2\%}} &132.9 \\ 

\cmidrule(r){2-5}
&\multirow{2}{*}{+L (LSS)}&   67.4\textit{\fontsize{6}{0}\selectfont\textcolor{darkgreen}{-3.4\%}}&70.4\textit{\fontsize{6}{0}\selectfont\textcolor{darkgreen}{-3.0\%}}&135.8\\ 
 &&\textit{\fontsize{6}{0}\selectfont\textcolor{red}{+6.6}}
 &\textit{\fontsize{6}{0}\selectfont\textcolor{red}{+4.7}}
 &\textit{\fontsize{6}{0}\selectfont\textcolor{red}{+2.9}}

 \\
 \cmidrule(r){2-5}
 &\multirow{2}{*}{+G (Deformable)}&    63.1\textit{\fontsize{6}{0}\selectfont\textcolor{darkgreen}{-9.6\%}}&67.2\textit{\fontsize{6}{0}\selectfont\textcolor{darkgreen}{-6.3\%}}&137.9\\  
 &&\textit{\fontsize{6}{0}\selectfont\textcolor{red}{+2.3}}
 &\textit{\fontsize{6}{0}\selectfont\textcolor{red}{+1.5}}
 &\textit{\fontsize{6}{0}\selectfont\textcolor{red}{+5.0}}
 \\
 \cmidrule(r){2-5}
 &\textbf{GraphBEV++ (LSS)}&   69.3\textit{\fontsize{6}{0}\selectfont\textcolor{darkgreen}{-2.0\%}}&72.3\textit{\fontsize{6}{0}\selectfont\textcolor{darkgreen}{-1.2\%}}&141.2\\ 
 &&\textit{\fontsize{6}{0}\selectfont\textcolor{red}{+8.5}}
 &\textit{\fontsize{6}{0}\selectfont\textcolor{red}{+6.6}}
 &\textit{\fontsize{6}{0}\selectfont\textcolor{red}{+8.3}}
 \\
\bottomrule
\end{tabular} }

\label{tab_nuscenes_misalignment_lss}
\end{table}

\begin{table}[t]
% \scriptsize
\centering
  \caption[]{Roles of different modules in GraphBEV++ (Query) for feature alignment on  nuScenes-C (\textcolor{red}{noisy}) dataset.  `+L (Query)' indicates the addition of only the LocalAlign-v2 (Query) module, and `+G (Diffusion)' indicates only the GlobalAlign-v2 (Diffusion) module. GraphBEV++ (Query) denotes the addition of both LocalAlign-v2 (LSS) and GlobalAlign-v2 (Diffusion) modules. All latency measurements are conducted on the same workstation with an A100 GPU.
  }
  \renewcommand\arraystretch{1}
  \tabcolsep=7.8mm %%%%%%%%%
  \resizebox{\linewidth}{!}{
  %\begin{tabular*}{\linewidth} {@{}@{\extracolsep{\fill}}!{\color{white}\vline}l|c|c|c|c|c|c|c|c|c|c|c|c @{}}
  \begin{tabular}{l cc }
    \toprule
 Method            & mAP  & NDS    \\ 
\midrule
BEVFormer-M~\cite{BEVFormerpami}&  63.2 & 66.3  \\
\multirow{2}{*}{+L (Query)}& 66.3&69.0\\
 
 &\textit{\fontsize{6}{0}\selectfont\textcolor{black}{+3.1}}
 &\textit{\fontsize{6}{0}\selectfont\textcolor{black}{+2.7}}
 \\

\multirow{2}{*}{+G (Diffusion)}&  66.2 & 68.1 \\
 
 &\textit{\fontsize{6}{0}\selectfont\textcolor{black}{+3.0}}
 &\textit{\fontsize{6}{0}\selectfont\textcolor{black}{+1.8}}

 \\
\midrule
\rowcolor{pink!10}\textbf{GraphBEV++ (Query)}&   69.1&71.2\\ 
 
 &\textit{\fontsize{6}{0}\selectfont\textcolor{black}{+5.9}}
 &\textit{\fontsize{6}{0}\selectfont\textcolor{black}{+4.9}}

 \\

\bottomrule
\end{tabular} }

\label{tab_nuscenes_misalignment_query}
\end{table}

\begin{table}[t]
\centering
\caption[]{\textcolor{black}{Effect of diffusion steps $T$ on accuracy and inference speed under the nuScenes-C noisy setting.}}
  \renewcommand\arraystretch{1}
  \tabcolsep=9.8mm %%%%%%%%%
  \resizebox{\linewidth}{!}{
\begin{tabular}{c|ccc}
\toprule
$T$ & mAP & NDS & FPS \\
\midrule
1 & 67.8 & 69.6 & 5.21 \\
2 & 68.5 & 70.3 & 5.06 \\
\rowcolor{pink!10}4 & 69.1 & 71.2 & 4.90 \\
8 & 69.2 & 71.3 & 4.57 \\
\bottomrule
\end{tabular}}
\label{tab_diffusion_steps}
\end{table}

\begin{table}[htp]
\scriptsize
\centering
  \caption[]{Evaluation of depth prediction on the \textbf{nuScenes} validation set. Following BEVDepth~\cite{bevdepth}, `soft’ and `hard’ refer to Gaussian and one-hot depth randomization, respectively. `learned' denotes GraphBEV++ (LSS).}
  \renewcommand\arraystretch{1}
  \tabcolsep=3.3mm %%%%%%%%%
  \resizebox{\linewidth}{!}{
  %\begin{tabular*}{\linewidth} {@{}@{\extracolsep{\fill}}!{\color{white}\vline}l|c|c|c|c|c|c|c|c|c|c|c|c @{}}
  \begin{tabular}{l|ccc}
    \toprule
$D^{\mathrm{pred}}$          & mAP↑& mATE↓& NDS↑\\
\midrule
random soft& 46.1& 68.3& 48.9\\
random hard &41.3& 74.6&43.2\\
learned& 70.7&27.6&73.2\\
\bottomrule
\end{tabular} }

\label{tab_nuscenes_bevdepth}
\end{table}

\begin{table*}[]
\centering
\caption[]{The performance of end-to-end methods (UniAD~\cite{uniad}, FusionAD~\cite{fusionad}, GraphBEV++ ) under nuScene (clean) and nuScenes-C (misalignment) conditions. ``↑" indicates better performance with higher metrics, while ``↓" indicates better performance with lower metrics.}
\renewcommand\arraystretch{1}
\tabcolsep=0.3mm %%%%%%%%%
\resizebox{\linewidth}{!}{
\begin{tabular}{l|c|cc|cc|ccc|cccc|cc}
\toprule
\multirow{2}{*}{Method} & \multirow{2}{*}{Setting} & \multicolumn{2}{c|}{Tracking} & \multicolumn{2}{c|}{Mapping} &\multicolumn{3}{c|}{Motion Forecasting} & \multicolumn{4}{c|}{Occupancy} & \multicolumn{2}{c}{Planning} \\
                        & & AMOTA  ↑         & AMOTP  ↓        & IoU-Lanes  ↑  &IoU-Drivable  ↑    & minADE  ↓       & minFDE  ↓    &  MR  ↓    & IoU-n.  ↑ & IoU-f.  ↑ & VPQ-n.  ↑ & VPQ-f. ↑    & avg.L2↓      & avg.Col↓      \\
                      \midrule

\multirow{2}{*}{FusionAD~\cite{fusionad}}& \textcolor{blue}{clean}  & 50.2 & 1.059 & 36.8 & 73.2& 0.38 & 0.61 & 8.4 &70.5 & 51.0 &64.9 &50.3 &0.70 & 0.11\\
% \cmidrule{2-15}
 & \textcolor{red}{Noisy} & 38.8	& 1.245 & 33.0 & 68.3 &0.46&	0.72	& 9.5 & 65.5	& 42.6 & 60.2 &	41.7 &	0.77 & 0.12\\

                       %& $Noisy$ & 0.398	& 1.245 & 0.334 & 0.683 &0.459&	0.725	& 0.095 & 65.5	& 42.6 & 60.2 &	41.7 &	0.775 & 0.121\\ 
                       
\midrule

\multirow{2}{*}{UniAD~\cite{uniad}}& \textcolor{blue}{clean}  & 35.6 & 1.343& 31.3 & 69.1 & 0.70	&1.01	&14.9 &55.2	&34.2 & 63.8	&40.5 & 1.04	&0.29\\
% \cmidrule{2-15}
 & \textcolor{red}{Noisy} & 21.5	& 1.566 & 25.1 & 61.6 &0.84&	1.32	& 26.1 & 49.2	& 26.4 & 38.1 &	21.5 &	1.23 & 0.48\\
\midrule
\multirow{2}{*}{\textbf{GraphBEV++ (LSS)}} & \textcolor{blue}{clean}& 49.8  & 1.082&34.1 & 73.9&0.40      & 0.59  &8.5& 72.6    &  52.4   &  66.5   & 51.4 &0.67&0.21 

\\

% \cmidrule{2-15}
& \textcolor{red}{Noisy} &44.5&1.102& 31.4&69.1&0.44&0.63&8.8&66.9&48.1&62.5&48.2&0.73&0.22

\\
\midrule
\multirow{2}{*}{\textbf{GraphBEV++ (Query)}} & \textcolor{blue}{clean}
 &51.1          & 1.022 & 37.1 & 73.4&0.38      & 0.52  &7.7& 72.3    &  52.2   & 66.1   & 51.5&0.70&0.13
\\

% \cmidrule{2-15}

 &\textcolor{red}{Noisy} &47.0&1.124&34.6&69.6&0.45&0.57&8.4&66.5&47.8&63.2&47.9&0.76&0.14

\\
\bottomrule
\end{tabular}
\label{tab_nuscenes_e2e_misalignment}
}
\end{table*}

\begin{figure*}
    \centering
    \includegraphics[width=\linewidth]{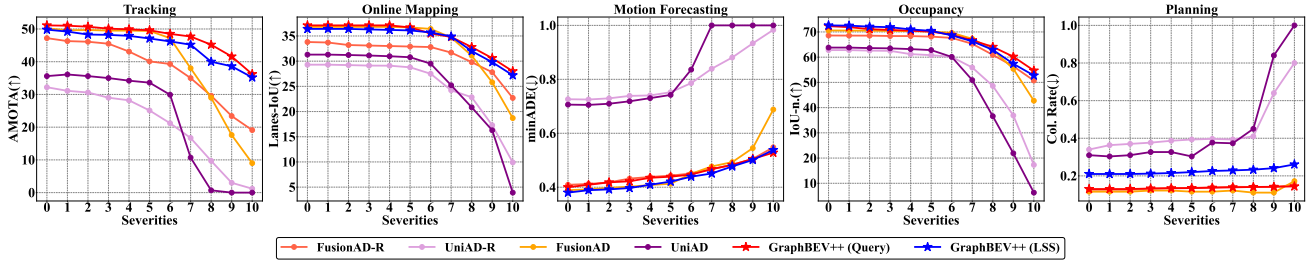}
    \caption[]{End-to-end model performance with respect to the severity of misalignment noise in autonomous driving tasks. In autonomous driving tasks, we followed the alignment noise augmentation method by Dong et al.~\cite{zhujun_benchmarking}, introducing noise of varying severities to the projection matrices used in both the multi-modal approach, FusionAD~\cite{fusionad}, and the vision-based approach, UniAD~\cite{uniad}. The severity scale ranges from 0, indicating no noise (Clean), to 1 for slight noise, with increasing levels up to 10 for severe noise. As the noise level increased, we observed a significant misalignment impact on UniAD~\cite{uniad}, followed by FusionAD~\cite{fusionad}. Our proposed multi-modal end-to-end autonomous driving framework, GraphBEV++ (LSS) and GraphBEV++ (Query), leverages the advantage of neighbor alignment and performs better under misaligned conditions. Note that `FusionAD-R' and `UniAD-R' denote the variants of FusionAD and UniAD, respectively, trained with the RoarNet-based data augmentation strategy~\cite{shin2019roarnet}.
 }
    \label{fig:misalignment_performance}
\end{figure*}

\begin{table*}[htp]
\centering
  \caption[]{Effect of the hyperparameters $K_{\text{graph}}$ for feature misalignment. We analyze the effect of hyperparameter $K_{\text{graph}}$ in LocalAlign-v2 module for feature alignment under \textcolor{red}{noisy} misalignment settings on the nuScenes validation set. `LT(ms)' represents latency. All latency measurements are conducted on the same workstation with an A100 GPU.}
  \renewcommand\arraystretch{1}
  \tabcolsep=1.6mm %%%%%%%%%
  \resizebox{\linewidth}{!}{
  %\begin{tabular*}{\linewidth} {@{}@{\extracolsep{\fill}}!{\color{white}\vline}l|c|c|c|c|c|c|c|c|c|c|c|c @{}}
\begin{tabular}{ccc|ccc|ccc|ccc|ccc|ccc}
    \toprule
\multicolumn{3}{c|}{Baseline~\cite{bevfusion-mit}}        
&\multicolumn{3}{c|}{$K_{\text{graph}}=5$}    
&\multicolumn{3}{c|}{$K_{\text{graph}}=8$} 
&\multicolumn{3}{c|}{$K_{\text{graph}}=12$} 
&\multicolumn{3}{c|}{$K_{\text{graph}}=16$} 
&\multicolumn{3}{c}{$K_{\text{graph}}=25$}  \\ 

mAP&NDS&LT 
&mAP&NDS&LT
&mAP&NDS&LT 
&mAP&NDS&LT 
&mAP&NDS&LT 
&mAP&NDS&LT \\
\midrule

60.8& 65.7& \textbf{132.9}&     67.1& 70.9& 138.2&    \textbf{70.1}& \textbf{72.9}& 140.9&    69.8& 72.2& 143.4&               
 68.8& 70.5& 145.3&    67.1& 69.9& 150.0
\\ 

\bottomrule
\end{tabular} }

\label{tab_nuscenes_K}
\end{table*}

\begin{table*}[]
\centering
\caption[]{\textcolor{black}{Unified ablation study of LocalAlign-v2 (Query) and GlobalAlign-v2 (Diffusion) on UniAD \cite{uniad} across multiple autonomous driving tasks, including tracking, mapping, motion forecasting, occupancy prediction, and planning. ``↑" indicates better performance with higher metrics, while ``↓" indicates better performance with lower metrics.}}
\renewcommand\arraystretch{1}
\tabcolsep=0.3mm %%%%%%%%%
\resizebox{\linewidth}{!}{
\begin{tabular}{cc|cc|cc|ccc|cccc|cc}
\toprule
 \multirow{2}{*}{ LocalAlign-v2 (Query) } & \multirow{2}{*}{ GlobalAlign-v2 (Diffusion) }& \multicolumn{2}{c|}{Tracking} & \multicolumn{2}{c|}{Mapping} &\multicolumn{3}{c|}{Motion Forecasting} & \multicolumn{4}{c|}{Occupancy} & \multicolumn{2}{c}{Planning} \\
                       & & AMOTA  ↑         & AMOTP  ↓        & IoU-Lanes  ↑  &IoU-Drivable  ↑    & minADE  ↓       & minFDE  ↓    &  MR  ↓    & IoU-n.  ↑ & IoU-f.  ↑ & VPQ-n.  ↑ & VPQ-f. ↑    & avg.L2↓      & avg.Col↓      \\

\midrule
& &21.5	& 1.566 & 25.1 & 61.6 &0.84&	1.32	& 26.1 & 49.2	& 26.4 & 38.1 &	21.5 &	1.23 & 0.48\\
\checkmark &  & 40.3&1.208&31.4&69.1&0.56&0.81&11.7&61.4&41.5&63.1&44.3&0.87&0.26\\
 &\checkmark & 31.9&1.337&28.7&65.9&0.63&0.93&15.4&58.1&37.6&54.9&37.2&0.95&0.33\\

\checkmark  &\checkmark  & 44.5&1.102& 31.4&69.1&0.44&0.63&8.8&66.9&48.1&62.5&48.2&0.73&0.22

\\

\bottomrule
\end{tabular}
\label{tab_nuscenes_e2e_misalignment_ablation}
}
\end{table*}

\begin{table*}[htp]
\scriptsize
\centering
  \caption[]{Robustness to weather conditions, different ego distances, different sizes on nuScenes~\cite{nuscenes} clean validation set. It is important to note that the evaluation metric is mAP (\%). }
  \renewcommand\arraystretch{1}
    \tabcolsep=3.5mm %%%%%%%%%
  \resizebox{\linewidth}{!}{
  %\begin{tabular*}{\linewidth} {@{}@{\extracolsep{\fill}}!{\color{white}\vline}l|c|c|c|c|c|c|c|c|c|c|c @{}}
  \begin{tabular}{l|cccc|ccc| ccc }
    \toprule
  \multirow{2}{*}{Method} &\multicolumn{4}{c|}{  Different Weather Conditions }  &\multicolumn{3}{c|}{Different Ego Distances}&\multicolumn{3}{c}{Different Object Sizes}  \\
          & Sunny  & Rainy  & Day  & Night & Near & Middle  & Far & Small & Moderate & Large  \\ 
\midrule

Baseline~\cite{bevfusion-mit}& 68.2& 69.9& 68.5& 42.8 & 79.4& 64.9& 40.0 & 50.3 &58.7 &64.0 \\ 
\rowcolor{pink!10}\textbf{GraphBEV++ (LSS)}  &70.7&70.4& 69.8& 45.2&79.9& 65.6& 42.2&55.5& 59.5& 64.6\\ 
 &\textit{\fontsize{6}{0}\selectfont\textcolor{red}{+2.5}}
 &\textit{\fontsize{6}{0}\selectfont\textcolor{red}{+0.5}}
 &\textit{\fontsize{6}{0}\selectfont\textcolor{red}{+1.3}}
 &\textit{\fontsize{6}{0}\selectfont\textcolor{red}{+2.4}}
 &\textit{\fontsize{6}{0}\selectfont\textcolor{red}{+0.5}} 
 &\textit{\fontsize{6}{0}\selectfont\textcolor{red}{+0.7}} 
 &\textit{\fontsize{6}{0}\selectfont\textcolor{red}{+2.2}}
 &\textit{\fontsize{6}{0}\selectfont\textcolor{red}{+5.2}}
 &\textit{\fontsize{6}{0}\selectfont\textcolor{red}{+0.8}}
 &\textit{\fontsize{6}{0}\selectfont\textcolor{red}{+0.6}}

 \\

% \midrule
% \midrule
% Baseline~\cite{bevfusion-mit}& \multirow{3}{*}{\textcolor{red}{Noisy}}& 60.8 & 65.7 & 83.1 & 50.3 & 26.5& 66.4 & 38.0 & 65.0 & 64.9 & 52.8 \\ 
% \textbf{GraphBEV++}&   &69.1&72.0& 88.1& 63.5& 30.0& 75.1& 42.7& 75.3& 79.8& 64.9\\ 
\bottomrule
\end{tabular} }

\label{nuscenes_robustness}
\end{table*}

% \begin{table*}[htp]
% \scriptsize
% \centering
%   \caption[]{Main results for multi-tasks, end-to-end learning. * denotes evaluation using checkpoints from official implementation.}
%   \renewcommand\arraystretch{1}
%   \tabcolsep=0.5mm %%%%%%%%%
%   \resizebox{\linewidth}{!}{
%   %\begin{tabular*}{\linewidth} {@{}@{\extracolsep{\fill}}!{\color{white}\vline}l|c|c|c|c|c|c|c|c|c|c|c|c @{}}
%   \begin{tabular}{l|cc|ccc|cc| cccc | cccc | ccc }
%     \toprule
% \multirow{2}{*}{Method} &     \multicolumn{2}{c|}{Detection} & \multicolumn{3}{c|}{Tracking} & \multicolumn{2}{c|}{Mapping} & \multicolumn{4}{c|}{Prediction} & \multicolumn{4}{c|}{Occupancy} &
% \multicolumn{3}{c}{Planning} \\ 
% &mAP↑&NDS↑    
% &AMOTA↑& AMOTP↓&IDS↓ %tracking 
% & IoU-Lane↑& IoU-D↑ %Online mapping          
% & minADE↓& minFDE↓& MR↓& EPA↑      % Motion forecasting
% & VPQ-n↑& VPQ-f↑& IoU-n↑& IoU-f↑&        %Occupancy prediction   
% L2(m)↓& . Rate(\%)↓ %Planning
% \\

% \midrule

%  % \\
% \bottomrule
% \end{tabular} }

% \label{tab_nuscenes_bevmap}
% \end{table*}

\subsubsection{3D Object Detection}
\noindent \textbf{NuScenes (val).}
We first compare GraphBEV++ with recent SOTA methods on the nuScenes validation set for 3D object detection, as shown in Table~\ref{tab_nuscenes_val_test}. GraphBEV++ achieves competitive performance under both LSS-based and query-based BEV paradigms. Specifically, GraphBEV++ (LSS) obtains 70.7% mAP and 73.2% NDS, outperforming the original GraphBEV by +0.6% mAP and +0.3% NDS. Moreover, GraphBEV++ (Query) further improves the performance to 71.4% mAP and 73.4% NDS, achieving the best mAP among all compared methods on the validation set. These results demonstrate that the proposed alignment modules are effective for both explicit LSS-based BEV fusion and implicit query-based BEV representations.

Among the multi-modal methods, point-level approaches such as PointPainting~\cite{pointpainting}, PointAugmenting~\cite{wang2021pointaugmenting}, and MVP~\cite{mvp} directly augment LiDAR points with image features, but they cannot explicitly handle feature misalignment in the BEV space. Feature-level methods, including GraphAlign~\cite{graphalign}, AutoAlignV2~\cite{autoalignv2}, TransFusion~\cite{Transfusion}, and DeepInteraction~\cite{DeepInteraction}, alleviate cross-modal misalignment before BEV construction, but they do not directly address the misalignment introduced during the camera-to-BEV transformation. In contrast, GraphBEV++ explicitly models both local projection-induced misalignment and global BEV-level misalignment, leading to more robust multi-modal fusion.

It is also noteworthy that GraphBEV++ shows clear advantages on several challenging object categories. For example, GraphBEV++ (Query) achieves the best performance on Car, Truck, Barrier, Motor., and Ped., while GraphBEV++ (LSS) achieves the best result on T.C. These improvements indicate that feature alignment is particularly beneficial for categories that are sensitive to geometric projection errors and local feature inconsistency. Furthermore, compared with BEVFormer-M, GraphBEV++ (Query) improves mAP from 71.2\% to 71.4\% and NDS from 73.2\% to 73.4\%, confirming the compatibility of the proposed method with query-based BEV architectures.

\noindent \textbf{NuScenes (test).}
As shown in Table~\ref{tab_nuscenes_val_test}, GraphBEV++ (LSS) also achieves strong performance on the nuScenes \textcolor{red}{test} set. Compared with BEVFusion-MIT~\cite{bevfusion-mit}, GraphBEV++ improves mAP from 70.2\% to 72.0\% and NDS from 72.9\% to 74.0\%, yielding gains of +1.8\% mAP and +1.1\% NDS. Compared with the conference version GraphBEV, GraphBEV++ improves mAP by +0.3\% and NDS by +0.4\%, demonstrating that the upgraded LocalAlign-v2 and GlobalAlign-v2 modules further enhance the detection capability.

Although the nuScenes test set contains relatively clean sensor calibration, feature misalignment can still occur due to projection noise, calibration deviation, and depth estimation errors, as illustrated in Fig.~\ref{fig:motivation}(a). By incorporating neighborhood-aware depth features through KD-Tree search, GraphBEV++ effectively improves the robustness of camera-to-BEV transformation. In particular, GraphBEV++ achieves the best results on Car, C.V., Ped., and T.C., and obtains second-best performance on Bus and Trailer. These gains are especially meaningful for small or structurally complex objects, whose projected image features are more sensitive to slight LiDAR-camera misalignment. Overall, the test-set results verify that GraphBEV++ provides a robust and generalizable feature alignment solution for BEV-based multi-modal 3D object detection.

\noindent \textbf{NuScenes-C.}
As shown in Table~\ref{tab_nuscenes_C_misalignment}, we conduct comparative experiments under the nuScenes-C misalignment noise setting, including SparseFusion, BEVFusion, BEVFormer, and GraphBEV. BEVFusion suffers a performance drop of 11.2\% in mAP and 8.2\% in NDS when transitioning from the clean to the noisy setting. Similarly, BEVFormer exhibits a drop of 10.8\% in mAP and 9.2\% in NDS, highlighting the challenge of feature misalignment.  Moreover, GraphBEV++ (LSS) and GraphBEV++ (Query) further reduce the performance degradation to 2.0\% and 3.2\% mAP, respectively. These results verify that GraphBEV++ not only achieves better performance under clean conditions, but also maintains high robustness in noisy, real-world scenarios.
\textcolor{black}{Although GraphBEV++ (LSS) shows a slightly larger relative mAP drop than GraphBEV, it consistently achieves higher performance under noisy conditions, improving mAP from 69.1 to 69.3 and NDS from 72.0 to 72.3. These results indicate that GraphBEV++ enhances the performance ceiling while preserving strong robustness against feature misalignment.}

As shown in Figure~\ref{fig:misalignment_performance}, we introduce varying levels of misalignment noise into the BEV perception module to evaluate its impact on end-to-end autonomous driving performance. Both FusionAD and UniAD are tested under two training settings: using clean data and employing misalignment-based data augmentation (RoarNet). While the augmented models show improved robustness under severe noise conditions, their performance on clean test data noticeably degrades. This reveals a trade-off: random data augmentation helps models generalize under noise but lacks the capacity to accurately simulate diverse real-world misalignment scenarios. In contrast, our GraphBEV++ addresses the misalignment problem at the algorithmic level, leading to a more principled and effective solution.

\noindent \textbf{NuScenes (Radar and Camera).}
As shown in Table~\ref{tab_nus_C_noise_radar}, we compare GraphBEV++ with existing radar-camera fusion methods under both clean and noisy settings. Compared with HVDetFusion~\cite{lei2023hvdetfusion}, GraphBEV++ improves mAP from 45.1\% to 45.9\% under clean settings and from 40.1\% to 45.0\% under noisy settings. Notably, the improvement under noisy conditions (+4.9\%) is significantly larger than that under clean conditions (+0.8\%), demonstrating the robustness of the proposed alignment strategy against sensor misalignment.
Furthermore, GraphBEV++ achieves competitive performance compared with recent radar-camera fusion methods, such as CRN~\cite{kim2023crn} and HyDRa~\cite{wolters2025unleashing}. When integrated into the state-of-the-art RaCFormer~\cite{chu2025racformer}, GraphBEV++ further improves mAP from 54.1\% to 54.8\% under clean settings and from 50.6\% to 54.2\% under noisy settings. These results indicate that the proposed LocalAlign-v2 generalizes well to sparse radar features and can consistently improve the robustness of different radar-camera fusion architectures.

\noindent \textbf{Waymo and Waymo-C.}
As shown in Table~\ref{tab_waymo_C_noise}, GraphBEV++ consistently surpasses UniTR~\cite{wang2023unitr} across all evaluation metrics under both clean and noisy settings on the Waymo-C benchmark (10\% training data). In the noisy scenario, it achieves substantial improvements in L2 mAP and mAPH (+4.69\%/+4.73\%), demonstrating superior robustness to sensor misalignment noise.

\noindent \textbf{Argoverse 2.}
As shown in Table~\ref{tab_Argoverse2_val}, GraphBEV++ achieves the best overall performance with 46.7\% mAP, outperforming the previous GraphBEV by +0.6\% mAP and surpassing all existing LiDAR-only and multi-modal methods. These results demonstrate the effectiveness of the proposed LocalAlign-v2 and GlobalAlign-v2 modules in improving feature alignment and object localization.
Furthermore, GraphBEV++ consistently improves performance across almost all categories. In particular, noticeable gains are observed on small, sparse, and long-tail categories, such as Bicyclist (43.0\%), Motorcycle (53.3\%), School Bus (52.8\%), Vehicle Trailer (37.2\%), and Large Vehicle (36.4\%). Since these categories are more sensitive to feature misalignment and localization errors, the improvements indicate that GraphBEV++ can better preserve fine-grained geometric details during multi-modal fusion. The strong performance on the Argoverse2 benchmark further validates the generalization ability of GraphBEV++ under large-scale and long-tail autonomous driving scenarios.

\subsubsection{BEV Map Segmentation}
To evaluate 3D object detection, we also assess the generalization capability in the BEV Map Segmentation (semantic segmentation) tasks on the nuScenes validation set, as shown in Table~\ref{tab_nuscenes_bevmap}. Following the same training strategy as the baseline BEVFusion, we have conducted evaluations within the [-50m, 50m]×[-50m, 50m] region around an ego car for each frame. We then report Intersection-over-Union (IoU) scores for drivable area, pedestrian crossing, walkway, stop line, car park, and divider. Significant improvements are observed for drivable area, pedestrian crossing, walkway, stop line, and car park, with only a minor decrease for divider. Overall, our GraphBEV++ demonstrates not only significant performance in 3D object detection but also strong generalization capability in BEV Map Segmentation.

\subsubsection{3D  Occupancy Prediction}
\textcolor{black}{To further validate the generality of GraphBEV++, we extend our feature alignment framework to the 3D semantic occupancy prediction task by integrating the GraphBEV++ (Query) variant into GaussianFormer-T. As shown in Table~\ref{tab_nusc_occ}, GraphBEV++ consistently outperforms the baseline under both clean and noisy settings. Specifically, GraphBEV++ improves IoU/mIoU from 31.34/20.42 to 32.47/21.90 under the clean setting and from 27.86/17.63 to 29.41/19.37 under the noisy setting. Moreover, the performance degradation from clean to noisy conditions is reduced from 3.48 to 3.06 IoU and from 2.79 to 2.53 mIoU. These results demonstrate that the proposed LocalAlign-v2 and GlobalAlign-v2 modules effectively improve multi-view feature alignment and can generalize beyond 3D object detection to robust 3D occupancy prediction.
}

\subsubsection{End-to-End Autonomous Driving}
To evaluate end-to-end autonomous driving, we follow UniAD~\cite{uniad}, reporting the performance of each task (perception, prediction, and planning) sequentially. We compare the performance of our GraphBEV++ (Query) and GraphBEV++ (LSS)  with SOTA methods on the nuScenes validation set.

\noindent \textbf{Perception Results.}
For multi-object tracking in Table~\ref{tab_nuscenes_multi_object_tracking}, our GraphBEV++ (Query) achieves SOTA performance compared to end-to-end autonomous driving methods like UniAD~\cite{uniad}, SparseDrive~\cite{sun2024sparsedrive}, and FusionAD~\cite{fusionad}. As a multi-modal end-to-end approach, our GraphBEV++ (Query) achieves 51.1\% AMOTA(\%)↑ and 1.022m AMOTP(m)↓, significantly surpassing the baseline UniAD (35.9 \% AMOTA and 1.320m AMOTP) as well as SparseDrive (38.6\% AMOTA and 1.254m AMOTP) and FusionAD (50.1\% AMOTA and 1.065m AMOTP). For online mapping in Table~\ref{tab_nuscenes_online_mapping}, our GraphBEV++ (Query) performs well on segmenting lanes (\textbf{+5.8 IoU(\%)} compared to UniAD~\cite{uniad}), which is crucial for downstream agentroad interaction in the motion module. It is noteworthy that another version of our method, GraphBEV++ (LSS), performs similarly to GraphBEV++ (Query) on the aforementioned tasks.
We have shown significant performance improvements in the perception module, attributed to our effective utilization of LiDAR, which offers new prospects for multi-modal end-to-end systems.

\noindent \textbf{Prediction Results.}
As shown in Table~\ref{tab_nuscenes_motion_forecasting}, GraphBEV++ achieves the best overall motion forecasting performance. GraphBEV++ (Query) obtains the lowest minADE (0.38,m), minFDE (0.52,m), and MR (7.7\%), outperforming recent methods such as SparseDrive and MomAD. In addition, GraphBEV++ achieves the highest EPA score of 64.7\%, indicating superior trajectory quality and prediction reliability. Compared with GraphBEV++ (LSS), the query-based variant further improves forecasting accuracy, demonstrating the effectiveness of the proposed alignment strategy for query-based BEV representations. These results verify that accurate multi-modal feature alignment benefits not only perception but also downstream motion forecasting.

\noindent \textbf{Planning Results (nuScenes).}
In the planning results shown in Table~\ref{tab_nuscenes_planning}, our GraphBEV++ (Query) 
 and GraphBEV++ (LSS) achieve superior performance compared to end-to-end autonomous driving methods such as ST-P3~\cite{ST_P3}, GPT-Driver~\cite{gpt_driver}, VAD~\cite{jiang2023vad}, and UniAD~\cite{uniad}. GraphBEV++ (Query) reaches SOTA results in collision rate, while GraphBEV++ (LSS) achieves SOTA results in L2 error. By leveraging multi-modal information from LiDAR and camera data, our GraphBEV++ (Query) reduces the planning L2 error and collision rate by 32.0\% and 58.1\%, respectively, compared to UniAD~\cite{uniad}, based on average values for the planning horizon.

 \noindent \textbf{Planning Results (Bench2Drive).}
As shown in Table~\ref{tab_b2d}, we conduct experiments on the Bench2Drive closed-loop benchmark, using the MomAD (VAD version) as the baseline. Our GraphBEV++ achieves consistent improvements on this benchmark, demonstrating its potential for deployment in closed-loop autonomous driving systems. It is worth noting that the feature alignment in this closed-loop setting is ideal; this experiment primarily serves to validate the practical applicability of our method.

 \noindent \textbf{Planning Results (NAVSIM).}
As shown in Table~\ref{tab_navsim}, we conduct end-to-end evaluations on the closed-loop NAVSIM benchmark, using WoTE~\cite{wote} as the baseline. Compared methods include UniAD, LTF, PARA-Drive, VADv2, Hydra-MDP, and DiffusionDrive. Notably, WoTE achieves a PDMS score of 88.3, while our GraphBEV++ (Query) reaches 88.7, indicating that our method also brings performance gains on NAVSIM.
This demonstrates the effectiveness and generalizability of GraphBEV++ in closed-loop end-to-end driving scenarios.

\subsection{Ablation Study}
\subsubsection{Efficiency Analysis of GraphBEV++}
\textcolor{black}{Table~\ref{tab_efficiency_comparison} compares the efficiency of GraphBEV++ with representative multi-modal perception methods. GraphBEV++ (LSS) achieves a favorable accuracy--efficiency trade-off, reducing FLOPs from 535.7G to 457.6G compared with GraphBEV while maintaining a similar inference speed (7.2 FPS vs. 7.1 FPS). Moreover, it is substantially more efficient than RoboFusion and DeepInteraction. These results demonstrate that the proposed alignment modules improve robustness and feature alignment quality with only marginal computational overhead.}

\subsubsection{Roles of Different Modules in GraphBEV++}
To analyze the impact of misalignment, we conduct comparative experiments between our GraphBEV++ (LSS) and BEVFusion~\cite{bevfusion-mit}. It is noteworthy that in Table~\ref{tab_nuscenes_misalignment_lss}, we introduce misalignment into the nuScenes validation set, rather than in the training and testing sets, following Ref.~\cite{zhujun_benchmarking}. We train on the clean nuScenes~\cite{nuscenes} training dataset and evaluate performance under both clean and noisy misalignment conditions. In the clean setting, GraphBEV++ (LSS) outperforms BEVFusion~\cite{bevfusion-mit} significantly, while in the noisy setting, the performance improvement is substantial. Furthermore, it is evident that BEVFusion~\cite{bevfusion-mit} exhibits a significant decrease in metrics such as mAP and NDS when transitioning from clean to noisy conditions, whereas GraphBEV++ (LSS) demonstrates a small decrease in performance metrics. Notably, adding the LocalAlign-v2 (LSS) or GlobalAlign-v2 (Deformable) module to BEVFusion~\cite{bevfusion-mit} has minimal impact on latency compared to BEVFusion~\cite{bevfusion-mit} alone, and the latency is lower than that of TransFusion~\cite{Transfusion}. When only the LocalAlign-v2 (LSS) module is added to BEVFusion~\cite{bevfusion-mit} and the KD-Tree algorithm is used to build proximity relationships, significant enhancements are observed in both clean and noisy misalignment settings by fusing projected depth with neighbor depth to prevent feature misalignment. Adding only the GlobalAlign-v2 (Deformable) module to BEVFusion~\cite{bevfusion-mit} also leads to noticeable improvements. Particularly, the simultaneous addition of the LocalAlign-v2 (LSS) and GlobalAlign-v2 (Deformable) modules exhibits strong performance in both clean and noisy settings.

We further evaluate the effectiveness of \textbf{LocalAlign-v2 (Query)} and \textbf{GlobalAlign-v2 (Diffusion)} on the BEVFormer-M baseline. As shown in Table~\ref{tab_nuscenes_misalignment_query}, the baseline achieves 63.2 mAP and 66.3 NDS on the nuScenes-C benchmark. Incorporating LocalAlign-v2 (Query) improves performance by 3.1\% mAP and 2.7\% NDS, while GlobalAlign-v2 (Diffusion) leads to a gain of 3.0\% mAP and 1.8\% NDS. The full GraphBEV++ (Query) model achieves 69.1 mAP and 71.2 NDS, outperforming the baseline by 5.9\% and 4.9\% respectively. These results validate the effectiveness of our method in addressing feature misalignment in query-based BEV frameworks. 

\textcolor{black}{
Table~\ref{tab_diffusion_steps} presents the trade-off between diffusion steps, detection performance, and inference speed. Increasing $T$ consistently improves alignment quality, with mAP rising from 67.8 to 69.1 and NDS from 69.6 to 71.2 when $T$ increases from 1 to 4. Nevertheless, the performance gain becomes saturated beyond $T=4$, while the inference speed decreases from 4.90 FPS to 4.57 FPS. Therefore, a lightweight diffusion configuration with $T=4$ is adopted throughout the paper, providing an effective balance between alignment performance and runtime efficiency.
}

\subsubsection{Effect of Depth Noise on GraphBEV++}

As shown in Table~\ref{tab_nuscenes_bevdepth}, we follow the noise injection protocol of BEVDepth~\cite{bevdepth}, applying both Gaussian noise and one-hot random noise (which directly replaces depth values) to simulate degraded depth inputs. These perturbations significantly impact our method, as GraphBEV++ relies heavily on accurate depth information. Our approach is specifically designed to address the misalignment issues caused by inaccurate point cloud projection in BEVDepth. The variant labeled as \textit{learned} refers to our GraphBEV++ framework, in which depth information is not only derived from projected LiDAR points but also encoded through a learnable depth feature representation. Overall, our method demonstrates strong robustness against depth noise.

\subsubsection{Effect of GraphBEV++ in End-to-End Autonomous Driving for Feature Misalignment}
As shown in Table~\ref{tab_nuscenes_e2e_misalignment}, we analyze the robustness of GraphBEV++ (LSS), GraphBEV++ (Query), UniAD~\cite{uniad}, and FusionAD~\cite{fusionad} in end-to-end tasks under varying severity levels of misalignment conditions. The misalignment issue directly impacts the performance of models in various end-to-end autonomous driving tasks (Tracking, Mapping, Motion Forecasting, Occupancy and Planning), which highlights the importance of addressing misalignment to maintain performance in end-to-end autonomous driving tasks. Meanwhile, compared to mono modal method UniAD~\cite{uniad}, multi-modal methods FusionAD and GraphBEV++ (Query) are more robust when facing the issues of misalignment. This is due to the effective utilization of the complementary information from different modalities in multi-modal methods. However, although multimodal methods are robust to misalignment issues, they are still affected by modal misalignment. Notably, by utilizing neighborhood information as guidance and correction for feature alignment, GraphBEV++ (Query) and GraphBEV++ (LSS) effectively achieve modal alignment from both local and global perspectives. This enables GraphBEV++ to significantly outperform existing methods like UniAD~\cite{uniad} and FusionAD~\cite{fusionad} under feature misalignment conditions. Noteworthy, by utilizing neighborhood information as a bridge for alignment between different modalities and achieving modality alignment from a global perspective, GraphBEV++ significantly outperforms the SOTA methods UniAD~\cite{uniad} and FusionAD~\cite{fusionad} under varying levels of misalignment severity. Overall, our GraphBEV++ (LSS) and GraphBEV++ (Query) improve performance in misalignment scenarios through accurate feature alignment.

\subsubsection{Effect of the Hyperparameters $K_{\text{graph}}$ for Feature Misalignment}

As shown in Table~\ref{tab_nuscenes_K}, to analyze the impact of the hyperparameter $K_{\text{graph}}$ in the LocalAlign-v2 module on feature misalignment, we have studied its effects under noisy misalignment settings on the nuScenes validation set. $K_{\text{graph}}$, which is the number of nearest depths for LiDAR-to-camera projected depth in the LocalAlign-v2 module, influences the expressive capability of neighboring depth features. It is observed that our GraphBEV++ achieves optimal overall performance when $K_{\text{graph}}$ is set to 8. Therefore, selecting an appropriate value for $K_{\text{graph}}$ is crucial and may vary across different datasets. Furthermore, despite significant fluctuations in mAP resulting from changes in $K_{\text{graph}}$, the overall performance still surpasses that of BEVFusion.

\subsubsection{Effect of LocalAlign-v2 and GlobalAlign-v2 Across Multiple Autonomous Driving Tasks}

\textcolor{black}{As shown in Table~\ref{tab_nuscenes_e2e_misalignment_ablation}, we conduct a unified ablation study on UniAD to investigate the contributions of LocalAlign-v2 (Query) and GlobalAlign-v2 (Diffusion) across multiple autonomous driving tasks, including tracking, mapping, motion forecasting, occupancy prediction, and planning.
It can be observed that both alignment modules consistently improve performance over the UniAD baseline. Specifically, introducing only LocalAlign-v2 yields substantial gains across all tasks, improving AMOTA from 21.5 to 40.3, IoU-Lanes from 25.1 to 31.4, and reducing the planning error (avg.L2) from 1.23 to 0.87. In comparison, GlobalAlign-v2 alone also provides consistent improvements, increasing AMOTA to 31.9 and reducing avg.L2 to 0.95. These results demonstrate that both local and global alignment errors negatively affect the entire autonomous driving pipeline.
Furthermore, combining LocalAlign-v2 and GlobalAlign-v2 achieves the best overall performance on nearly all metrics. Compared with the UniAD baseline, the complete GraphBEV++ improves AMOTA by +23.0, IoU-f by +21.7, and VPQ-f by +26.7, while reducing minADE from 0.84 to 0.44, minFDE from 1.32 to 0.63, and avg.Col from 0.48 to 0.22. These results indicate that the two modules are highly complementary, where LocalAlign-v2 primarily addresses local feature correspondence while GlobalAlign-v2 further mitigates global feature misalignment. Their combination provides the most effective alignment solution and consistently benefits all downstream autonomous driving tasks.}

\subsection{Robustness Study}

\subsubsection{Robustness to Weather Conditions}
Adverse weather amplifies sensor degradation and calibration uncertainty, thereby intensifying feature misalignment between LiDAR and camera modalities. Therefore, evaluating performance under different weather conditions serves as an indirect yet effective way to assess the model’s ability to address feature misalignment.
As shown in Table~\ref{nuscenes_robustness}, we present a robustness analysis of our GraphBEV++ with respect to different weather conditions. Various weather conditions influence 3D object detection tasks. Following the approach of BEVFusion~\cite{bevfusion-mit}, we partition the scenes in the validation set into sunny, rainy, day, and night conditions. We outperform BEVFusion~\cite{bevfusion-mit} under different weather conditions, especially in night scenes. Overall, our GraphBEV++ improves performance in sunny weather through accurate feature alignment and enhances performance in adverse weather conditions.
% jfy we partitioned -> partition
\subsubsection{Robustness to Ego Distances and Object Sizes}
As shown in Table~\ref{nuscenes_robustness}, we analyze the impact of different ego distances and object sizes on the performance of GraphBEV++. We categorize annotation and prediction ego distances into three groups: Near (0-20m), Middle (20-30m), and Far (>30m), and summarize the size distributions for each category, defining three equal-proportion size levels: Small, Moderate, and Large. It is evident that GraphBEV++ demonstrates significant performance improvements for distant and small objects. Compared to BEVFusion~\cite{bevfusion-mit}, our GraphBEV++ consistently enhances performance across all ego distances and object sizes, further narrowing the performance gaps. Overall, our GraphBEV++ exhibits great robustness to changes in ego distances and object sizes.
% jfy Futhermore -> Furthermore 拼错
% jfy summarized -> summarize

% \subsection{Extension to the end-to-end planning}

\section{Conclusion} \label{sec:conclusion}

In this work, we propose GraphBEV++, a robust fusion framework designed to address the feature misalignment problem in autonomous driving. Our framework is applicable to both 3D object detection and end-to-end autonomous driving tasks.
To tackle local feature misalignment caused by inaccurately projected depth from LiDAR, we introduce the LocalAlign-v2 module, which comes in two variants: LSS-based and Query-based. This design not only mitigates the misalignment issues inherent in LSS-dependent methods such as BEVFusion but also addresses the projection inaccuracies of 3D reference points in methods like BEVFormer, by incorporating neighbor-aware depth features via graph matching.
To further handle global-level misalignment between LiDAR and camera BEV features during fusion, we propose the GlobalAlign-v2 module, featuring two variants: Deformable-based and Diffusion-based. These variants resolve global misalignment through learned spatial offsets and diffusion-based feature alignment, respectively.
Overall, GraphBEV++ provides a unified and principled solution to both local and global misalignment, significantly enhancing the robustness of multi-modal BEV perception under real-world deployment scenarios.

\section*{Acknowledgments}
This work was supported by the National Natural Science Foundation of China (NSFC) under Grants No. 62536001 (Key Program) and No. 62576026.

\section*{Data Availability} The datasets generated and/or analyzed during this study are publicly available in the original repositories: nuScenes ~\cite{nuscenes}  \url{https://www.nuscenes.org} and Argoverse 2 ~\cite{Argoverse2}  \url{https://www.argoverse.org/av2.html}.

\bibliographystyle{spmpsci}
\bibliography{sample}% common bib file

@String(CVPR  = {IEEE Conf. Comput. Vis. Pattern Recog.})

@String(ICCV  = {Int. Conf. Comput. Vis.})

@String(ECCV  = {Eur. Conf. Comput. Vis.})

@String(NeurIPS = {Adv. Neural Inform. Process. Syst.})

@String(AAAI  = {AAAI})

@String(CVPR  = {CVPR})

@String(ICCV  = {ICCV})

@String(ECCV  = {ECCV})

@String(NeurIPS = {NeurIPS})

@inproceedings{murez2020atlas,
  title={Atlas: End-to-end 3d scene reconstruction from posed images},
  author={Murez, Zak and Van As, Tarrence and Bartolozzi, James and Sinha, Ayan and Badrinarayanan, Vijay and Rabinovich, Andrew},
  booktitle={European conference on computer vision},
  pages={414--431},
  year={2020},
  organization={Springer}
}

@inproceedings{li2023voxformer,
  title={Voxformer: Sparse voxel transformer for camera-based 3d semantic scene completion},
  author={Li, Yiming and Yu, Zhiding and Choy, Christopher and Xiao, Chaowei and Alvarez, Jose M and Fidler, Sanja and Feng, Chen and Anandkumar, Anima},
  booktitle={Proceedings of the IEEE/CVF conference on computer vision and pattern recognition},
  pages={9087--9098},
  year={2023}
}

@article{liao2025cot,
  title={Cot-drive: Efficient motion forecasting for autonomous driving with llms and chain-of-thought prompting},
  author={Liao, Haicheng and Kong, Hanlin and Wang, Bonan and Wang, Chengyue and Wang, Kanye Ye and He, Zhengbing and Xu, Chengzhong and Li, Zhenning},
  journal={IEEE Transactions on Artificial Intelligence},
  number={2},
  pages={625--641},
  year={2025},
  publisher={IEEE}
}

@article{gan2024comprehensive,
  title={A comprehensive framework for 3d occupancy estimation in autonomous driving},
  author={Gan, Wanshui and Mo, Ningkai and Xu, Hongbin and Yokoya, Naoto},
  journal={IEEE Transactions on Intelligent Vehicles},
  year={2024},
  publisher={IEEE}
}

@inproceedings{wang2023openoccupancy,
  title={Openoccupancy: A large scale benchmark for surrounding semantic occupancy perception},
  author={Wang, Xiaofeng and Zhu, Zheng and Xu, Wenbo and Zhang, Yunpeng and Wei, Yi and Chi, Xu and Ye, Yun and Du, Dalong and Lu, Jiwen and Wang, Xingang},
  booktitle={Proceedings of the IEEE/CVF International Conference on Computer Vision},
  pages={17850--17859},
  year={2023}
}

@article{zuo2023pointocc,
  title={Pointocc: Cylindrical tri-perspective view for point-based 3d semantic occupancy prediction},
  author={Zuo, Sicheng and Zheng, Wenzhao and Huang, Yuanhui and Zhou, Jie and Lu, Jiwen},
  journal={arXiv preprint arXiv:2308.16896},
  year={2023}
}

@inproceedings{liu2024fully,
  title={Fully sparse 3d occupancy prediction},
  author={Liu, Haisong and Chen, Yang and Wang, Haiguang and Yang, Zetong and Li, Tianyu and Zeng, Jia and Chen, Li and Li, Hongyang and Wang, Limin},
  booktitle={European Conference on Computer Vision},
  pages={54--71},
  year={2024},
  organization={Springer}
}

@inproceedings{huang2024gaussianformer,
  title={Gaussianformer: Scene as gaussians for vision-based 3d semantic occupancy prediction},
  author={Huang, Yuanhui and Zheng, Wenzhao and Zhang, Yunpeng and Zhou, Jie and Lu, Jiwen},
  booktitle={European Conference on Computer Vision},
  pages={376--393},
  year={2024},
  organization={Springer}
}

@article{jia2025drivetransformer,
  title={Drivetransformer: Unified transformer for scalable end-to-end autonomous driving},
  author={Jia, Xiaosong and You, Junqi and Zhang, Zhiyuan and Yan, Junchi},
  journal={arXiv preprint arXiv:2503.07656},
  year={2025}
}

@inproceedings{chu2025racformer,
  title={Racformer: Towards high-quality 3d object detection via query-based radar-camera fusion},
  author={Chu, Xiaomeng and Deng, Jiajun and You, Guoliang and Duan, Yifan and Li, Houqiang and Zhang, Yanyong},
  booktitle={Proceedings of the IEEE/CVF Conference on Computer Vision and Pattern Recognition},
  pages={17081--17091},
  year={2025}
}

@article{liu2026driveworld,
  title={DriveWorld-VLA: Unified Latent-Space World Modeling with Vision-Language-Action for Autonomous Driving},
  author={Jia, Feiyang and Liu, Lin and Song, Ziying and Jia, Caiyan and Ye, Hangjun and Hao, Xiaoshuai and Chen, Long and others},
  journal={arXiv preprint arXiv:2602.06521},
  year={2026}
}

@inproceedings{huang2025gaussianformer,
  title={Gaussianformer-2: Probabilistic gaussian superposition for efficient 3d occupancy prediction},
  author={Huang, Yuanhui and Thammatadatrakoon, Amonnut and Zheng, Wenzhao and Zhang, Yunpeng and Du, Dalong and Lu, Jiwen},
  booktitle={Proceedings of the computer vision and pattern recognition conference},
  pages={27477--27486},
  year={2025}
}

@inproceedings{lv2026gau,
  title={Gau-occ: Geometry-completed gaussians for multi-modal 3d occupancy prediction},
  author={Lv, Chengxin and Li, Yihui and Yang, Hongyu and Wang, YunHong},
  booktitle={Proceedings of the IEEE/CVF Conference on Computer Vision and Pattern Recognition},
  pages={14198--14207},
  year={2026}
}

@inproceedings{zhang2023occformer,
  title={Occformer: Dual-path transformer for vision-based 3d semantic occupancy prediction},
  author={Zhang, Yunpeng and Zhu, Zheng and Du, Dalong},
  booktitle={Proceedings of the IEEE/CVF International Conference on Computer Vision},
  pages={9433--9443},
  year={2023}
}

@inproceedings{huang2023tri,
  title={Tri-perspective view for vision-based 3d semantic occupancy prediction},
  author={Huang, Yuanhui and Zheng, Wenzhao and Zhang, Yunpeng and Zhou, Jie and Lu, Jiwen},
  booktitle={Proceedings of the IEEE/CVF conference on computer vision and pattern recognition},
  pages={9223--9232},
  year={2023}
}

@inproceedings{cao2022monoscene,
  title={Monoscene: Monocular 3d semantic scene completion},
  author={Cao, Anh-Quan and De Charette, Raoul},
  booktitle={Proceedings of the IEEE/CVF Conference on Computer Vision and Pattern Recognition},
  pages={3991--4001},
  year={2022}
}

@article{tian2023occ3d,
  title={Occ3d: A large-scale 3d occupancy prediction benchmark for autonomous driving},
  author={Tian, Xiaoyu and Jiang, Tao and Yun, Longfei and Mao, Yucheng and Yang, Huitong and Wang, Yue and Wang, Yilun and Zhao, Hang},
  journal={Advances in Neural Information Processing Systems},
  volume={36},
  pages={64318--64330},
  year={2023}
}

@inproceedings{zuo2025gaussianworld,
  title={Gaussianworld: Gaussian world model for streaming 3d occupancy prediction},
  author={Zuo, Sicheng and Zheng, Wenzhao and Huang, Yuanhui and Zhou, Jie and Lu, Jiwen},
  booktitle={Proceedings of the Computer Vision and Pattern Recognition Conference},
  pages={6772--6781},
  year={2025}
}

@inproceedings{wei2023surroundocc,
  title={Surroundocc: Multi-camera 3d occupancy prediction for autonomous driving},
  author={Wei, Yi and Zhao, Linqing and Zheng, Wenzhao and Zhu, Zheng and Zhou, Jie and Lu, Jiwen},
  booktitle={Proceedings of the IEEE/CVF International Conference on Computer Vision},
  pages={21729--21740},
  year={2023}
}

@inproceedings{zheng2024occworld,
  title={Occworld: Learning a 3d occupancy world model for autonomous driving},
  author={Zheng, Wenzhao and Chen, Weiliang and Huang, Yuanhui and Zhang, Borui and Duan, Yueqi and Lu, Jiwen},
  booktitle={European conference on computer vision},
  pages={55--72},
  year={2024},
  organization={Springer}
}

@article{feng2020deep,
  title={Deep multi-modal object detection and semantic segmentation for autonomous driving: Datasets, methods, and challenges},
  author={Feng, Di and Haase-Sch{\"u}tz, Christian and Rosenbaum, Lars and Hertlein, Heinz and Glaeser, Claudius and Timm, Fabian and Wiesbeck, Werner and Dietmayer, Klaus},
  journal={IEEE Transactions on Intelligent Transportation Systems},
  volume={22},
  number={3},
  pages={1341--1360},
  year={2020},
  publisher={IEEE}
}

@inproceedings{feng2020leveraging,
  title={Leveraging uncertainties for deep multi-modal object detection in autonomous driving},
  author={Feng, Di and Cao, Yifan and Rosenbaum, Lars and Timm, Fabian and Dietmayer, Klaus},
  booktitle={2020 IEEE Intelligent Vehicles Symposium (IV)},
  pages={877--884},
  year={2020},
  organization={IEEE}
}

@inproceedings{shin2019roarnet,
  title={Roarnet: A robust 3d object detection based on region approximation refinement},
  author={Shin, Kiwoo and Kwon, Youngwook Paul and Tomizuka, Masayoshi},
  booktitle={2019 IEEE intelligent vehicles symposium (IV)},
  pages={2510--2515},
  year={2019},
  organization={IEEE}
}

@inproceedings{bijelic2020seeing,
  title={Seeing through fog without seeing fog: Deep multimodal sensor fusion in unseen adverse weather},
  author={Bijelic, Mario and Gruber, Tobias and Mannan, Fahim and Kraus, Florian and Ritter, Werner and Dietmayer, Klaus and Heide, Felix},
  booktitle={Proceedings of the IEEE/CVF Conference on Computer Vision and Pattern Recognition},
  pages={11682--11692},
  year={2020}
}

@inproceedings{drews2022deepfusion,
  title={Deepfusion: A robust and modular 3d object detector for lidars, cameras and radars},
  author={Drews, Florian and Feng, Di and Faion, Florian and Rosenbaum, Lars and Ulrich, Michael and Gl{\"a}ser, Claudius},
  booktitle={2022 IEEE/RSJ International Conference on Intelligent Robots and Systems (IROS)},
  pages={560--567},
  year={2022},
  organization={IEEE}
}

@inproceedings{pfeuffer2018optimal,
  title={Optimal sensor data fusion architecture for object detection in adverse weather conditions},
  author={Pfeuffer, Andreas and Dietmayer, Klaus},
  booktitle={2018 21st International Conference on Information Fusion (FUSION)},
  pages={1--8},
  year={2018},
  organization={IEEE}
}

@inproceedings{frossard2021strobe,
  title={Strobe: Streaming object detection from lidar packets},
  author={Frossard, Davi and Da Suo, Shun and Casas, Sergio and Tu, James and Urtasun, Raquel},
  booktitle={Conference on Robot Learning},
  pages={1174--1183},
  year={2021},
  organization={PMLR}
}

@article{gpt_driver,
  title={Gpt-driver: Learning to drive with gpt},
  author={Mao, Jiageng and Qian, Yuxi and Zhao, Hang and Wang, Yue},
  journal={arXiv preprint arXiv:2310.01415},
  year={2023}
}

@inproceedings{EO,
  title={Differentiable raycasting for self-supervised occupancy forecasting},
  author={Khurana, Tarasha and Hu, Peiyun and Dave, Achal and Ziglar, Jason and Held, David and Ramanan, Deva},
  booktitle={European Conference on Computer Vision},
  pages={353--369},
  year={2022},
  organization={Springer}
}

@inproceedings{FF,
  title={Safe local motion planning with self-supervised freespace forecasting},
  author={Hu, Peiyun and Huang, Aaron and Dolan, John and Held, David and Ramanan, Deva},
  booktitle={Proceedings of the IEEE/CVF Conference on Computer Vision and Pattern Recognition},
  pages={12732--12741},
  year={2021}
}

@inproceedings{NMP,
  title={End-to-end interpretable neural motion planner},
  author={Zeng, Wenyuan and Luo, Wenjie and Suo, Simon and Sadat, Abbas and Yang, Bin and Casas, Sergio and Urtasun, Raquel},
  booktitle={Proceedings of the IEEE/CVF Conference on Computer Vision and Pattern Recognition},
  pages={8660--8669},
  year={2019}
}

@inproceedings{ST_P3,
  title={St-p3: End-to-end vision-based autonomous driving via spatial-temporal feature learning},
  author={Hu, Shengchao and Chen, Li and Wu, Penghao and Li, Hongyang and Yan, Junchi and Tao, Dacheng},
  booktitle={European Conference on Computer Vision},
  pages={533--549},
  year={2022},
  organization={Springer}
}

@inproceedings{hu2021fiery,
  title={Fiery: Future instance prediction in bird's-eye view from surround monocular cameras},
  author={Hu, Anthony and Murez, Zak and Mohan, Nikhil and Dudas, Sof{\'\i}a and Hawke, Jeffrey and Badrinarayanan, Vijay and Cipolla, Roberto and Kendall, Alex},
  booktitle={Proceedings of the IEEE/CVF International Conference on Computer Vision},
  pages={15273--15282},
  year={2021}
}

@article{VPN,
  title={Cross-view semantic segmentation for sensing surroundings},
  author={Pan, Bowen and Sun, Jiankai and Leung, Ho Yin Tiga and Andonian, Alex and Zhou, Bolei},
  journal={IEEE Robotics and Automation Letters},
  volume={5},
  number={3},
  pages={4867--4873},
  year={2020},
  publisher={IEEE}
}

@inproceedings{zhang2022mutr3d,
  title={Mutr3d: A multi-camera tracking framework via 3d-to-2d queries},
  author={Zhang, Tianyuan and Chen, Xuanyao and Wang, Yue and Wang, Yilun and Zhao, Hang},
  booktitle={Proceedings of the IEEE/CVF Conference on Computer Vision and Pattern Recognition},
  pages={4537--4546},
  year={2022}
}

@article{chaabane2021deft,
  title={Deft: Detection embeddings for tracking},
  author={Chaabane, Mohamed and Zhang, Peter and Beveridge, J Ross and O'Hara, Stephen},
  journal={arXiv preprint arXiv:2102.02267},
  year={2021}
}

@article{QD3DT,
  title={Monocular quasi-dense 3d object tracking},
  author={Hu, Hou-Ning and Yang, Yung-Hsu and Fischer, Tobias and Darrell, Trevor and Yu, Fisher and Sun, Min},
  journal={IEEE Transactions on Pattern Analysis and Machine Intelligence},
  volume={45},
  number={2},
  pages={1992--2008},
  year={2022},
  publisher={IEEE}
}

@inproceedings{DQTrack,
  title={End-to-end 3d tracking with decoupled queries},
  author={Li, Yanwei and Yu, Zhiding and Philion, Jonah and Anandkumar, Anima and Fidler, Sanja and Jia, Jiaya and Alvarez, Jose},
  booktitle={Proceedings of the IEEE/CVF International Conference on Computer Vision},
  pages={18302--18311},
  year={2023}
}

@article{fischer2022cc,
  title={Cc-3dt: Panoramic 3d object tracking via cross-camera fusion},
  author={Fischer, Tobias and Yang, Yung-Hsu and Kumar, Suryansh and Sun, Min and Yu, Fisher},
  journal={arXiv preprint arXiv:2212.01247},
  year={2022}
}

@inproceedings{pang2023standing,
  title={Standing between past and future: Spatio-temporal modeling for multi-camera 3d multi-object tracking},
  author={Pang, Ziqi and Li, Jie and Tokmakov, Pavel and Chen, Dian and Zagoruyko, Sergey and Wang, Yu-Xiong},
  booktitle={Proceedings of the IEEE/CVF conference on computer vision and pattern recognition},
  pages={17928--17938},
  year={2023}
}

@inproceedings{sun2025sparsedrive,
  title={Sparsedrive: End-to-end autonomous driving via sparse scene representation},
  author={Sun, Wenchao and Lin, Xuewu and Shi, Yining and Zhang, Chuang and Wu, Haoran and Zheng, Sifa},
  booktitle={2025 IEEE International Conference on Robotics and Automation (ICRA)},
  pages={8795--8801},
  year={2025},
  organization={IEEE}
}

@inproceedings{zhang2025bridging,
  title={Bridging past and future: End-to-end autonomous driving with historical prediction and planning},
  author={Zhang, Bozhou and Song, Nan and Jin, Xin and Zhang, Li},
  booktitle={Proceedings of the Computer Vision and Pattern Recognition Conference},
  pages={6854--6863},
  year={2025}
}

@article{wu2026uc,
  title={UC-Track: Uncertainty-Aware and Task-Coupled 3-D Multi-Object Tracking},
  author={Wu, Di and Peng, Jiankun and Yu, Shuangzhi and Xu, Ke and Chen, Zhijun and Ma, Chunye},
  journal={IEEE Transactions on Intelligent Transportation Systems},
  year={2026},
  publisher={IEEE}
}

@article{doll2023star,
  title={Star-track: Latent motion models for end-to-end 3d object tracking with adaptive spatio-temporal appearance representations},
  author={Doll, Simon and Hanselmann, Niklas and Schneider, Lukas and Schulz, Richard and Enzweiler, Markus and Lensch, Hendrik PA},
  journal={IEEE Robotics and Automation Letters},
  volume={9},
  number={2},
  pages={1326--1333},
  year={2023},
  publisher={IEEE}
}

@inproceedings{gu2023vip3d,
  title={Vip3d: End-to-end visual trajectory prediction via 3d agent queries},
  author={Gu, Junru and Hu, Chenxu and Zhang, Tianyuan and Chen, Xuanyao and Wang, Yilun and Wang, Yue and Zhao, Hang},
  booktitle={Proceedings of the IEEE/CVF Conference on Computer Vision and Pattern Recognition},
  pages={5496--5506},
  year={2023}
}

@inproceedings{jia2023driveadapter,
  title={Driveadapter: Breaking the coupling barrier of perception and planning in end-to-end autonomous driving},
  author={Jia, Xiaosong and Gao, Yulu and Chen, Li and Yan, Junchi and Liu, Patrick Langechuan and Li, Hongyang},
  booktitle={Proceedings of the IEEE/CVF International Conference on Computer Vision},
  pages={7953--7963},
  year={2023}
}

@article{sun2024sparsedrive,
  title={SparseDrive: End-to-End Autonomous Driving via Sparse Scene Representation},
  author={Sun, Wenchao and Lin, Xuewu and Shi, Yining and Zhang, Chuang and Wu, Haoran and Zheng, Sifa},
  journal={arXiv preprint arXiv:2405.19620},
  year={2024}
}

@article{chen2024vadv2,
  title={Vadv2: End-to-end vectorized autonomous driving via probabilistic planning},
  author={Chen, Shaoyu and Jiang, Bo and Gao, Hao and Liao, Bencheng and Xu, Qing and Zhang, Qian and Huang, Chang and Liu, Wenyu and Wang, Xinggang},
  journal={arXiv preprint arXiv:2402.13243},
  year={2024}
}

@inproceedings{jiang2023vad,
  title={Vad: Vectorized scene representation for efficient autonomous driving},
  author={Jiang, Bo and Chen, Shaoyu and Xu, Qing and Liao, Bencheng and Chen, Jiajie and Zhou, Helong and Zhang, Qian and Liu, Wenyu and Huang, Chang and Wang, Xinggang},
  booktitle={Proceedings of the IEEE/CVF International Conference on Computer Vision},
  pages={8340--8350},
  year={2023}
}

@misc{contributors2023openscene,
  title={Openscene: The largest up-to-date 3d occupancy prediction benchmark in autonomous driving},
  author={Contributors, OpenScene},
  year={2023}
}

@article{caesar2021nuplan,
  title={nuplan: A closed-loop ml-based planning benchmark for autonomous vehicles},
  author={Caesar, Holger and Kabzan, Juraj and Tan, Kok Seang and Fong, Whye Kit and Wolff, Eric and Lang, Alex and Fletcher, Luke and Beijbom, Oscar and Omari, Sammy},
  journal={arXiv preprint arXiv:2106.11810},
  year={2021}
}

@InProceedings{CARLA,
  title = 	 {{CARLA}: {An} Open Urban Driving Simulator},
  author = 	 {Dosovitskiy, Alexey and Ros, German and Codevilla, Felipe and Lopez, Antonio and Koltun, Vladlen},
  booktitle = 	 {Proceedings of the 1st Annual Conference on Robot Learning},
  pages = 	 {1--16},
  year = 	 {2017},
  editor = 	 {Levine, Sergey and Vanhoucke, Vincent and Goldberg, Ken},
  volume = 	 {78},
  series = 	 {Proceedings of Machine Learning Research},
  month = 	 {13--15 Nov},
  publisher =    {PMLR},
  url = 	 {https://proceedings.mlr.press/v78/dosovitskiy17a.html}
}

@inproceedings{momad,
  title={Don't Shake the Wheel: Momentum-Aware Planning in End-to-End Autonomous Driving},
  author={Song, Ziying and Jia, Caiyan and Liu, Lin and Pan, Hongyu and Zhang, Yongchang and Wang, Junming and Zhang, Xingyu and Xu, Shaoqing and Yang, Lei and Luo, Yadan},
  booktitle={Proceedings of the Computer Vision and Pattern Recognition Conference},
  pages={22432--22441},
  year={2025}
}

@article{wote,
  title={End-to-end driving with online trajectory evaluation via bev world model},
  author={Li, Yingyan and Wang, Yuqi and Liu, Yang and He, Jiawei and Fan, Lue and Zhang, Zhaoxiang},
  journal={arXiv preprint arXiv:2504.01941},
  year={2025}
}

@article{yang2024deepinteraction++,
  title={DeepInteraction++: Multi-Modality Interaction for Autonomous Driving},
  author={Yang, Zeyu and Song, Nan and Li, Wei and Zhu, Xiatian and Zhang, Li and Torr, Philip HS},
  journal={arXiv preprint arXiv:2408.05075},
  year={2024}
}

@article{jia2024bench2drive,
  title={Bench2drive: Towards multi-ability benchmarking of closed-loop end-to-end autonomous driving},
  author={Jia, Xiaosong and Yang, Zhenjie and Li, Qifeng and Zhang, Zhiyuan and Yan, Junchi},
  journal={Advances in Neural Information Processing Systems},
  volume={37},
  pages={819--844},
  year={2024}
}

@article{dauner2024navsim,
  title={Navsim: Data-driven non-reactive autonomous vehicle simulation and benchmarking},
  author={Dauner, Daniel and Hallgarten, Marcel and Li, Tianyu and Weng, Xinshuo and Huang, Zhiyu and Yang, Zetong and Li, Hongyang and Gilitschenski, Igor and Ivanovic, Boris and Pavone, Marco and others},
  journal={Advances in Neural Information Processing Systems},
  volume={37},
  pages={28706--28719},
  year={2024}
}

@article{e2etpamisurvey,
  title={End-to-end autonomous driving: Challenges and frontiers},
  author={Chen, Li and Wu, Penghao and Chitta, Kashyap and Jaeger, Bernhard and Geiger, Andreas and Li, Hongyang},
  journal={IEEE Transactions on Pattern Analysis and Machine Intelligence},
  year={2024},
  publisher={IEEE}
}

@misc{fusionad,
      title={FusionAD: Multi-modality Fusion for Prediction and Planning Tasks of Autonomous Driving}, 
      author={Tengju Ye and Wei Jing and Chunyong Hu and Shikun Huang and Lingping Gao and Fangzhen Li and Jingke Wang and Ke Guo and Wencong Xiao and Weibo Mao and Hang Zheng and Kun Li and Junbo Chen and Kaicheng Yu},
      year={2023},
      eprint={2308.01006},
      archivePrefix={arXiv},
      primaryClass={cs.CV},
      url={https://arxiv.org/abs/2308.01006}, 
}

@inproceedings{li2024enhancing_law,
  title={Enhancing end-to-end autonomous driving with latent world model},
  author={Li, Yingyan and Fan, Lue and He, Jiawei and Wang, Yuqi and Chen, Yuntao and Zhang, Zhaoxiang and Tan, Tieniu},
  booktitle={International Conference on Learning Representations},
  volume={2025},
  pages={42942--42959},
  year={2025}
}

@inproceedings{zheng2024genad,
  title={Genad: Generative end-to-end autonomous driving},
  author={Zheng, Wenzhao and Song, Ruiqi and Guo, Xianda and Zhang, Chenming and Chen, Long},
  booktitle={European Conference on Computer Vision},
  pages={87--104},
  year={2024},
  organization={Springer}
}

@article{diver,
  title={Breaking imitation bottlenecks: Reinforced diffusion powers diverse trajectory generation},
  author={Song, Ziying and Liu, Lin and Pan, Hongyu and Liao, Bencheng and Guo, Mingzhe and Yang, Lei and Zhang, Yongchang and Xu, Shaoqing and Jia, Caiyan and Luo, Yadan},
  journal={arXiv preprint arXiv:2507.04049},
  year={2025}
}

@article{sun2026focalad,
  title={FocalAD: Local Motion Planning for End-to-End Autonomous Driving: B. Sun et al.},
  author={Sun, Bin and Zhang, Boao and Lu, Jiayi and Feng, Xinjie and Shang, Jiachen and Cao, Rui and Zheng, Mengchao and Wang, Chuanye and Yang, Shichun and Cao, Yaoguang and others},
  journal={Automotive Innovation},
  pages={1--11},
  year={2026},
  publisher={Springer}
}

@inproceedings{liu2026guideflow,
  title={Guideflow: Constraint-guided flow matching for planning in end-to-end autonomous driving},
  author={Liu, Lin and Jia, Caiyan and Yu, Guanyi and Song, Ziying and Li, JunQiao and Jia, Feiyang and Wu, Peiliang and Hao, Xiaoshuai and Luo, Yadan},
  booktitle={Proceedings of the IEEE/CVF Conference on Computer Vision and Pattern Recognition},
  pages={3719--3728},
  year={2026}
}

@inproceedings{ji2025ocrfdet,
  title={Ocrfdet: Object-centric radiance fields for multi-view 3d object detection in autonomous driving},
  author={Ji, Mingqian and Zhang, Shanshan and Yang, Jian},
  booktitle={Proceedings of the IEEE/CVF International Conference on Computer Vision},
  pages={24933--24942},
  year={2025}
}

@article{kim2025bridgeta,
  title={BridgeTA: Bridging the Representation Gap in Knowledge Distillation via Teacher Assistant for Bird's Eye View Map Segmentation},
  author={Kim, Beomjun and Woo, Suhan and Heo, Sejong and Kim, Euntai},
  journal={arXiv preprint arXiv:2508.09599},
  year={2025}
}

@article{zuo2026quadricformer,
  title={Quadricformer: Scene as superquadrics for 3d semantic occupancy prediction},
  author={Zuo, Sicheng and Zheng, Wenzhao and Han, Xiaoyong and Yang, Longchao and Lu, Jiwen and others},
  journal={Advances in Neural Information Processing Systems},
  volume={38},
  pages={47779--47801},
  year={2026}
}

@inproceedings{ssr,
  title={Navigation-guided sparse scene representation for end-to-end autonomous driving},
  author={Li, Peidong and Cui, Dixiao},
  booktitle={International Conference on Learning Representations},
  volume={2025},
  pages={29118--29134},
  year={2025}
}

@inproceedings{li2024ego,
  title={Is ego status all you need for open-loop end-to-end autonomous driving?},
  author={Li, Zhiqi and Yu, Zhiding and Lan, Shiyi and Li, Jiahan and Kautz, Jan and Lu, Tong and Alvarez, Jose M},
  booktitle={Proceedings of the IEEE/CVF Conference on Computer Vision and Pattern Recognition},
  pages={14864--14873},
  year={2024}
}

@InProceedings{uniad,
    author    = {Hu, Yihan and Yang, Jiazhi and Chen, Li and Li, Keyu and Sima, Chonghao and Zhu, Xizhou and Chai, Siqi and Du, Senyao and Lin, Tianwei and Wang, Wenhai and Lu, Lewei and Jia, Xiaosong and Liu, Qiang and Dai, Jifeng and Qiao, Yu and Li, Hongyang},
    title     = {Planning-Oriented Autonomous Driving},
    booktitle = {Proceedings of the IEEE/CVF Conference on Computer Vision and Pattern Recognition (CVPR)},
    month     = {June},
    year      = {2023},
    pages     = {17853-17862}
}

@article{li2024hydra,
  title={Hydra-MDP: End-to-end Multimodal Planning with Multi-target Hydra-Distillation},
  author={Li, Zhenxin and Li, Kailin and Wang, Shihao and Lan, Shiyi and Yu, Zhiding and Ji, Yishen and Li, Zhiqi and Zhu, Ziyue and Kautz, Jan and Wu, Zuxuan and others},
  journal={arXiv preprint arXiv:2406.06978},
  year={2024}
}

@article{liao2024diffusiondrive,
  title={DiffusionDrive: Truncated Diffusion Model for End-to-End Autonomous Driving},
  author={Liao, Bencheng and Chen, Shaoyu and Yin, Haoran and Jiang, Bo and Wang, Cheng and Yan, Sixu and Zhang, Xinbang and Li, Xiangyu and Zhang, Ying and Zhang, Qian and others},
  journal={arXiv preprint arXiv:2411.15139},
  year={2024}
}

@ARTICLE{TransFuser,
  author={Chitta, Kashyap and Prakash, Aditya and Jaeger, Bernhard and Yu, Zehao and Renz, Katrin and Geiger, Andreas},
  journal={IEEE Transactions on Pattern Analysis and Machine Intelligence}, 
  title={TransFuser: Imitation With Transformer-Based Sensor Fusion for Autonomous Driving}, 
  year={2023},
  volume={45},
  number={11},
  pages={12878-12895},
  doi={10.1109/TPAMI.2022.3200245}}

@inproceedings{paradrive,
  title={Para-drive: Parallelized architecture for real-time autonomous driving},
  author={Weng, Xinshuo and Ivanovic, Boris and Wang, Yan and Wang, Yue and Pavone, Marco},
  booktitle={Proceedings of the IEEE/CVF Conference on Computer Vision and Pattern Recognition},
  pages={15449--15458},
  year={2024}
}

@article{song2024graphbev,
  title={Graphbev: Towards robust bev feature alignment for multi-modal 3d object detection},
  author={Song, Ziying and Yang, Lei and Xu, Shaoqing and Liu, Lin and Xu, Dongyang and Jia, Caiyan and Jia, Feiyang and Wang, Li},
  journal={arXiv preprint arXiv:2403.11848},
  year={2024}
}

@inproceedings{chen2023voxelnext,
 title={{VoxelNeXt: Fully Sparse VoxelNet for 3D Object Detection and Tracking}},
  author={Yukang Chen and Jianhui Liu and Xiangyu Zhang and Xiaojuan Qi and Jiaya Jia},
  booktitle={Proceedings of the IEEE/CVF Conference on Computer Vision and Pattern Recognition},
  year={2023}
}

@article{wang2023multi,
  title={Multi-modal 3D Object Detection in Autonomous Driving: A Survey and Taxonomy},
  author={Wang, Li and Zhang, Xinyu and Song, Ziying and Bi, Jiangfeng and Zhang, Guoxin and Wei, Haiyue and Tang, Liyao and Yang, Lei and Li, Jun and Jia, Caiyan and others},
  journal={IEEE Transactions on Intelligent Vehicles},
  year={2023},
  publisher={IEEE}
}

@ARTICLE{song2024robustness,
  author={Song, Ziying and Liu, Lin and Jia, Feiyang and Luo, Yadan and Jia, Caiyan and Zhang, Guoxin and Yang, Lei and Wang, Li},
  journal={IEEE Transactions on Intelligent Transportation Systems}, 
  title={Robustness-Aware 3D Object Detection in Autonomous Driving: A Review and Outlook}, 
  year={2024},
  volume={},
  number={},
  pages={1-30},
  doi={10.1109/TITS.2024.3439557},
  ISSN={1558-0016},
  month={},}

@InProceedings{CMT,
    author    = {Yan, Junjie and Liu, Yingfei and Sun, Jianjian and Jia, Fan and Li, Shuailin and Wang, Tiancai and Zhang, Xiangyu},
    title     = {Cross Modal Transformer: Towards Fast and Robust 3D Object Detection},
    booktitle = {Proceedings of the IEEE/CVF International Conference on Computer Vision (ICCV)},
    month     = {October},
    year      = {2023},
    pages     = {18268-18278}
}

@InProceedings{SparseFusion,
    author    = {Xie, Yichen and Xu, Chenfeng and Rakotosaona, Marie-Julie and Rim, Patrick and Tombari, Federico and Keutzer, Kurt and Tomizuka, Masayoshi and Zhan, Wei},
    title     = {SparseFusion: Fusing Multi-Modal Sparse Representations for Multi-Sensor 3D Object Detection},
    booktitle = {Proceedings of the IEEE/CVF International Conference on Computer Vision (ICCV)},
    month     = {October},
    year      = {2023},
    pages     = {17591-17602}
}

@InProceedings{MSMDFusion,
    author    = {Jiao, Yang and Jie, Zequn and Chen, Shaoxiang and Chen, Jingjing and Ma, Lin and Jiang, Yu-Gang},
    title     = {MSMDFusion: Fusing LiDAR and Camera at Multiple Scales With Multi-Depth Seeds for 3D Object Detection},
    booktitle = {Proceedings of the IEEE/CVF Conference on Computer Vision and Pattern Recognition (CVPR)},
    month     = {June},
    year      = {2023},
    pages     = {21643-21652}
}

@article{bevfusion-pku,
 title={Bevfusion: A simple and robust lidar-camera fusion framework},
  author={Liang, Tingting and Xie, Hongwei and Yu, Kaicheng and Xia, Zhongyu and Lin, Zhiwei and Wang, Yongtao and Tang, Tao and Wang, Bing and Tang, Zhi},
  journal={Advances in Neural Information Processing Systems},
  volume={35},
  pages={10421--10434},
  year={2022}
}

@article{wang2025mv2dfusion,
  title={Mv2dfusion: Leveraging modality-specific object semantics for multi-modal 3d detection},
  author={Wang, Zitian and Huang, Zehao and Gao, Yulu and Wang, Naiyan and Liu, Si},
  journal={IEEE Transactions on Pattern Analysis and Machine Intelligence},
  year={2025},
  publisher={IEEE}
}

@inproceedings{wang2023unitr,
  title={Unitr: A unified and efficient multi-modal transformer for bird's-eye-view representation},
  author={Wang, Haiyang and Tang, Hao and Shi, Shaoshuai and Li, Aoxue and Li, Zhenguo and Schiele, Bernt and Wang, Liwei},
  booktitle={Proceedings of the IEEE/CVF international conference on computer vision},
  pages={6792--6802},
  year={2023}
}

@article{chen2026sparsebev,
  title={SparseBEV: A Fully Sparse Framework for Multi-View 3D Object Detection},
  author={Chen, Yang and Liu, Haisong and Wang, Limin},
  journal={IEEE Transactions on Pattern Analysis and Machine Intelligence},
  year={2026},
  publisher={IEEE}
}

@inproceedings{li2026refine3d,
  title={Refine3D: Scene-Adaptive Reference Point Refinement for Sparse 3D Object Detection},
  author={Li, Fan and Lu, Jing and Xu, Yunlu and Wu, Changhong and Xu, Tao and Xiang, Zhaoyi and Niu, Yi},
  booktitle={Proceedings of the AAAI Conference on Artificial Intelligence},
  volume={40},
  number={8},
  pages={6073--6081},
  year={2026}
}

@inproceedings{zhang2025dgfsd,
  title={DGFSD: Bridging the Gap between Dense and Sparse for Fully Sparse 3D Object Detection},
  author={Zhang, Guoxin and Ou, Zhonghong and Xue, Kaiwen and Sun, Jiangfeng and Zhu, Yifan and Yao, Siyuan and Shen, Yiran and Song, Meina},
  booktitle={Proceedings of the 33rd ACM International Conference on Multimedia},
  pages={4669--4678},
  year={2025}
}

@article{fan2024fsd,
  title={Fsd v2: Improving fully sparse 3d object detection with virtual voxels},
  author={Fan, Lue and Wang, Feng and Wang, Naiyan and Zhang, Zhaoxiang},
  journal={IEEE Transactions on Pattern Analysis and Machine Intelligence},
  volume={47},
  number={2},
  pages={1279--1292},
  year={2024},
  publisher={IEEE}
}

@article{ji2026depthfusion,
  title={Depthfusion: Depth-aware hybrid feature fusion for LiDAR-camera 3D object detection},
  author={Ji, Mingqian and Zhang, Shanshan and Yang, Jian},
  journal={IEEE Transactions on Multimedia},
  year={2026},
  publisher={IEEE}
}

@article{jing2026hd,
  title={HD-Fusion: Hierarchical Dynamic Fusion of LiDAR-Camera for Robust 3-D Object Detection},
  author={Jing, Weiming and Chen, Xiyuan and Nie, Shuhan and Jiao, Zhiyuan and Ma, Jianghui},
  journal={IEEE Transactions on Industrial Informatics},
  year={2026},
  publisher={IEEE}
}

@article{bevfusion-mit,
  title={Bevfusion: Multi-task multi-sensor fusion with unified bird's-eye view representation},
  author={Liu, Zhijian and Tang, Haotian and Amini, Alexander and Yang, Xinyu and Mao, Huizi and Rus, Daniela L and Han, Song},
  booktitle={2023 IEEE international conference on robotics and automation (ICRA)},
  pages={2774--2781},
  year={2023},
  organization={IEEE}
}

@article{lei2023hvdetfusion,
  title={Hvdetfusion: A simple and robust camera-radar fusion framework},
  author={Lei, Kai and Chen, Zhan and Jia, Shuman and Zhang, Xiaoteng},
  journal={arXiv preprint arXiv:2307.11323},
  year={2023}
}

@inproceedings{xu2024sparseinteraction,
  title={SparseInteraction: Sparse semantic guidance for radar and camera 3D object detection},
  author={Xu, Shaoqing and Jiang, Shengyin and Li, Fang and Liu, Li and Song, Ziying and Yang, Bo and Yang, Zhi-xin},
  booktitle={Proceedings of the 32nd ACM International Conference on Multimedia},
  pages={9224--9233},
  year={2024}
}

@inproceedings{wang2023exploring,
  title={Exploring object-centric temporal modeling for efficient multi-view 3d object detection},
  author={Wang, Shihao and Liu, Yingfei and Wang, Tiancai and Li, Ying and Zhang, Xiangyu},
  booktitle={Proceedings of the IEEE/CVF international conference on computer vision},
  pages={3621--3631},
  year={2023}
}

@inproceedings{chu2024rayformer,
  title={Rayformer: Improving query-based multi-camera 3d object detection via ray-centric strategies},
  author={Chu, Xiaomeng and Deng, Jiajun and You, Guoliang and Duan, Yifan and Li, Yao and Zhang, Yanyong},
  booktitle={Proceedings of the 32nd ACM International Conference on Multimedia},
  pages={4620--4629},
  year={2024}
}

@inproceedings{lin2024rcbevdet,
  title={RCBEVDet: Radar-camera fusion in bird's eye view for 3D object detection},
  author={Lin, Zhiwei and Liu, Zhe and Xia, Zhongyu and Wang, Xinhao and Wang, Yongtao and Qi, Shengxiang and Dong, Yang and Dong, Nan and Zhang, Le and Zhu, Ce},
  booktitle={Proceedings of the IEEE/CVF Conference on Computer Vision and Pattern Recognition},
  pages={14928--14937},
  year={2024}
}

@inproceedings{jiang2024far3d,
  title={Far3d: Expanding the horizon for surround-view 3d object detection},
  author={Jiang, Xiaohui and Li, Shuailin and Liu, Yingfei and Wang, Shihao and Jia, Fan and Wang, Tiancai and Han, Lijin and Zhang, Xiangyu},
  booktitle={Proceedings of the AAAI conference on artificial intelligence},
  volume={38},
  number={3},
  pages={2561--2569},
  year={2024}
}

@article{zhang2023hednet,
  title={Hednet: A hierarchical encoder-decoder network for 3d object detection in point clouds},
  author={Zhang, Gang and Junnan, Chen and Gao, Guohuan and Li, Jianmin and Hu, Xiaolin},
  journal={Advances in Neural Information Processing Systems},
  volume={36},
  pages={53076--53089},
  year={2023}
}

@inproceedings{zhang2024safdnet,
  title={Safdnet: A simple and effective network for fully sparse 3d object detection},
  author={Zhang, Gang and Chen, Junnan and Gao, Guohuan and Li, Jianmin and Liu, Si and Hu, Xiaolin},
  booktitle={Proceedings of the IEEE/CVF conference on computer vision and pattern recognition},
  pages={14477--14486},
  year={2024}
}

@article{liu2024lion,
  title={Lion: Linear group rnn for 3d object detection in point clouds},
  author={Liu, Zhe and Hou, Jinghua and Wang, Xinyu and Ye, Xiaoqing and Wang, Jingdong and Zhao, Hengshuang and Bai, Xiang},
  journal={Advances in Neural Information Processing Systems},
  volume={37},
  pages={13601--13626},
  year={2024}
}

@inproceedings{jin2025unimamba,
  title={UniMamba: Unified spatial-channel representation learning with group-efficient mamba for LiDAR-based 3D object detection},
  author={Jin, Xin and Su, Haisheng and Liu, Kai and Ma, Cong and Wu, Wei and Hui, Fei and Yan, Junchi},
  booktitle={Proceedings of the Computer Vision and Pattern Recognition Conference},
  pages={1407--1417},
  year={2025}
}

@inproceedings{jin2025geoformer,
  title={GeoFormer: Geometry Point Encoder for 3D Object Detection with Graph-based Transformer},
  author={Jin, Xin and Su, Haisheng and Ma, Cong and Liu, Kai and Wu, Wei and Hui, Fei and Yan, Junchi},
  booktitle={Proceedings of the IEEE/CVF International Conference on Computer Vision},
  pages={26879--26889},
  year={2025}
}

@inproceedings{liu2025fshnet,
  title={FSHNet: Fully sparse hybrid network for 3D object detection},
  author={Liu, Shuai and Cui, Mingyue and Li, Boyang and Liang, Quanmin and Hong, Tinghe and Huang, Kai and Shan, Yunxiao},
  booktitle={Proceedings of the Computer Vision and Pattern Recognition Conference},
  pages={8900--8909},
  year={2025}
}

@inproceedings{li2026look,
  title={Look Before You Fuse: 2D-Guided Cross-Modal Alignment for Robust 3D Detection},
  author={Li, Xiang and Hu, Zhangchi and Xiao, Xu and Kong, Bin},
  booktitle={Proceedings of the IEEE/CVF Conference on Computer Vision and Pattern Recognition},
  pages={11589--11598},
  year={2026}
}

@article{yuan2023m,
  title={M3Net: Multilevel, Mixed and Multistage Attention Network for Salient Object Detection},
  author={Yuan, Yao and Gao, Pan and Tan, XiaoYang},
  journal={arXiv preprint arXiv:2309.08365},
  year={2023}
}

@inproceedings{cvt,
  title={Cross-view transformers for real-time map-view semantic segmentation},
  author={Zhou, Brady and Kr{\"a}henb{\"u}hl, Philipp},
  booktitle={Proceedings of the IEEE/CVF conference on computer vision and pattern recognition},
  pages={13760--13769},
  year={2022}
}

@inproceedings{kim2023crn,
  title={Crn: Camera radar net for accurate, robust, efficient 3d perception},
  author={Kim, Youngseok and Shin, Juyeb and Kim, Sanmin and Lee, In-Jae and Choi, Jun Won and Kum, Dongsuk},
  booktitle={Proceedings of the IEEE/CVF International Conference on Computer Vision},
  pages={17615--17626},
  year={2023}
}

@inproceedings{wolters2025unleashing,
  title={Unleashing hydra: Hybrid fusion, depth consistency and radar for unified 3d perception},
  author={Wolters, Philipp and Gilg, Johannes and Teepe, Torben and Herzog, Fabian and Laouichi, Anouar and Hofmann, Martin and Rigoll, Gerhard},
  booktitle={2025 IEEE International Conference on Robotics and Automation (ICRA)},
  pages={7467--7474},
  year={2025},
  organization={IEEE}
}

@article{song2024contrastalign,
  title={ContrastAlign: Toward Robust BEV Feature Alignment via Contrastive Learning for Multi-Modal 3D Object Detection},
  author={Song, Ziying and Jia, Feiyang and Pan, Hongyu and Luo, Yadan and Jia, Caiyan and Zhang, Guoxin and Liu, Lin and Ji, Yang and Yang, Lei and Wang, Li},
  journal={arXiv preprint arXiv:2405.16873},
  year={2024}
}

@InProceedings{ObjectFusion,
    author    = {Cai, Qi and Pan, Yingwei and Yao, Ting and Ngo, Chong-Wah and Mei, Tao},
    title     = {ObjectFusion: Multi-modal 3D Object Detection with Object-Centric Fusion},
    booktitle = {Proceedings of the IEEE/CVF International Conference on Computer Vision (ICCV)},
    month     = {October},
    year      = {2023},
    pages     = {18067-18076}
}

@article{song2024robofusion,
  title={RoboFusion: Towards robust multi-modal 3D object detection via SAM},
  author={Song, Ziying and Zhang, Guoxing and Liu, Lin and Yang, Lei and Xu, Shaoqing and Jia, Caiyan and Jia, Feiyang and Wang, Li},
  journal={arXiv preprint arXiv:2401.03907},
  year={2024}
}

@InProceedings{ge2023metabev,
    author    = {Ge, Chongjian and Chen, Junsong and Xie, Enze and Wang, Zhongdao and Hong, Lanqing and Lu, Huchuan and Li, Zhenguo and Luo, Ping},
    title     = {MetaBEV: Solving Sensor Failures for 3D Detection and Map Segmentation},
    booktitle = {Proceedings of the IEEE/CVF International Conference on Computer Vision (ICCV)},
    month     = {October},
    year      = {2023},
    pages     = {8721-8731}
}

@inproceedings{ji2023ddp,
  title={Ddp: Diffusion model for dense visual prediction},
  author={Ji, Yuanfeng and Chen, Zhe and Xie, Enze and Hong, Lanqing and Liu, Xihui and Liu, Zhaoqiang and Lu, Tong and Li, Zhenguo and Luo, Ping},
  booktitle={Proceedings of the IEEE/CVF International Conference on Computer Vision},
  pages={21741--21752},
  year={2023}
}

@inproceedings{chen2024residual,
  title={Residual graph convolutional network for bird's-eye-view semantic segmentation},
  author={Chen, Qiuxiao and Qi, Xiaojun},
  booktitle={Proceedings of the IEEE/CVF Winter Conference on Applications of Computer Vision},
  pages={3324--3331},
  year={2024}
}

@inproceedings{mvx-net,
  title={Mvx-net: Multimodal voxelnet for 3d object detection},
  author={Sindagi, Vishwanath A and Zhou, Yin and Tuzel, Oncel},
  booktitle={2019 International Conference on Robotics and Automation (ICRA)},
  pages={7276--7282},
  year={2019},
  organization={IEEE}
}

@inproceedings{pointpainting,
  title={Pointpainting: Sequential fusion for 3d object detection},
  author={Vora, Sourabh and Lang, Alex H and Helou, Bassam and Beijbom, Oscar},
  booktitle={Proceedings of the IEEE/CVF conference on computer vision and pattern recognition},
  pages={4604--4612},
  year={2020}
}

@inproceedings{epnet,
  title={Epnet: Enhancing point features with image semantics for 3d object detection},
  author={Huang, Tengteng and Liu, Zhe and Chen, Xiwu and Bai, Xiang},
  booktitle={European Conference on Computer Vision},
  pages={35--52},
  year={2020},
  organization={Springer}
}

@article{epnet++,
  author={Liu, Zhe and Huang, Tengteng and Li, Bingling and Chen, Xiwu and Wang, Xi and Bai, Xiang},
  journal={IEEE Transactions on Pattern Analysis and Machine Intelligence}, 
  title={EPNet++: Cascade Bi-Directional Fusion for Multi-Modal 3D Object Detection}, 
  year={2023},
  volume={45},
  number={7},
  pages={8324-8341},
  doi={10.1109/TPAMI.2022.3228806}}

@inproceedings{wang2021pointaugmenting,
  title={Pointaugmenting: Cross-modal augmentation for 3d object detection},
  author={Wang, Chunwei and Ma, Chao and Zhu, Ming and Yang, Xiaokang},
  booktitle={Proceedings of the IEEE/CVF Conference on Computer Vision and Pattern Recognition},
  pages={11794--11803},
  year={2021}
}

@article{bi2024dyfusion,
  title={DyFusion: Cross-Attention 3D Object Detection with Dynamic Fusion},
  author={Bi, Jiangfeng and Wei, Haiyue and Zhang, Guoxin and Yang, Kuihe and Song, Ziying},
  journal={IEEE Latin America Transactions},
  volume={22},
  number={2},
  pages={106--112},
  year={2024},
  publisher={IEEE}
}

@article{mvp,
  title={Multimodal virtual point 3d detection},
  author={Yin, Tianwei and Zhou, Xingyi and Kr{\"a}henb{\"u}hl, Philipp},
  journal={Advances in Neural Information Processing Systems},
  volume={34},
  pages={16494--16507},
  year={2021}
}

@inproceedings{zhang2023urformer,
  title={Urformer: Unified representation lidar-camera 3d object detection with transformer},
  author={Zhang, Guoxin and Xie, Jun and Liu, Lin and Wang, Zhepeng and Yang, Kuihe and Song, Ziying},
  booktitle={Chinese Conference on Pattern Recognition and Computer Vision (PRCV)},
  pages={401--413},
  year={2023},
  organization={Springer}
}

@ARTICLE{VoxelNextFusion,
  author={Song, Ziying and Zhang, Guoxin and Xie, Jun and Liu, Lin and Jia, Caiyan and Xu, Shaoqing and Wang, Zhepeng},
  journal={IEEE Transactions on Geoscience and Remote Sensing}, 
  title={VoxelNextFusion: A Simple, Unified, and Effective Voxel Fusion Framework for Multimodal 3-D Object Detection}, 
  year={2023},
  volume={61},
  number={},
  pages={1-12},
  doi={10.1109/TGRS.2023.3331893}}

@InProceedings{Transfusion,
    author    = {Bai, Xuyang and Hu, Zeyu and Zhu, Xinge and Huang, Qingqiu and Chen, Yilun and Fu, Hongbo and Tai, Chiew-Lan},
    title     = {TransFusion: Robust LiDAR-Camera Fusion for 3D Object Detection With Transformers},
    booktitle = {Proceedings of the IEEE/CVF Conference on Computer Vision and Pattern Recognition (CVPR)},
    month     = {June},
    year      = {2022},
    pages     = {1090-1099}
}

@InProceedings{DeepFusion,
    author    = {Li, Yingwei and Yu, Adams Wei and Meng, Tianjian and Caine, Ben and Ngiam, Jiquan and Peng, Daiyi and Shen, Junyang and Lu, Yifeng and Zhou, Denny and Le, Quoc V. and Yuille, Alan and Tan, Mingxing},
    title     = {DeepFusion: Lidar-Camera Deep Fusion for Multi-Modal 3D Object Detection},
    booktitle = {Proceedings of the IEEE/CVF Conference on Computer Vision and Pattern Recognition (CVPR)},
    month     = {June},
    year      = {2022},
    pages     = {17182-17191}
}

@inproceedings{DeepInteraction,
 author = {Yang, Zeyu and Chen, Jiaqi and Miao, Zhenwei and Li, Wei and Zhu, Xiatian and Zhang, Li},
 booktitle = {Advances in Neural Information Processing Systems},
 editor = {S. Koyejo and S. Mohamed and A. Agarwal and D. Belgrave and K. Cho and A. Oh},
 pages = {1992--2005},
 publisher = {Curran Associates, Inc.},
 title = {DeepInteraction: 3D Object Detection via Modality Interaction},
 url = {https://proceedings.neurips.cc/paper_files/paper/2022/file/0d18ab3b5fabfa6fe47c62e711af02f0-Paper-Conference.pdf},
 volume = {35},
 year = {2022}
}

@inproceedings{autoalign,  
 title={AutoAlign: Pixel-Instance Feature Aggregation for Multi-Modal 3D Object Detection}, 
 DOI={10.24963/ijcai.2022/116}, 
 booktitle={Proceedings of the Thirty-First International Joint Conference on Artificial Intelligence}, 
 author={Chen, Zehui and Li, Zhenyu and Zhang, Shiquan and Fang, Liangji and Jiang, Qinhong and Zhao, Feng and Zhou, Bolei and Zhao, Hang}, 
 year={2022}, 
 month={Jul}, 
 language={en-US} 
 }

@InProceedings{autoalignv2,
author="Chen, Zehui
and Li, Zhenyu
and Zhang, Shiquan
and Fang, Liangji
and Jiang, Qinhong
and Zhao, Feng",
editor="Avidan, Shai
and Brostow, Gabriel
and Ciss{\'e}, Moustapha
and Farinella, Giovanni Maria
and Hassner, Tal",
title="Deformable Feature Aggregation for Dynamic Multi-modal 3D Object Detection",
booktitle="Computer Vision -- ECCV 2022",
year="2022",
publisher="Springer Nature Switzerland",
address="Cham",
pages="628--644",
isbn="978-3-031-20074-8"
}

@inproceedings{graphalign,
  title={GraphAlign: Enhancing Accurate Feature Alignment by Graph matching for Multi-Modal 3D Object Detection},
  author={Song, Ziying and Wei, Haiyue and Bai, Lin and Yang, Lei and Jia, Caiyan},
  booktitle={Proceedings of the IEEE/CVF International Conference on Computer Vision},
  pages={3358--3369},
  year={2023}
}

@inproceedings{3dcvf,
  title={3d-cvf: Generating joint camera and lidar features using cross-view spatial feature fusion for 3d object detection},
  author={Yoo, Jin Hyeok and Kim, Yecheol and Kim, Jisong and Choi, Jun Won},
  booktitle={Computer Vision--ECCV 2020: 16th European Conference, Glasgow, UK, August 23--28, 2020, Proceedings, Part XXVII 16},
  pages={720--736},
  year={2020},
  organization={Springer}
}

@inproceedings{HMFI,
  title={Homogeneous multi-modal feature fusion and interaction for 3D object detection},
  author={Li, Xin and Shi, Botian and Hou, Yuenan and Wu, Xingjiao and Ma, Tianlong and Li, Yikang and He, Liang},
  booktitle={European Conference on Computer Vision},
  pages={691--707},
  year={2022},
  organization={Springer}
}

@article{Robust-FusionNet,
  title={Robust-FusionNet: Deep multimodal sensor fusion for 3-D object detection under severe weather conditions},
  author={Zhang, Cheng and Wang, Hai and Cai, Yingfeng and Chen, Long and Li, Yicheng and Sotelo, Miguel Angel and Li, Zhixiong},
  journal={IEEE Transactions on Instrumentation and Measurement},
  volume={71},
  pages={1--13},
  year={2022},
  publisher={IEEE}
}

@InProceedings{Logonet,
    author    = {Li, Xin and Ma, Tao and Hou, Yuenan and Shi, Botian and Yang, Yuchen and Liu, Youquan and Wu, Xingjiao and Chen, Qin and Li, Yikang and Qiao, Yu and He, Liang},
    title     = {LoGoNet: Towards Accurate 3D Object Detection With Local-to-Global Cross-Modal Fusion},
    booktitle = {Proceedings of the IEEE/CVF Conference on Computer Vision and Pattern Recognition (CVPR)},
    month     = {June},
    year      = {2023},
    pages     = {17524-17534}
}

@article{3DDualFusion,
  title={3D Dual-Fusion: Dual-Domain Dual-Query Camera-LiDAR Fusion for 3D Object Detection},
  author={Kim, Yecheol and Park, Konyul and Kim, Minwook and Kum, Dongsuk and Choi, Jun Won},
  journal={arXiv preprint arXiv:2211.13529},
  year={2022}
}

@InProceedings{CAT-Det,
    author    = {Zhang, Yanan and Chen, Jiaxin and Huang, Di},
    title     = {CAT-Det: Contrastively Augmented Transformer for Multi-Modal 3D Object Detection},
    booktitle = {Proceedings of the IEEE/CVF Conference on Computer Vision and Pattern Recognition (CVPR)},
    month     = {June},
    year      = {2022},
    pages     = {908-917}
}

@article{graphalign++,
  title={Graphalign++: An accurate feature alignment by graph matching for multi-modal 3d object detection},
  author={Song, Ziying and Jia, Caiyan and Yang, Lei and Wei, Haiyue and Liu, Lin},
  journal={IEEE Transactions on Circuits and Systems for Video Technology},
  year={2023},
  publisher={IEEE}
}

@inproceedings{bevdepth,
  title={{Bevdepth: Acquisition of reliable depth for multi-view 3d object detection}},
  author={Li, Yinhao and Ge, Zheng and Yu, Guanyi and Yang, Jinrong and Wang, Zengran and Shi, Yukang and Sun, Jianjian and Li, Zeming},
  booktitle={Proceedings of the AAAI Conference on Artificial Intelligence},
  volume={37},
  number={2},
  pages={1477--1485},
  year={2023}
}

@inproceedings{lss,
  title={Lift, splat, shoot: Encoding images from arbitrary camera rigs by implicitly unprojecting to 3d},
  author={Philion, Jonah and Fidler, Sanja},
  booktitle={Computer Vision--ECCV 2020: 16th European Conference, Glasgow, UK, August 23--28, 2020, Proceedings, Part XIV 16},
  pages={194--210},
  year={2020},
  organization={Springer}
}

@article{m2bev,
  title={M2BEV: Multi-camera joint 3D detection and segmentation with unified birds-eye view representation},
  author={Xie, Enze and Yu, Zhiding and Zhou, Daquan and Philion, Jonah and Anandkumar, Anima and Fidler, Sanja and Luo, Ping and Alvarez, Jose M},
  journal={arXiv preprint arXiv:2204.05088},
  year={2022}
}

@article{hao2025mapfusion,
  title={Mapfusion: A novel bev feature fusion network for multi-modal map construction},
  author={Hao, Xiaoshuai and Diao, Yunfeng and Wei, Mengchuan and Yang, Yifan and Hao, Peng and Yin, Rong and Zhang, Hui and Li, Weiming and Zhao, Shu and Liu, Yu},
  journal={Information Fusion},
  volume={119},
  pages={103018},
  year={2025},
  publisher={Elsevier}
}

@article{li2026nrseg,
  title={NRSeg: Noise-Resilient Learning for BEV Semantic Segmentation via Driving World Models},
  author={Li, Siyu and Teng, Fei and Cao, Yihong and Yang, Kailun and Li, Zhiyong and Wang, Yaonan},
  journal={IEEE Transactions on Image Processing},
  year={2026},
  publisher={IEEE}
}

@inproceedings{zhu2023mapprior,
  title={Mapprior: Bird's-eye view map layout estimation with generative models},
  author={Zhu, Xiyue and Zyrianov, Vlas and Liu, Zhijian and Wang, Shenlong},
  booktitle={Proceedings of the IEEE/CVF International Conference on Computer Vision},
  pages={8228--8239},
  year={2023}
}

@inproceedings{borse2023x,
  title={X-align: Cross-modal cross-view alignment for bird's-eye-view segmentation},
  author={Borse, Shubhankar and Klingner, Marvin and Kumar, Varun Ravi and Cai, Hong and Almuzairee, Abdulaziz and Yogamani, Senthil and Porikli, Fatih},
  booktitle={Proceedings of the IEEE/CVF Winter Conference on Applications of Computer Vision},
  pages={3287--3297},
  year={2023}
}

@InProceedings{yukaicheng_benchmarking,
    author    = {Yu, Kaicheng and Tao, Tang and Xie, Hongwei and Lin, Zhiwei and Liang, Tingting and Wang, Bing and Chen, Peng and Hao, Dayang and Wang, Yongtao and Liang, Xiaodan},
    title     = {Benchmarking the Robustness of LiDAR-Camera Fusion for 3D Object Detection},
    booktitle = {Proceedings of the IEEE/CVF Conference on Computer Vision and Pattern Recognition (CVPR) Workshops},
    month     = {June},
    year      = {2023},
    pages     = {3188-3198}
}

@InProceedings{zhujun_benchmarking,
    author    = {Dong, Yinpeng and Kang, Caixin and Zhang, Jinlai and Zhu, Zijian and Wang, Yikai and Yang, Xiao and Su, Hang and Wei, Xingxing and Zhu, Jun},
    title     = {Benchmarking Robustness of 3D Object Detection to Common Corruptions},
    booktitle = {Proceedings of the IEEE/CVF Conference on Computer Vision and Pattern Recognition (CVPR)},
    month     = {June},
    year      = {2023},
    pages     = {1022-1032}
}

@inproceedings{nuscenes,
  title={nuscenes: A multimodal dataset for autonomous driving},
  author={Caesar, Holger and Bankiti, Varun and Lang, Alex H and Vora, Sourabh and Liong, Venice Erin and Xu, Qiang and Krishnan, Anush and Pan, Yu and Baldan, Giancarlo and Beijbom, Oscar},
  booktitle={Proceedings of the IEEE/CVF conference on computer vision and pattern recognition},
  pages={11621--11631},
  year={2020}
}

@inproceedings{waymo,
  title={Scalability in perception for autonomous driving: Waymo open dataset},
  author={Sun, Pei and Kretzschmar, Henrik and Dotiwalla, Xerxes and Chouard, Aurelien and Patnaik, Vijaysai and Tsui, Paul and Guo, James and Zhou, Yin and Chai, Yuning and Caine, Benjamin and others},
  booktitle={Proceedings of the IEEE/CVF conference on computer vision and pattern recognition},
  pages={2446--2454},
  year={2020}
}

@INPROCEEDINGS { Argoverse2,
  author = {Benjamin Wilson and William Qi and Tanmay Agarwal and John Lambert and Jagjeet Singh and Siddhesh Khandelwal and Bowen Pan and Ratnesh Kumar and Andrew Hartnett and Jhony Kaesemodel Pontes and Deva Ramanan and Peter Carr and James Hays},
  title = {{Argoverse 2: Next Generation Datasets for Self-Driving Perception and Forecasting}},
  booktitle = {Proceedings of the Neural Information Processing Systems Track on Datasets and Benchmarks (NeurIPS Datasets and Benchmarks 2021)},
  year = {2021}
}

@inproceedings{Pointnet,
  title={Pointnet: Deep learning on point sets for 3d classification and segmentation},
  author={Qi, Charles R and Su, Hao and Mo, Kaichun and Guibas, Leonidas J},
  booktitle={Proceedings of the IEEE conference on computer vision and pattern recognition},
  pages={652--660},
  year={2017}
}

@article{Pointnet++,
  title={Pointnet++: Deep hierarchical feature learning on point sets in a metric space},
  author={Qi, Charles Ruizhongtai and Yi, Li and Su, Hao and Guibas, Leonidas J},
  journal={Advances in neural information processing systems},
  volume={30},
  year={2017}
}

@inproceedings{lidarrcnn,
  title={Lidar r-cnn: An efficient and universal 3d object detector},
  author={Li, Zhichao and Wang, Feng and Wang, Naiyan},
  booktitle={Proceedings of the IEEE/CVF Conference on Computer Vision and Pattern Recognition},
  pages={7546--7555},
  year={2021}
}

@inproceedings{Frustumpointnets,
  title={Frustum pointnets for 3d object detection from rgb-d data},
  author={Qi, Charles R and Liu, Wei and Wu, Chenxia and Su, Hao and Guibas, Leonidas J},
  booktitle={Proceedings of the IEEE conference on computer vision and pattern recognition},
  pages={918--927},
  year={2018}
}

@inproceedings{Pointrcnn,
  title={Pointrcnn: 3d object proposal generation and detection from point cloud},
  author={Shi, Shaoshuai and Wang, Xiaogang and Li, Hongsheng},
  booktitle={Proceedings of the IEEE/CVF conference on computer vision and pattern recognition},
  pages={770--779},
  year={2019}
}

@inproceedings{Voxelnet,
  title={Voxelnet: End-to-end learning for point cloud based 3d object detection},
  author={Zhou, Yin and Tuzel, Oncel},
  booktitle={Proceedings of the IEEE conference on computer vision and pattern recognition},
  pages={4490--4499},
  year={2018}
}

@inproceedings{Voxelrcnn,
  title={Voxel r-cnn: Towards high performance voxel-based 3d object detection},
  author={Deng, Jiajun and Shi, Shaoshuai and Li, Peiwei and Zhou, Wengang and Zhang, Yanyong and Li, Houqiang},
  booktitle={Proceedings of the AAAI Conference on Artificial Intelligence},
  volume={35},
  number={2},
  pages={1201--1209},
  year={2021}
}

@article{Second,
  title={Second: Sparsely embedded convolutional detection},
  author={Yan, Yan and Mao, Yuxing and Li, Bo},
  journal={Sensors},
  volume={18},
  number={10},
  pages={3337},
  year={2018},
  publisher={MDPI}
}

@inproceedings{Pointpillars,
  title={Pointpillars: Fast encoders for object detection from point clouds},
  author={Lang, Alex H and Vora, Sourabh and Caesar, Holger and Zhou, Lubing and Yang, Jiong and Beijbom, Oscar},
  booktitle={Proceedings of the IEEE/CVF conference on computer vision and pattern recognition},
  pages={12697--12705},
  year={2019}
}

@article{SATGCN,
  title={SAT-GCN: Self-attention graph convolutional network-based 3D object detection for autonomous driving},
  author={Wang, Li and Song, Ziying and Zhang, Xinyu and Wang, Chenfei and Zhang, Guoxin and Zhu, Lei and Li, Jun and Liu, Huaping},
  journal={Knowledge-Based Systems},
  volume={259},
  pages={110080},
  year={2023},
  publisher={Elsevier}
}

@inproceedings{LargeKernel3D,
  title={LargeKernel3D: Scaling Up Kernels in 3D Sparse CNNs},
  author={Chen, Yukang and Liu, Jianhui and Zhang, Xiangyu and Qi, Xiaojuan and Jia, Jiaya},
  booktitle={Proceedings of the IEEE/CVF Conference on Computer Vision and Pattern Recognition},
  pages={13488--13498},
  year={2023}
}

@article{transformer,
  title={Attention is all you need},
  author={Vaswani, Ashish and Shazeer, Noam and Parmar, Niki and Uszkoreit, Jakob and Jones, Llion and Gomez, Aidan N and Kaiser, {\L}ukasz and Polosukhin, Illia},
  journal={Advances in neural information processing systems},
  volume={30},
  year={2017}
}

@InProceedings{vsettransformer,
    author    = {He, Chenhang and Li, Ruihuang and Li, Shuai and Zhang, Lei},
    title     = {Voxel Set Transformer: A Set-to-Set Approach to 3D Object Detection From Point Clouds},
    booktitle = {Proceedings of the IEEE/CVF Conference on Computer Vision and Pattern Recognition (CVPR)},
    month     = {June},
    year      = {2022},
    pages     = {8417-8427}
}

@inproceedings{voxeltransformer,
  title={Voxel transformer for 3d object detection},
  author={Mao, Jiageng and Xue, Yujing and Niu, Minzhe and Bai, Haoyue and Feng, Jiashi and Liang, Xiaodan and Xu, Hang and Xu, Chunjing},
  booktitle={Proceedings of the IEEE/CVF International Conference on Computer Vision},
  pages={3164--3173},
  year={2021}
}

@inproceedings{Sparsetransformer,  
 title={Embracing Single Stride 3D Object Detector with Sparse Transformer}, 
 DOI={10.1109/cvpr52688.2022.00827}, 
 booktitle={2022 IEEE/CVF Conference on Computer Vision and Pattern Recognition (CVPR)}, 
 author={Fan, Lue and Pang, Ziqi and Zhang, Tianyuan and Wang, Yu-Xiong and Zhao, Hang and Wang, Feng and Wang, Naiyan and Zhang, Zhaoxiang}, 
 year={2022}, 
 month={Jun}, 
 language={en-US} 
 }

@article{pvcnn,
  title={Point-voxel cnn for efficient 3d deep learning},
  author={Liu, Zhijian and Tang, Haotian and Lin, Yujun and Han, Song},
  journal={Advances in Neural Information Processing Systems},
  volume={32},
  year={2019}
}

@inproceedings{PVGNet,  
 title={PVGNet: A Bottom-Up One-Stage 3D Object Detector with Integrated Multi-Level Features}, 
 DOI={10.1109/cvpr46437.2021.00329}, 
 booktitle={2021 IEEE/CVF Conference on Computer Vision and Pattern Recognition (CVPR)}, 
 author={Miao, Zhenwei and Chen, Jikai and Pan, Hongyu and Zhang, Ruiwen and Liu, Kaixuan and Hao, Peihan and Zhu, Jun and Wang, Yang and Zhan, Xin}, 
 year={2021}, 
 month={Jun}, 
 language={en-US} 
 }

@article{pvrcnn++,
  title={PV-RCNN++: Point-voxel feature set abstraction with local vector representation for 3D object detection},
  author={Shi, Shaoshuai and Jiang, Li and Deng, Jiajun and Wang, Zhe and Guo, Chaoxu and Shi, Jianping and Wang, Xiaogang and Li, Hongsheng},
  journal={International Journal of Computer Vision},
  volume={131},
  number={2},
  pages={531--551},
  year={2023},
  publisher={Springer}
}

@inproceedings{Std,
  title={Std: Sparse-to-dense 3d object detector for point cloud},
  author={Yang, Zetong and Sun, Yanan and Liu, Shu and Shen, Xiaoyong and Jia, Jiaya},
  booktitle={Proceedings of the IEEE/CVF international conference on computer vision},
  pages={1951--1960},
  year={2019}
}

@article{vpnet,
  title={VP-Net: Voxels as Points for 3D Object Detection},
  author={Song, Ziying and Wei, Haiyue and Jia, Caiyan and Xia, Yongchao and Li, Xiaokun and Zhang, Chao},
  journal={IEEE Transactions on Geoscience and Remote Sensing},
  year={2023},
  publisher={IEEE}
}

@inproceedings{brazil2019m3d,
  title={M3d-rpn: Monocular 3d region proposal network for object detection},
  author={Brazil, Garrick and Liu, Xiaoming},
  booktitle={Proceedings of the IEEE/CVF International Conference on Computer Vision},
  pages={9287--9296},
  year={2019}
}

@inproceedings{xu2018multi,
  title={Multi-level fusion based 3d object detection from monocular images},
  author={Xu, Bin and Chen, Zhenzhong},
  booktitle={Proceedings of the IEEE conference on computer vision and pattern recognition},
  pages={2345--2353},
  year={2018}
}

@inproceedings{simonelli2019disentangling,
  title={Disentangling monocular 3d object detection},
  author={Simonelli, Andrea and Bulo, Samuel Rota and Porzi, Lorenzo and L{\'o}pez-Antequera, Manuel and Kontschieder, Peter},
  booktitle={Proceedings of the IEEE/CVF International Conference on Computer Vision},
  pages={1991--1999},
  year={2019}
}

@inproceedings{yang2023bevheight,
  title={BEVHeight: A Robust Framework for Vision-based Roadside 3D Object Detection},
  author={Yang, Lei and Yu, Kaicheng and Tang, Tao and Li, Jun and Yuan, Kun and Wang, Li and Zhang, Xinyu and Chen, Peng},
  booktitle={Proceedings of the IEEE/CVF Conference on Computer Vision and Pattern Recognition},
  pages={21611--21620},
  year={2023}
}

@article{yang2023bevheight++,
  title={Bevheight++: Toward robust visual centric 3d object detection},
  author={Yang, Lei and Tang, Tao and Li, Jun and Chen, Peng and Yuan, Kun and Wang, Li and Huang, Yi and Zhang, Xinyu and Yu, Kaicheng},
  journal={arXiv preprint arXiv:2309.16179},
  year={2023}
}

@article{BEVStereo, title={BEVStereo: Enhancing Depth Estimation in Multi-View 3D Object Detection with Temporal Stereo}, volume={37}, url={https://ojs.aaai.org/index.php/AAAI/article/view/25234}, DOI={10.1609/aaai.v37i2.25234}, number={2}, journal={Proceedings of the AAAI Conference on Artificial Intelligence}, author={Li, Yinhao and Bao, Han and Ge, Zheng and Yang, Jinrong and Sun, Jianjian and Li, Zeming}, year={2023}, month={Jun.}, pages={1486-1494} }

@inproceedings{park2022time,
title={Time Will Tell: New Outlooks and A Baseline for Temporal Multi-View 3D Object Detection},
author={Jinhyung Park and Chenfeng Xu and Shijia Yang and Kurt Keutzer and Kris M. Kitani and Masayoshi Tomizuka and Wei Zhan},
booktitle={The Eleventh International Conference on Learning Representations },
year={2023},
url={https://openreview.net/forum?id=H3HcEJA2Um}
}

@inproceedings{detr,
  title={End-to-end object detection with transformers},
  author={Carion, Nicolas and Massa, Francisco and Synnaeve, Gabriel and Usunier, Nicolas and Kirillov, Alexander and Zagoruyko, Sergey},
  booktitle={European conference on computer vision},
  pages={213--229},
  year={2020},
  organization={Springer}
}

@article{deformabledetr,
  title={Deformable detr: Deformable transformers for end-to-end object detection},
  author={Zhu, Xizhou and Su, Weijie and Lu, Lewei and Li, Bin and Wang, Xiaogang and Dai, Jifeng},
  journal={arXiv preprint arXiv:2010.04159},
  year={2020}
}

@inproceedings{wang2022detr3d,
  title={Detr3d: 3d object detection from multi-view images via 3d-to-2d queries},
  author={Wang, Yue and Guizilini, Vitor Campagnolo and Zhang, Tianyuan and Wang, Yilun and Zhao, Hang and Solomon, Justin},
  booktitle={Conference on Robot Learning},
  pages={180--191},
  year={2022},
  organization={PMLR}
}

@inproceedings{jiang2023polarformer,
  title={Polarformer: Multi-camera 3d object detection with polar transformer},
  author={Jiang, Yanqin and Zhang, Li and Miao, Zhenwei and Zhu, Xiatian and Gao, Jin and Hu, Weiming and Jiang, Yu-Gang},
  booktitle={Proceedings of the AAAI Conference on Artificial Intelligence},
  volume={37},
  number={1},
  pages={1042--1050},
  year={2023}
}

@inproceedings{liu2022petr,
  title={Petr: Position embedding transformation for multi-view 3d object detection},
  author={Liu, Yingfei and Wang, Tiancai and Zhang, Xiangyu and Sun, Jian},
  booktitle={European Conference on Computer Vision},
  pages={531--548},
  year={2022},
  organization={Springer}
}

@article{lin2023sparse4d,
  title={Sparse4d v2: Recurrent temporal fusion with sparse model},
  author={Lin, Xuewu and Lin, Tianwei and Pei, Zixiang and Huang, Lichao and Su, Zhizhong},
  journal={arXiv preprint arXiv:2305.14018},
  year={2023}
}

@inproceedings{liu2023petrv2,
  title={Petrv2: A unified framework for 3d perception from multi-camera images},
  author={Liu, Yingfei and Yan, Junjie and Jia, Fan and Li, Shuailin and Gao, Aqi and Wang, Tiancai and Zhang, Xiangyu},
  booktitle={Proceedings of the IEEE/CVF International Conference on Computer Vision},
  pages={3262--3272},
  year={2023}
}

@inproceedings{bevformerv2,
  title={BEVFormer v2: Adapting Modern Image Backbones to Bird's-Eye-View Recognition via Perspective Supervision},
  author={Yang, Chenyu and Chen, Yuntao and Tian, Hao and Tao, Chenxin and Zhu, Xizhou and Zhang, Zhaoxiang and Huang, Gao and Li, Hongyang and Qiao, Yu and Lu, Lewei and others},
  booktitle={Proceedings of the IEEE/CVF Conference on Computer Vision and Pattern Recognition},
  pages={17830--17839},
  year={2023}
}

@article{fsd,
  title={Fully sparse 3d object detection},
  author={Fan, Lue and Wang, Feng and Wang, Naiyan and Zhang, Zhao-Xiang},
  journal={Advances in Neural Information Processing Systems},
  volume={35},
  pages={351--363},
  year={2022}
}

@inproceedings{bevformer,
  title={{Bevformer: Learning bird’s-eye-view representation from multi-camera images via spatiotemporal transformers}},
  author={Li, Zhiqi and Wang, Wenhai and Li, Hongyang and Xie, Enze and Sima, Chonghao and Lu, Tong and Qiao, Yu and Dai, Jifeng},
  booktitle={European conference on computer vision},
  pages={1--18},
  year={2022},
  organization={Springer}
}

@ARTICLE{BEVFormerpami,
  author={Li, Zhiqi and Wang, Wenhai and Li, Hongyang and Xie, Enze and Sima, Chonghao and Lu, Tong and Yu, Qiao and Dai, Jifeng},
  journal={IEEE Transactions on Pattern Analysis and Machine Intelligence}, 
  title={BEVFormer: Learning Bird’s-Eye-View Representation From LiDAR-Camera via Spatiotemporal Transformers}, 
  year={2025},
  volume={47},
  number={3},
  pages={2020-2036},
  doi={10.1109/TPAMI.2024.3515454}}

@inproceedings{swimtransformer,
  title={Swin transformer: Hierarchical vision transformer using shifted windows},
  author={Liu, Ze and Lin, Yutong and Cao, Yue and Hu, Han and Wei, Yixuan and Zhang, Zheng and Lin, Stephen and Guo, Baining},
  booktitle={Proceedings of the IEEE/CVF international conference on computer vision},
  pages={10012--10022},
  year={2021}
}

@article{paszke2019pytorch,
  title={Pytorch: An imperative style, high-performance deep learning library},
  author={Paszke, Adam and Gross, Sam and Massa, Francisco and Lerer, Adam and Bradbury, James and Chanan, Gregory and Killeen, Trevor and Lin, Zeming and Gimelshein, Natalia and Antiga, Luca and others},
  journal={Advances in neural information processing systems},
  volume={32},
  year={2019}
}

@misc{openpcdet,
	title={OpenPCDet: An Open-source Toolbox for 3D Object Detection from Point Clouds},
	author={OpenPCDet Development Team},
	howpublished = {\url{https://github.com/open-mmlab/OpenPCDet}},
	year={2020}
}

@inproceedings{resnet,
  title={Deep residual learning for image recognition},
  author={He, Kaiming and Zhang, Xiangyu and Ren, Shaoqing and Sun, Jian},
  booktitle={Proceedings of the IEEE conference on computer vision and pattern recognition},
  pages={770--778},
  year={2016}
}

@inproceedings{wang2021fcos3d,
  title={Fcos3d: Fully convolutional one-stage monocular 3d object detection},
  author={Wang, Tai and Zhu, Xinge and Pang, Jiangmiao and Lin, Dahua},
  booktitle={Proceedings of the IEEE/CVF international conference on computer vision},
  pages={913--922},
  year={2021}
}

@inproceedings{dcn,
  title={Deformable convolutional networks},
  author={Dai, Jifeng and Qi, Haozhi and Xiong, Yuwen and Li, Yi and Zhang, Guodong and Hu, Han and Wei, Yichen},
  booktitle={Proceedings of the IEEE international conference on computer vision},
  pages={764--773},
  year={2017}
}

@inproceedings{fpn,
  title={Feature pyramid networks for object detection},
  author={Lin, Tsung-Yi and Doll{\'a}r, Piotr and Girshick, Ross and He, Kaiming and Hariharan, Bharath and Belongie, Serge},
  booktitle={Proceedings of the IEEE conference on computer vision and pattern recognition},
  pages={2117--2125},
  year={2017}
}

@inproceedings{maskrcnn,
  title={Mask r-cnn},
  author={He, Kaiming and Gkioxari, Georgia and Doll{\'a}r, Piotr and Girshick, Ross},
  booktitle={Proceedings of the IEEE international conference on computer vision},
  pages={2961--2969},
  year={2017}
}

@article{CBGS,
  title={Class-balanced grouping and sampling for point cloud 3d object detection},
  author={Zhu, Benjin and Jiang, Zhengkai and Zhou, Xiangxin and Li, Zeming and Yu, Gang},
  journal={arXiv preprint arXiv:1908.09492},
  year={2019}
}

@article{adam,
  title={Adam: A method for stochastic optimization},
  author={Kingma, Diederik P and Ba, Jimmy},
  journal={arXiv preprint arXiv:1412.6980},
  year={2014}
}

@inproceedings{Centerpoint,  
 title={Center-based 3D Object Detection and Tracking.}, 
 DOI={10.1109/cvpr46437.2021.01161}, 
 booktitle={2021 IEEE/CVF Conference on Computer Vision and Pattern Recognition (CVPR)}, 
 author={Yin, Tianwei and Zhou, Xingyi and Krahenbuhl, Philipp}, 
 year={2021}, 
 month={Jun}, 
 language={en-US} 
 }

@InProceedings{thiktwice,
    author    = {Jia, Xiaosong and Wu, Penghao and Chen, Li and Xie, Jiangwei and He, Conghui and Yan, Junchi and Li, Hongyang},
    title     = {Think Twice Before Driving: Towards Scalable Decoders for End-to-End Autonomous Driving},
    booktitle = {Proceedings of the IEEE/CVF Conference on Computer Vision and Pattern Recognition (CVPR)},
    month     = {June},
    year      = {2023},
    pages     = {21983-21994}
}

@article{shang2025drivedpo,
  title={Drivedpo: Policy learning via safety dpo for end-to-end autonomous driving},
  author={Shang, Shuyao and Chen, Yuntao and Wang, Yuqi and Li, Yingyan and Zhang, Zhaoxiang},
  journal={arXiv preprint arXiv:2509.17940},
  year={2025}
}

@article{Raw2Drive,
  title={Raw2Drive: Reinforcement learning with aligned world models for end-to-end autonomous driving (in carla v2)},
  author={Yang, Zhenjie and Jia, Xiaosong and Li, Qifeng and Yang, Xue and Yao, Maoqing and Yan, Junchi},
  journal={arXiv preprint arXiv:2505.16394},
  year={2025}
}

\end{document}